\documentclass{article}
\PassOptionsToPackage{table}{xcolor}
\usepackage[T1]{fontenc}
\usepackage{iclr2027_conference,times}

\usepackage{amsmath,amsfonts,bm}

\def\eqref#1{equation~\ref{#1}}

\def\1{\bm{1}}

\DeclareMathAlphabet{\mathsfit}{\encodingdefault}{\sfdefault}{m}{sl}
\SetMathAlphabet{\mathsfit}{bold}{\encodingdefault}{\sfdefault}{bx}{n}

\usepackage{hyperref}
\usepackage{url}
\usepackage{amsmath,amssymb}
\usepackage{graphicx}
\usepackage{fontawesome5}
\usepackage{capt-of}
\usepackage{subcaption}
\usepackage{tabularx}
\usepackage{multirow}
\usepackage{array}
\usepackage{enumitem}
\usepackage{tabularray}
\SetTblrTemplate{caption}{plain}
\SetTblrTemplate{capcont}{plain}
\usepackage{adjustbox}
\UseTblrLibrary{booktabs}
\usepackage{booktabs,fvextra}
\usepackage{longtable}
\usepackage{wrapfig}
\usepackage{colortbl}
\usepackage[table]{xcolor}
\usepackage[most]{tcolorbox}
\definecolor{promptblue}{RGB}{0,0,180}

\definecolor{evgopsdblue}{RGB}{232,240,251} %
\newcommand{\increase}[1]{%
    \rlap{%
        {\scriptsize\textcolor{green!60!black}{+#1}}%
    }%
}
\newcommand{\decrease}[1]{%
    \rlap{%
        {\scriptsize\textcolor{red!75!black}{-#1}}%
    }%
}

\newcommand{\stepscreenshot}[1]{%
    \adjustbox{valign=t}{%
        \includegraphics[width=\linewidth]{#1}%
    }%
}

\newcommand{\trajectorytool}[1]{%
    \par
    \texttt{\textless tool\_call\textgreater}%
    \par
    #1%
    \par
    \texttt{\textless/tool\_call\textgreater}%
}

\newcommand{\expertannotation}[4]{%
    \{"consistency": #1, "effectiveness": #2,%
    \par
    "guide": "#3",%
    \par
    "avoid": "#4"\}%
}

\providecommand{\stepscreenshot}[1]{%
    \adjustbox{valign=t}{\includegraphics[width=\linewidth]{#1}}%
}
\providecommand{\caseScreenshot}[1]{%
    \IfFileExists{#1}{\stepscreenshot{#1}}{%
        \adjustbox{valign=t}{%
            \fbox{\parbox[c][24mm][c]{\dimexpr\linewidth-2\fboxsep-2\fboxrule\relax}{%
                \centering\small\itshape Screenshot placeholder}}%
        }%
    }%
}
\providecommand{\caseToolCall}[1]{%
    \par\texttt{\textless tool\_call\textgreater}%
    \par #1%
    \par\texttt{\textless/tool\_call\textgreater}%
}

\title{ComputerSD: Online Self-Distillation from Real-Time Feedback for Computer-Use Agents}

\newif\ifComputerSDPreprint
\ComputerSDPreprinttrue
\author{
\begin{minipage}[t]{\dimexpr\textwidth-2\tabcolsep\relax}
\centering
Yong Du\textsuperscript{1}, Tongbo Chen\textsuperscript{1},
Zhengxi Lu\textsuperscript{1}, Yizhou Liu\textsuperscript{1}, Bofan Chen\textsuperscript{1},\\
Tao Jiang\textsuperscript{2}, Wenhao Xu\textsuperscript{2}, Yongliang Shen\textsuperscript{1\textdagger}\\
{\normalfont\small
\textsuperscript{1}Zhejiang University \quad
\textsuperscript{2}Ant Group\\
\texttt{\{duyong123,syl\}@zju.edu.cn} \\[0.5em]
\faGithub\ Code: \href{https://github.com/ZJU-REAL/ComputerSD}{\texttt{\textcolor{cyan}{https://github.com/ZJU-REAL/ComputerSD}}}
}
\end{minipage}%
}
\ifComputerSDPreprint
\iclrfinalcopy
\hypersetup{
  pdftitle={ComputerSD: Online Self-Distillation from Real-Time Feedback for Computer-Use Agents},
  pdfauthor={Yong Du, Tongbo Chen, Zhengxi Lu, Yizhou Liu, Bofan Chen, Tao Jiang, Wenhao Xu, Yongliang Shen}
}
\fi

\begin{document}
\maketitle
\ifComputerSDPreprint
\lhead{Preprint}
\begingroup
\renewcommand{\thefootnote}{\fnsymbol{footnote}}
\footnotetext[2]{Corresponding Author}
\endgroup
\fi

\begin{abstract}
Online training enables computer-use agents (CUAs) to improve through interaction with executable environments.
However, existing methods primarily rely on sparse outcome rewards, which provide no supervision for intermediate actions.
On-policy self-distillation (OPSD) offers token-level learning signals through privileged rescoring, but directly applying it to CUA online training presents two challenges: fixed guidance may become misaligned with the student's current state, and guidance-induced probability shifts may conflict with step-level correctness.
We introduce \textbf{ComputerSD}, an online self-distillation method for CUAs that converts real-time feedback from executed GUI transitions into guidance for policy learning.
A fine-tuned GUI analyzer produces guidance and a step-level value score after each action; the guidance provides privileged context, while the score regulates the resulting OPSD signals.
ComputerSD jointly optimizes token-level OPSD and trajectory-level GRPO in a fully asynchronous training framework.
On OSWorld-Verified, ComputerSD outperforms outcome-only GRPO by 1.9 and 4.1 percentage points on the general-purpose Qwen3-VL-8B-Thinking and specialized EvoCUA-8B backbones, respectively. Evaluation in out-of-distribution settings further supports the generalizability of ComputerSD. These results demonstrate the effectiveness of learning from real-time feedback through online self-distillation for CUAs.
\end{abstract}

\begin{figure}[h]
    \centering
    \includegraphics[width=\linewidth]{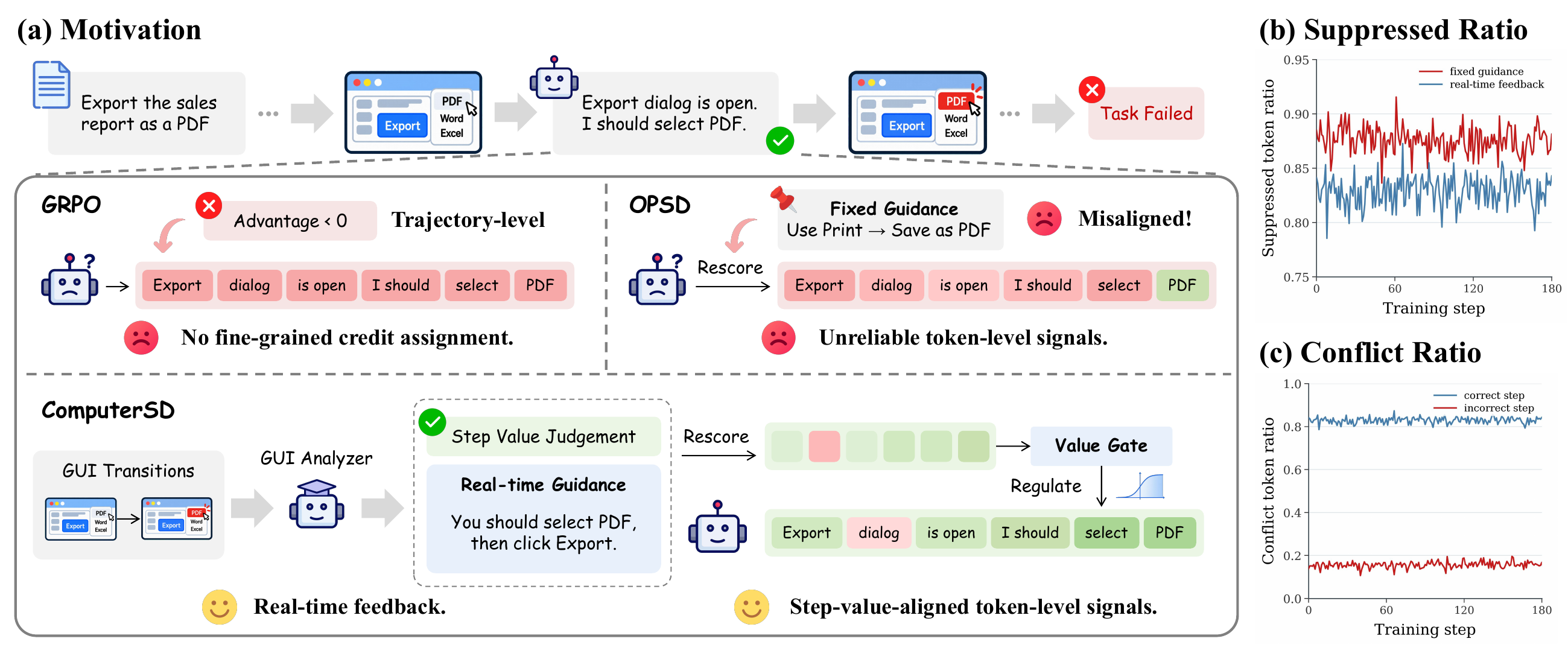}
    \caption{\textbf{Motivation and preliminaries.} \textbf{(a)} Comparison of GRPO, OPSD and ComputerSD. During online training, we measure \textbf{(b)} \emph{suppressed ratio} (the ratio of tokens at correct steps whose log-probability is lowered) under fixed guidance and real-time feedback and \textbf{(c)} \emph{conflict ratio} (the ratio of tokens whose shift contradicts the step-level judgment) in correct and incorrect steps.}
    \label{fig:motivation}
\end{figure}

\section{Introduction}
\label{sec:introduction}

Developing computer-use agents (CUAs) capable of operating graphical user interfaces (GUIs) is an essential step toward autonomous computer use~\citep{qin2025ui,wang2025opencuaopenfoundationscomputeruse,xu2024aguvis}.
To bridge the gap between offline demonstrations and real-world interaction, recent studies have increasingly turned to online training in interactive environments~\citep{wang2025uitars2technicalreportadvancing,zhou2025maiui,lai2025computerrl,lu2025uis1}.
These methods rely primarily on outcome rewards produced by environment verifiers.
Such rewards indicate whether a task is completed but not which of the many actions in an episode were productive, redundant, or erroneous, making credit assignment a persistent challenge in online CUA training~\citep{chen2025guishepherd,feng2025gigpo}.

On-policy self-distillation (OPSD), recently extended to multi-turn agents~\citep{lu2026selfdistilledagenticreinforcementlearning,yang2026opid,wu2026seed}, recovers this fine-grained credit at the token level~\citep{zhao2026opsd,hubotter2026sdpo}: the policy rescores its own sampled response under privileged information such as a reference trajectory or task-relevant skills~\citep{wang2026skillsd,lu2026skill0}, and the resulting log-probability shifts serve as supervision on which tokens to reinforce or suppress.

Applying OPSD to online CUA training, however, raises two problems, illustrated in Figure~\ref{fig:motivation}a.
The first problem is that privileged information fixed before the rollout becomes \emph{misaligned} with the student's state.
A CUA task usually admits multiple valid solutions, and once the student leaves the path that a reference trajectory or pre-written guidance assumes, the guidance no longer matches the observed state~\citep{shenfeld2026sdft,harne2026privileged,liu2026smrcsd}.
Distilling toward such guidance then suppresses the tokens of steps that are correct on the student's own path (Figure~\ref{fig:motivation}b).
The second problem is that the token-level signals are \emph{unreliable} even when the guidance is relevant.
Guidance shifts the probability of every token in the response, and the shift at a given token need not agree with whether the step was correct: over 80\% of tokens at correct steps are suppressed and nearly 20\% of tokens at incorrect ones are reinforced throughout training (Figure~\ref{fig:motivation}c).
Without regulation, these signals would pull the policy away from the task objective.

Our key insight is that the real-time observation after each action can resolve both problems at once.
Once an action is executed, its consequence appears on the next screen, and feedback written from this transition, which we call \emph{real-time feedback}, describes exactly the state the student reached and removes the misalignment.
The same feedback can also judge whether the action was correct, so \emph{the signal that constructs the privileged context can also decide how far to trust the token-level supervision it induces}.
Recent work likewise conditions the teacher on the post-action observation~\citep{liu2026hero,li2026ghd,wang2026openclawrl}, but stops at state matching, which still leaves the induced shifts unreliable. We additionally use the feedback's own judgment of the step to regulate every token-level signal it produces.

Building on this insight, we propose \textbf{ComputerSD}, an online self-distillation method that converts real-time feedback into value-gated token-level supervision (Figure~\ref{fig:method}).
After each action, a GUI analyzer reads the screenshots before and after it and returns a step-level value score together with guidance on what to do and what to avoid from the pre-action state; since the untuned policy often misjudges GUI transitions, we fine-tune the analyzer on expert annotations.
The guidance then serves as the privileged context for rescoring, and a value gate weights each token's shift by its agreement with the value score.
We optimize this token-level objective jointly with trajectory-level GRPO, so that step-level feedback refines credit within each episode while outcome rewards keep learning anchored to task success.
To prevent per-step analysis from stalling rollout, we design a fully asynchronous pipeline for ComputerSD.

On OSWorld-Verified~\citep{xie2024osworldbenchmarkingmultimodalagents}, ComputerSD outperforms outcome-only GRPO on both the general-purpose Qwen3-VL-8B-Thinking~\citep{bai2025qwen3vltechnicalreport} and the computer-use model EvoCUA-8B~\citep{xue2026evocuaevolvingcomputeruse}, raising the success rate from 37.9\% to 39.8\% and from 43.8\% to 47.9\%.
The gains are largest on application categories held out from training (from 21.6\% to 27.5\% and from 30.2\% to 35.4\%), indicating that the step-level supervision transfers beyond the training distribution.
Ablations mirror the two failures identified above: either replacing real-time feedback with fixed guidance or removing the value gate drops performance below GRPO on Qwen3-VL-8B-Thinking, demonstrating the necessity of our design.
Finally, the asynchronous pipeline raises training throughput fivefold over its synchronous counterpart, keeping per-step feedback affordable.

Our main contributions are summarized as follows:

\begin{itemize}
    \item We propose ComputerSD, an online self-distillation method in which a fine-tuned GUI analyzer writes real-time feedback after each action and a value gate weights each token-level signal by its agreement with the analyzer's judgment of the step.
    \item We develop a fully asynchronous training pipeline that overlaps environment interaction, GUI analysis, privileged rescoring, and policy optimization, so that rollout workers keep collecting trajectories while each executed step is analyzed and rescored.
    \item We show on OSWorld-Verified with two 8B backbones with different levels of computer-use specialization that ComputerSD outperforms outcome-only GRPO, and that removing either real-time feedback or the value gate lowers performance below GRPO.
\end{itemize}

\section{Related Work}
\label{sec:related_work}

\paragraph{Online Training for Computer-Use Agents.}
Recent CUAs improve by scaling verifiable training tasks~\citep{xue2026evocuaevolvingcomputeruse,lv2026scalecua} and by online reinforcement learning in executable environments, where ComputerRL and UI-TARS-2 run rollouts over parallel environments~\citep{lai2025computerrl,wang2025uitars2technicalreportadvancing} and DART decouples rollout from training~\citep{li2025dart}.
To supervise intermediate steps, GUI-Shepherd learns a process reward model~\citep{chen2025guishepherd} and GiGPO estimates step-level advantages from repeated states~\citep{feng2025gigpo}; both assign a single scalar to each step.
ComputerSD instead turns the feedback on each executed step into token-level supervision, and its asynchronous pipeline overlaps this analysis with rollout and training.

\paragraph{On-Policy Self-Distillation.}
On-policy self-distillation (OPSD) obtains a teacher by conditioning the policy on privileged information and distills its token-level predictions on the policy's own samples~\citep{agarwal2024gkd,zhao2026opsd,hubotter2026sdpo}.
For multi-turn agents, OPID and SEED derive this information from completed trajectories~\citep{yang2026opid,wu2026seed}, and SDAR gates the resulting signals by the teacher--student gap~\citep{lu2026selfdistilledagenticreinforcementlearning}.
Because information fixed in advance can mismatch the states the student reaches, HERO conditions the teacher on a diagnosis of the observation after each action~\citep{liu2026hero}, and GHD and OpenClaw-RL bring this idea to GUI agents via the next screenshot and via hints from a prompted judge~\citep{li2026ghd,wang2026openclawrl}.
ComputerSD obtains both the guidance and a step-level judgment from a fine-tuned GUI analyzer and gates each token-level signal by this judgment, a check at the level of the step rather than of the teacher--student gap or the trajectory outcome~\citep{lin2026opdvr}.

\section{Method}\label{sec:methods}
\begin{figure}[t]
    \centering
    \includegraphics[width=\linewidth]{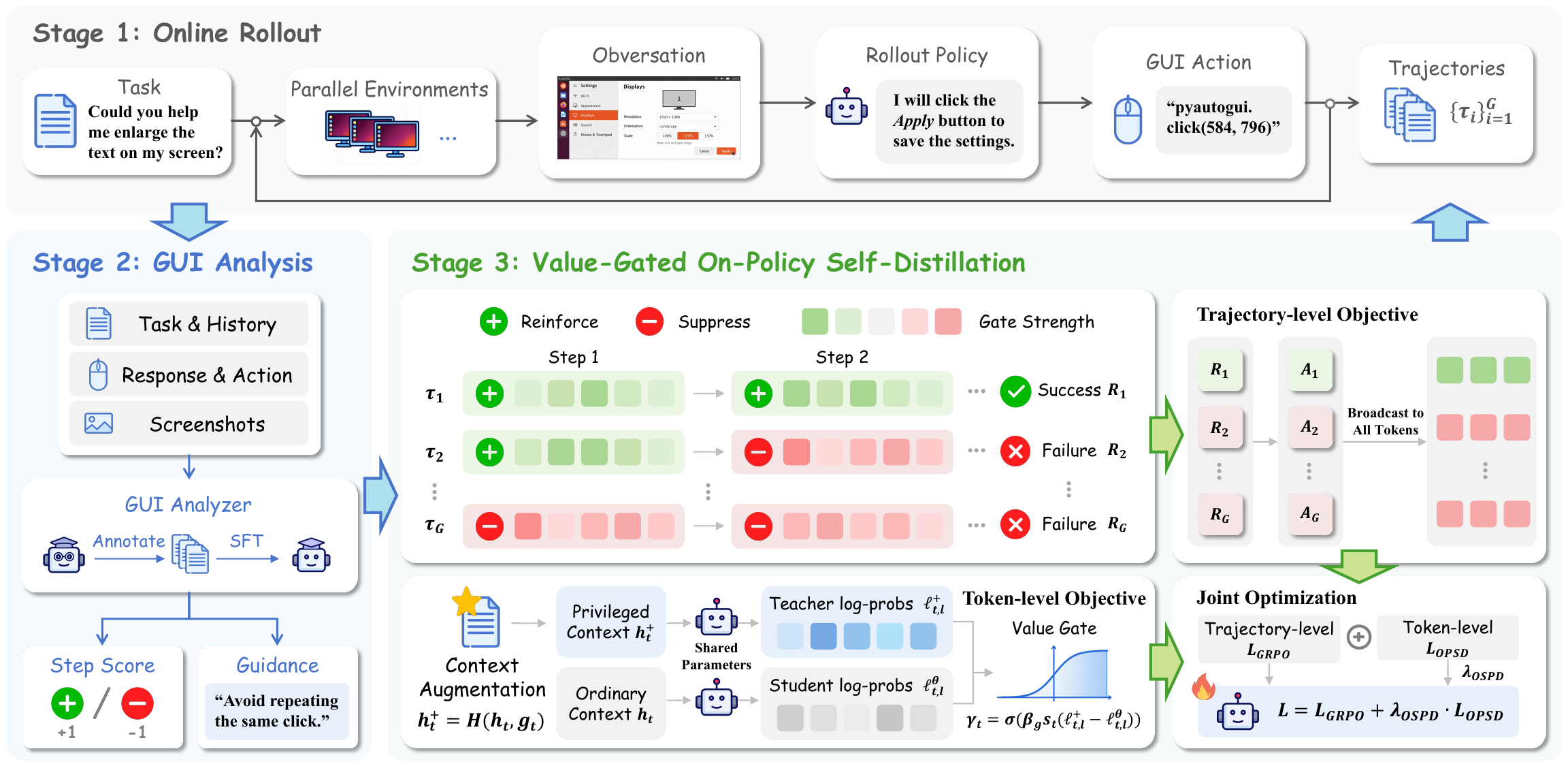}
    \caption{Overview of ComputerSD. The base policy samples trajectories online, while a GUI analyzer provides step-level value scores and guidance. Privileged rescoring produces token-level supervision, which is combined with trajectory-level GRPO to update the policy.}
    \label{fig:method}
\end{figure}

We present \textbf{ComputerSD}, an online self-distillation method for computer-use agents. As illustrated in Figure~\ref{fig:method}, ComputerSD first samples multiple trajectories through online interaction with parallel computer environments, followed by a GUI analyzer that provides step-level guidance based on the interaction process, which is then used to value-gate the token-level reward signals and control their strength according to the step-level value. Finally, the resulting token-level signals are combined with environment rewards to optimize the policy model.

\subsection{Problem Formulation}
\label{sec:problem}

We formulate CUA task execution as a partially observable Markov decision process. Given a task instruction $x$, the agent receives an observation $o_t$ at each time step $t$, which may include a screenshot, an accessibility tree, or other environment feedback. From the ordinary context $h_t=(x,o_0,y_0,\ldots,o_t)$, the student samples $y_t=(r_t,a_t)\sim\pi_\theta(\cdot\mid h_t)$, where $r_t$ denotes reasoning text and $a_t$ is an executable action. Executing $a_t$ yields the next observation $o_{t+1}$. A trajectory is represented as
\begin{equation}
 \tau=\bigl(x,\{h_t,y_t,o_{t+1}\}_{t=0}^{T-1},R\bigr),
 \label{eq:trajectory}
\end{equation}
where $T$ is the trajectory length and $R$ is the terminal reward from the environment verifier.

\subsection{GUI Analyzer Supervised Fine-Tuning}
\label{sec:expert}

To turn step-level environment feedback into structured, learnable signals for online policy training, we train a lightweight GUI analyzer that analyzes GUI transitions to provide real-time feedback.

\paragraph{Online trajectory collection.}
We first collect trajectories on a subset of OSWorld~\citep{xie2024osworldbenchmarkingmultimodalagents} using a base policy $\pi_{\phi}$. For each task, the policy samples $K$ trajectories, yielding a diverse pool of successful and unsuccessful interactions with varied action choices, state transitions, and failure modes. These trajectories provide the step contexts for subsequent expert annotation.

\paragraph{Expert annotation.}
A strong expert model is then used to annotate the collected trajectories. For each step $t$, the expert receives the step context $z_t$ and produces a value score $\hat{s}_t$ and guidance $\hat{g}_t$ according to $(\hat s_t,\hat g_t)\sim\pi_{\mathrm{exp}}(\cdot\mid z_t)$, which together form the training dataset
\begin{equation}
\mathcal D_{\mathrm{exp}}
=\{(z_j,\hat s_j,\hat g_j)\}_{j=1}^{N},
\label{eq:expert-data}
\end{equation}
where $N$ is the total number of annotated steps across all collected trajectories. Here, $\hat s_j$ and $\hat g_j$ are the corresponding score and guidance targets.

\paragraph{Supervised fine-tuning.}
Finally, we initialize the GUI analyzer $\pi_{\phi}$ from the same base policy and fine-tune it to predict the step-level value score and guidance. The optimization objective is to maximize the likelihood of the expert annotations:
\begin{equation}
\mathcal L_{\mathrm{SFT}}(\phi)
=-\mathbb E_{(z,\hat s,\hat g)\sim\mathcal D_{\mathrm{exp}}}
\left[\log\pi_\phi(\hat s,\hat g\mid z)\right].
\label{eq:sft}
\end{equation}
After supervised fine-tuning, the GUI analyzer is frozen and used to provide structured real-time feedback during subsequent online policy training.

\subsection{Value-Gated On-Policy Self-Distillation}
\label{sec:evg}

To improve the reliability of OPSD signals during online training, the trained GUI analyzer provides step-level feedback to gate token-level OPSD signals, which are combined with trajectory-level environment rewards for policy optimization.

\paragraph{Trajectory-level GRPO objective.}
For each task $x$, the policy samples a group of $G$ trajectories in parallel online environments. Based on their terminal rewards, the group-relative advantage $A_i$ is computed. The trajectory-level GRPO objective is computed as:
\begin{equation}
\mathcal L_{\mathrm{GRPO}}(\theta)
=-\mathbb E_{i,t,l}\left[
\min\!\left(\rho_{i,t,l}(\theta)A_i,
\operatorname{clip}\!\left(\rho_{i,t,l}(\theta),
1-\epsilon,1+\epsilon\right)A_i\right)
\right]+\beta_{\mathrm{KL}}\mathcal L_{\mathrm{KL}}(\theta),
\label{eq:grpo-loss}
\end{equation}
where $\rho_{i,t,l}(\theta)$ is the token-level importance sampling ratio, $\epsilon$ is the clipping threshold, and $\beta_{\mathrm{KL}}$ is the KL regularization coefficient.

\paragraph{Value-gated on-policy self-distillation objective.}
For the $i$-th trajectory, the GUI analyzer generates step-level real-time feedback $(s_{i,t}, g_{i,t})$ for each executed step $t$. The guidance $g_{i,t}$ is added to the ordinary context $h_{i,t}$ as privileged information, yielding $h_{i,t}^{+}=(h_{i,t},g_{i,t})$.

We then compute the token-level log-probabilities of the sampled response $y_{i,t}$ under two different contexts. The log-probability gap reflects how real-time guidance changes the likelihood of the original sampled response and is computed as:
\begin{equation}
\delta_{i,t,l}
=\log\pi_\theta(y_{i,t,l}\mid h^{+}_{i,t},y_{i,t,<l})
-\log\pi_\theta(y_{i,t,l}\mid h_{i,t},y_{i,t,<l}),
\label{eq:teacher-logprob}
\end{equation}
where $l$ indexes tokens in the sampled response and $y_{i,t,<l}$ denotes the original sampled prefix.

The log-probability shifts induced by privileged guidance are not always reliable. To improve the reliability of token-level OPSD signals, we introduce a value gate that modulates each signal according to its consistency with the step-level value judgment. Specifically, we define $\ell_{i,t,l}$ as the value-weighted log-probability gap and compute the value gate $\gamma_{i,t,l}$ as follows:
\begin{equation}
\ell_{i,t,l}=s_{i,t}\delta_{i,t,l},
\quad
\gamma_{i,t,l}=\sigma(\beta_{\mathrm{gate}}\ell_{i,t,l}),
\label{eq:value-gate}
\end{equation}
where $\sigma$ is the logistic sigmoid function, and $\beta_{\mathrm{gate}}$ denotes the gate sharpness. The resulting gate regulates the token-level OPSD signals, reinforcing signals aligned with the step-level value judgment and suppressing those that are misaligned. The value-gated OPSD objective is computed as:
\begin{equation}
\mathcal L_{\mathrm{OPSD}}(\theta)
=\mathbb E_{i,t,l}
\left[
\gamma_{i,t,l}\ell_{i,t,l}
\right].
\label{eq:opd-loss}
\end{equation}

\paragraph{Joint training objective.}
The final ComputerSD objective combines token-level OPSD with trajectory-level GRPO:
\begin{equation}
\mathcal L_{\mathrm{ComputerSD}}(\theta)
=\mathcal L_{\mathrm{GRPO}}(\theta)
+\lambda_{\mathrm{OPSD}}\mathcal L_{\mathrm{OPSD}}(\theta),
\label{eq:joint-loss}
\end{equation}
where $\lambda_{\mathrm{OPSD}}$ controls the contribution of the OPSD signals.

\begin{figure}[t]
\centering
\includegraphics[width=\linewidth]{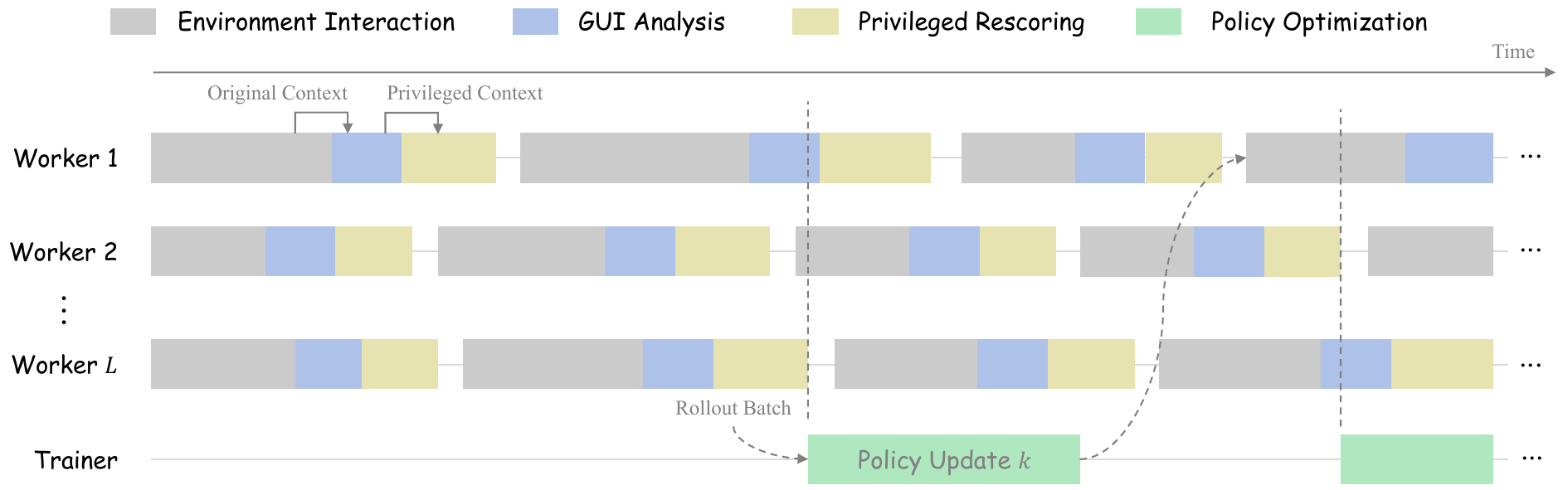}
\caption{Illustration of the fully asynchronous online training framework.}
\label{fig:async}
\end{figure}

\paragraph{Asynchronous online training.}
To improve training efficiency, we asynchronously overlap environment interaction, GUI analysis, privileged rescoring, and policy optimization, as illustrated in Figure~\ref{fig:async}. GUI analysis and privileged rescoring proceed alongside rollout, while the trainer updates the policy once a trajectory batch is collected. The updated parameters are then asynchronously published to the rollout workers for subsequent interaction.

\section{Experiments}
\label{sec:experiments}

\subsection{Experimental Settings}
\label{sec:experimental-settings}

\paragraph{Models.}
We initialize the GUI analyzer from Qwen3-VL-8B-Thinking~\citep{bai2025qwen3vltechnicalreport} and use Kimi K3~\citep{kimiteam2026kimik3openfrontier} as the expert model to annotate the collected trajectories. We then apply ComputerSD to two 8B-scale models: a general-purpose model Qwen3-VL-8B-Thinking~\citep{bai2025qwen3vltechnicalreport} and a specialized model EvoCUA-8B~\citep{xue2026evocuaevolvingcomputeruse}. This allows us to assess ComputerSD's effectiveness across policies with different levels of computer-use specialization.

\paragraph{Training and evaluation datasets.}
We conduct online training on OSWorld-Verified~\citep{xie2024osworldbenchmarkingmultimodalagents}, excluding tasks in the \textit{Multiple Apps} and \textit{Chrome} categories for out-of-distribution (OOD) evaluation. The evaluation set contains 222 in-domain tasks and 139 OOD tasks. We further evaluate the cross-platform generalization of ComputerSD on WindowsAgentArena~\citep{bonatti2024windowsagentarenaevaluating}. For both benchmarks, we report task success rates determined by their official environment verifiers.

\paragraph{Implementation details.}
During training, the rollout policy samples 8 trajectories for each of 4 tasks, yielding a batch of 32 trajectories. All training runs for 180 policy updates. Following SDAR~\citep{lu2026selfdistilledagenticreinforcementlearning}, we set the OPSD loss coefficient $\lambda_{\mathrm{OPSD}}$ to 0.01 and the gate sharpness $\beta_{\mathrm{gate}}$ to 5. Considering training efficiency, we set the maximum number of interaction steps to 30 during training and 50 during evaluation. During evaluation, models trained with ComputerSD use only the ordinary context, without GUI analyzer calls or additional guidance. Additional training details are provided in Appendix~\ref{app:training_details}.

\begin{table*}[t]
    \centering
    \small
    \setlength{\tabcolsep}{8pt}
    \caption{Performance comparison on OSWorld-Verified. Success rate is reported as Pass@1. Our experimental results are averaged over three independent evaluation runs to mitigate variance in online environments, while results for other models are taken from their official reports.}
    \label{tab:main-results}
    \begin{tabular}{lccc}
        \toprule
        \textbf{Model} & \textbf{Type} & \textbf{Max Steps} & \textbf{Success Rate (\%)} \\
        \midrule
        \rowcolor{gray!10}
        \multicolumn{4}{l}{\textit{Proprietary Models}} \\
        OpenAI CUA~\citep{cua2025} & Specialized & 50 & 31.3 \\
        Seed1.5-VL~\citep{guo2025seed15vltechnicalreport} & General & 100 & 36.7 \\
        Step-GUI-8B~\citep{yan2025stepguitechnicalreport} & Specialized & 100 & 40.2 \\
        Qwen3-VL-Flash~\citep{bai2025qwen3vltechnicalreport} & General & 100 & 41.6 \\
        UI-TARS-1.5~\citep{qin2025ui} & Specialized & 100 & 42.5 \\
        Claude-4-Sonnet~\citep{anthropic2025claude4} & General & 100 & 43.9 \\
        UI-TARS-2~\citep{wang2025uitars2technicalreportadvancing} & Specialized & 100 & 47.5 \\
        Claude-4.5-Sonnet~\citep{anthropic2025claudesonnet45} & General & 100 & 62.9 \\
        \midrule
        \rowcolor{gray!10}
        \multicolumn{4}{l}{\textit{Open-Source Models}} \\
        ScaleCUA-32B~\citep{liu2025scalecuascalingopensourcecomputer} & Specialized & 50 & 17.7 \\
        UI-TARS-72B-DPO~\citep{qin2025ui} & Specialized & 50 & 24.6 \\
        OpenCUA-7B~\citep{wang2025opencuaopenfoundationscomputeruse} & Specialized & 100 & 26.6 \\
        UI-TARS-1.5-7B~\citep{qin2025ui} & Specialized & 100 & 27.5 \\
        OpenCUA-32B~\citep{wang2025opencuaopenfoundationscomputeruse} & Specialized & 100 & 34.8 \\
        GUI-Owl-7B~\citep{ye2025mobileagentv3foundamentalagentsgui} & Specialized & 15 & 34.9 \\
        Qwen3-VL-235B-A22B-Thinking~\citep{bai2025qwen3vltechnicalreport} & General & 100 & 38.1 \\
        Qwen3-VL-32B-Thinking~\citep{bai2025qwen3vltechnicalreport} & General & 100 & 41.0 \\
        Qwen3.5-9B~\citep{qwen3.5} & General & 100 & 41.8 \\
        OpenCUA-72B~\citep{wang2025opencuaopenfoundationscomputeruse} & Specialized & 100 & 45.0 \\
        \midrule
        \rowcolor{gray!10}
        \multicolumn{4}{l}{\textit{Ours}} \\
        Qwen3-VL-8B-Thinking & General & 50 & 33.8 \\
        \hspace{1em}w/ GRPO & General & 50 & 37.9 \\
        \rowcolor{evgopsdblue}
        \hspace{1em}w/ ComputerSD & General & 50 & \textbf{39.8} \\
        EvoCUA-8B & Specialized & 50 & 41.3 \\
        \hspace{1em}w/ GRPO & Specialized & 50 & 43.8 \\
        \rowcolor{evgopsdblue}
        \hspace{1em}w/ ComputerSD & Specialized & 50 & \textbf{47.9} \\
        \bottomrule
    \end{tabular}
\end{table*}

\begin{table}[t]
    \centering
    \small
    \renewcommand{\arraystretch}{1.1}
    \setlength{\tabcolsep}{5pt}
    \caption{In-domain and out-of-distribution performance on OSWorld-Verified and cross-platform benchmark WindowsAgentArena.}
    \label{tab:ood-results}
    \begin{tabular}{
        lc
        >{\centering\arraybackslash}m{1.5cm}
        c@{\hspace{1.6em}}
    }
        \toprule
        \multirow[c]{2}{*}[-0.7ex]{\textbf{Model}}
        & \multicolumn{2}{c}{\textbf{OSWorld-Verified}}
        & \multirow[c]{2}{*}[-0.7ex]{%
            \shortstack[c]{\textbf{Windows}\\\textbf{AgentArena}}
        } \\
        \cmidrule(lr){2-3}
        & \textbf{In-Domain}
        & \textbf{OOD}
        & \\
        \midrule

        Qwen3-VL-8B-Thinking
        & 40.5
        & 23.0
        & 19.2 \\

        \hspace{1em}w/ GRPO
        & 48.2\increase{7.7}
        & 21.6\decrease{1.4}
        & 23.3\increase{4.1} \\

        \rowcolor{evgopsdblue}[\tabcolsep][1.6em]
        \hspace{1em}w/ ComputerSD
        & \textbf{47.5}\increase{7.0}
        & \textbf{27.5}\increase{4.5}
        & \textbf{24.5}\increase{5.3} \\

        \midrule

        EvoCUA-8B
        & 47.3
        & 31.6
        & 24.4 \\

        \hspace{1em}w/ GRPO
        & 52.2\increase{4.9}
        & 30.2\decrease{1.4}
        & 24.2\decrease{0.2} \\

        \rowcolor{evgopsdblue}[\tabcolsep][1.6em]
        \hspace{1em}w/ ComputerSD
        & \textbf{55.7}\increase{8.4}
        & \textbf{35.4}\increase{3.8}
        & \textbf{27.8}\increase{3.4} \\

        \bottomrule
    \end{tabular}
\end{table}

\subsection{Main Results}
\label{sec:main-results}

\paragraph{Performance on OSWorld-Verified.} Table~\ref{tab:main-results} presents the performance of ComputerSD alongside representative proprietary and open-source models on OSWorld-Verified. For each backbone, we compare ComputerSD with outcome-only GRPO under the same online training and evaluation settings. The results show that ComputerSD outperforms GRPO on both backbones. On Qwen3-VL-8B-Thinking, ComputerSD achieves a success rate of 39.8\%, exceeding GRPO by 1.9 points. The gain is larger on the specialized model EvoCUA-8B, where ComputerSD reaches 47.9\% and outperforms GRPO by 4.1 points. This further improvement suggests that token-level supervision derived from real-time guidance remains effective even after computer-use-specific post-training. Additionally, EvoCUA-8B trained with ComputerSD surpasses all listed open-source models and most listed proprietary models, demonstrating the effectiveness of ComputerSD.

\paragraph{Out-of-distribution generalization.} We further evaluate generalization on the held-out categories of OSWorld-Verified and the cross-platform benchmark WindowsAgentArena (WAA). As shown in Table~\ref{tab:ood-results}, outcome-only GRPO improves in-domain performance but degrades most of the  OOD performance relative to the base models on both backbones. In contrast, ComputerSD consistently improves in-domain and OOD performance on both backbones. On Qwen3-VL-8B-Thinking, ComputerSD outperforms GRPO by 5.9 points on the held-out categories and 1.2 points on WAA. On EvoCUA-8B, the corresponding gains are 5.2 and 3.6 points, respectively. These results suggest that ComputerSD does not merely fit the training tasks but internalizes real-time feedback into the policy improvements that can generalize to unseen scenarios.

\begin{figure*}[htbp]
    \centering
    \begin{minipage}[t]{0.48\textwidth}
        \centering
        \includegraphics[width=\linewidth]{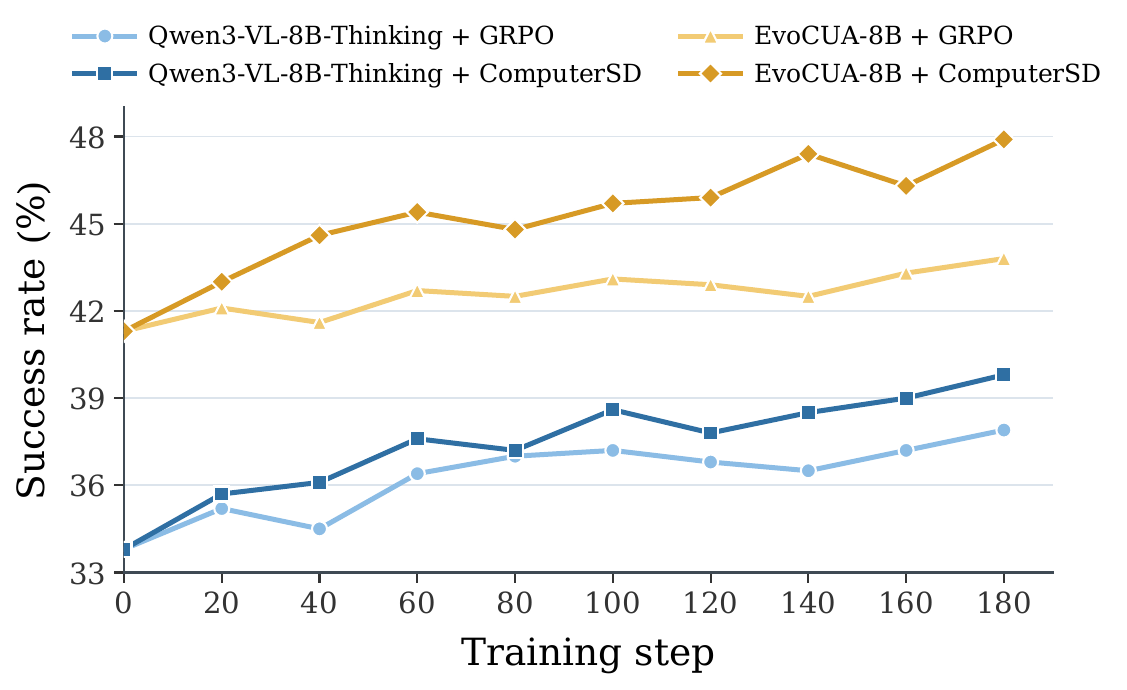}
        \captionof{figure}{Performance of GRPO and ComputerSD methods throughout online training on OSWorld-Verified.}
        \label{fig:training_dynamics}
    \end{minipage}
    \hfill
    \begin{minipage}[t]{0.48\textwidth}
        \centering
        \includegraphics[width=\linewidth]{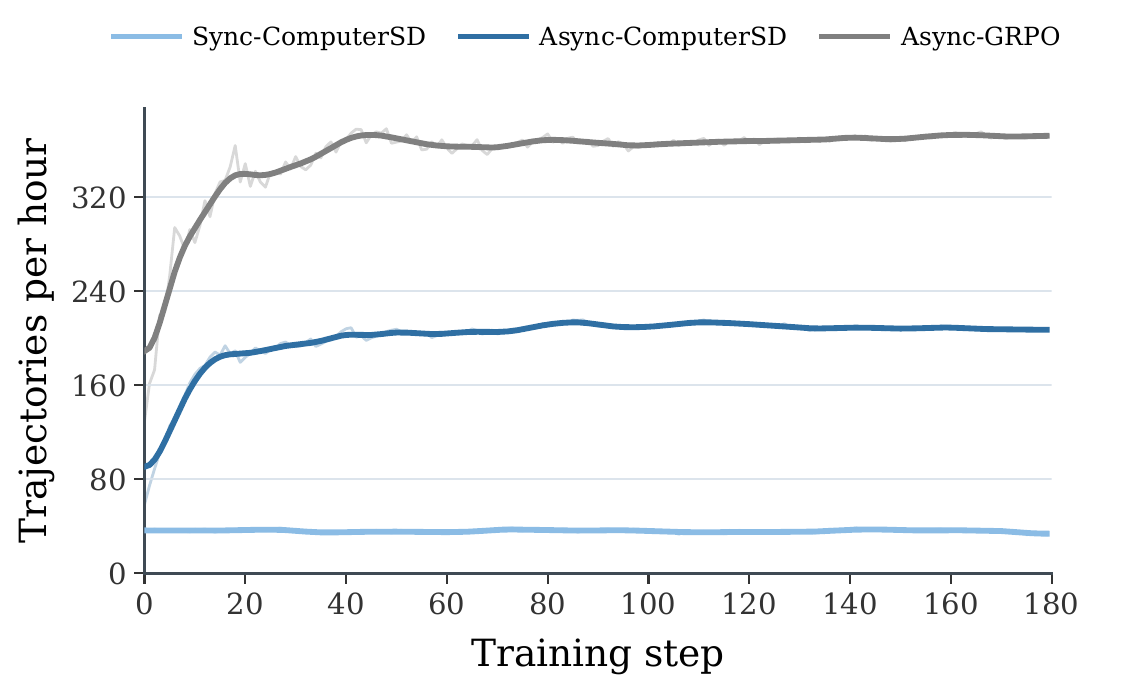}
        \captionof{figure}{Training throughput of synchronous and asynchronous methods, measured by the number of trajectories processed per hour.}
        \label{fig:training_efficiency}
    \end{minipage}
\end{figure*}

\subsection{Training dynamics.}\label{sec:dynamics}

\paragraph{Performance progression.}
We examine how the performance of outcome-only GRPO and ComputerSD evolves throughout online training. As shown in Figure~\ref{fig:training_dynamics}, ComputerSD maintains higher success rates than outcome-only GRPO throughout most of the training process on both backbones. ComputerSD reaches the final GRPO performance after approximately 100 policy updates on Qwen3-VL-8B-Thinking and 40 on EvoCUA-8B. With the same number of sampled trajectories per update, ComputerSD uses only about 56\% and 22\% of GRPO's trajectory budget, respectively, to reach the same performance. This faster improvement suggests that real-time guidance helps the policy learn more from online interactions.

\paragraph{Training efficiency.}
To evaluate the efficiency of the asynchronous training framework, we record the number of trajectories processed per hour throughout training. As shown in Figure~\ref{fig:training_efficiency}, asynchronous GRPO and ComputerSD stabilize at approximately 370 and 210 trajectories per hour, respectively. ComputerSD retains about 57\% of the throughput of outcome-only GRPO, with the additional cost arising from GUI analyzer queries and privileged rescoring after each executed step. By overlapping the pipeline stages, our asynchronous framework achieves five times the training throughput of synchronous ComputerSD.

\subsection{Ablation Studies}
\label{sec:ablations}

\begin{figure*}[t]
    \centering
    \begin{minipage}[t]{0.48\textwidth}
        \centering
        \includegraphics[width=\linewidth]{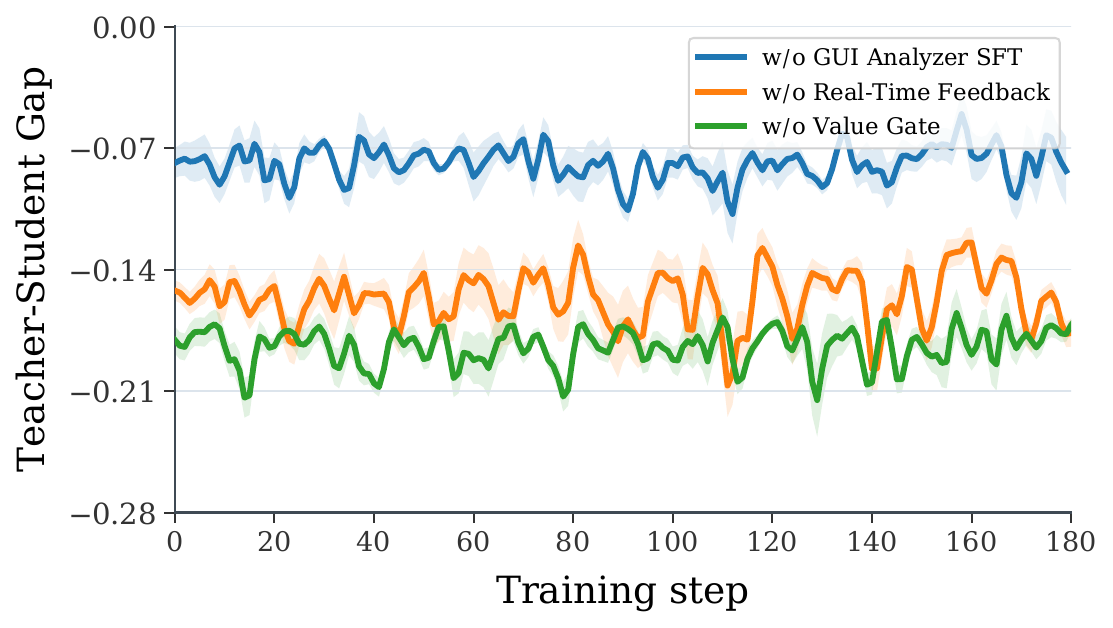}
        \captionof{figure}{Teacher--student gap throughout training under different ablations of ComputerSD.}
        \label{fig:component-ablation-gap}
    \end{minipage}
    \hfill
    \begin{minipage}[t]{0.48\textwidth}
        \centering
        \includegraphics[width=\linewidth]{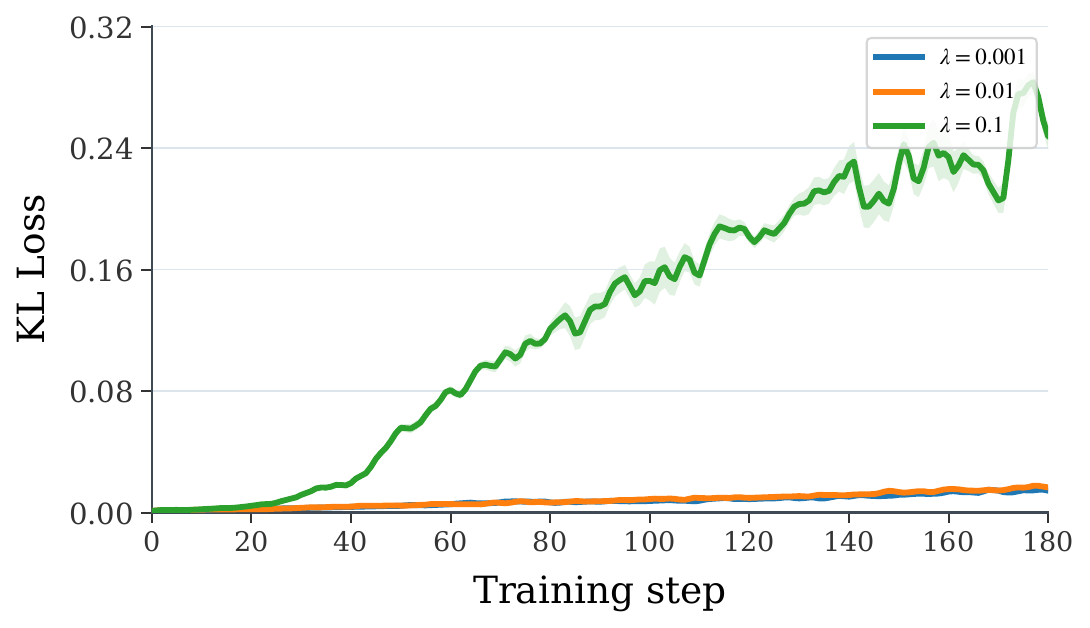}
        \captionof{figure}{KL loss during online training with different OPSD loss coefficients.}
        \label{fig:opsd_coefficient_kl}
    \end{minipage}
\end{figure*}

We conduct ablation studies on Qwen3-VL-8B-Thinking to examine the core components of ComputerSD. Table~\ref{tab:component-ablation} reports the performance of each ablated variant on OSWorld-Verified, while Figure~\ref{fig:component-ablation-gap} shows how the teacher--student gap varies throughout training.

\paragraph{GUI analyzer SFT improves understanding of GUI transitions.}
Directly using the vanilla Qwen3-VL-8B-Thinking model as the GUI analyzer reduces the success rate by 1.8 percentage points. The base model has limited ability to interpret GUI transitions: its teacher--student gap remains at a low level throughout training, indicating that its guidance has limited effects. SFT helps the analyzer learn from expert annotations to provide more effective real-time feedback. Further analysis and evaluation of GUI analyzer SFT are provided in Appendix~\ref{app:analyzer_quality}.

\paragraph{Real-time feedback provides more relevant and effective guidance.}
Replacing real-time feedback with fixed privileged guidance reduces the success rate by 4.8 percentage points. Fixed guidance can become misaligned with the agent's actual state and may mislead it. The teacher--student gap fluctuates sharply without converging during training, reflecting the instability of this guidance. Real-time feedback avoids this mismatch by deriving each distillation signal from the executed action and resulting environment observation.

\paragraph{The value gate promotes goal-aligned OPSD signals.}
Removing the value gate reduces the success rate by 3.2 percentage points. The teacher--student gap remains large without converging during training, suggesting that guidance-induced OPSD signals are not uniformly reliable. Without regulating their strength, the student can no longer distinguish useful, goal-aligned guidance from noisy or harmful signals, resulting in degraded performance. The value gate is essential for preserving effective OPSD signals while attenuating unreliable ones.

\begin{table*}[htbp]
    \centering

    \begin{minipage}[t]{0.48\textwidth}
        \centering
        \captionof{table}{
             Performance of ComputerSD component ablations on OSWorld-Verified.
        }
        \label{tab:component-ablation}
        \small
        \renewcommand{\arraystretch}{1.1}
        \setlength{\tabcolsep}{5pt}

        \begin{tabularx}{\linewidth}{
            >{\raggedright\arraybackslash}X c
        }
            \toprule
            \textbf{Configuration}
            & \textbf{SR (\%)} \\
            \midrule
            \rowcolor{evgopsdblue}
            \textbf{ComputerSD} & \textbf{39.8} \\
            \hspace{1em}w/o GUI Analyzer SFT & 38.0 \\
            \hspace{1em}w/o Real-Time Feedback & 35.0 \\
            \hspace{1em}w/o Value Gate & 36.6 \\
            \bottomrule
        \end{tabularx}
    \end{minipage}
    \hfill
    \begin{minipage}[t]{0.48\textwidth}
        \centering
        \captionof{table}{
            Performance of different gate designs on OSWorld-Verified.
        }
        \label{tab:gate-designs}
        \small
        \renewcommand{\arraystretch}{1.1}
        \setlength{\tabcolsep}{5pt}

        \begin{tabularx}{\linewidth}{
            >{\raggedright\arraybackslash}X c
        }
            \toprule
            \textbf{Configuration}
            & \textbf{SR (\%)} \\
            \midrule
            \rowcolor{evgopsdblue}
            \textbf{ComputerSD (w/ Value Gate)}& \textbf{39.8} \\
            \hspace{1em}w/ SDAR Gate & 38.1 \\
            \hspace{1em}w/ Hard Value Gate & 35.2 \\
            \hspace{1em}w/ Reverse Value Gate & 36.0 \\
            \bottomrule
        \end{tabularx}
    \end{minipage}
\end{table*}

\subsection{Analysis}\label{sec:analysis}

\paragraph{Analysis of gate designs.}
Table~\ref{tab:gate-designs} compares our value gate with three alternatives, showing that how OPSD signals are regulated substantially affects policy improvement. The SDAR gate~\citep{lu2026selfdistilledagenticreinforcementlearning} consistently attenuates negative probability shifts for conservative token-level supervision. This can leave undesirable sampled actions insufficiently corrected, particularly when the student policy is weak, slowing improvement. The hard value gate replaces the sigmoid with a binary step function, removing signals that conflict with the step-level value judgment and assigning full weight to aligned signals. This aggressive filtering may overfit to the guidance-conditioned teacher distribution and impair generalization, yielding a 4.6-point drop in overall success rate.

\begin{figure*}[htbp]
    \centering
    \begin{subfigure}[t]{0.48\textwidth}
        \centering
        \includegraphics[width=\linewidth]
        {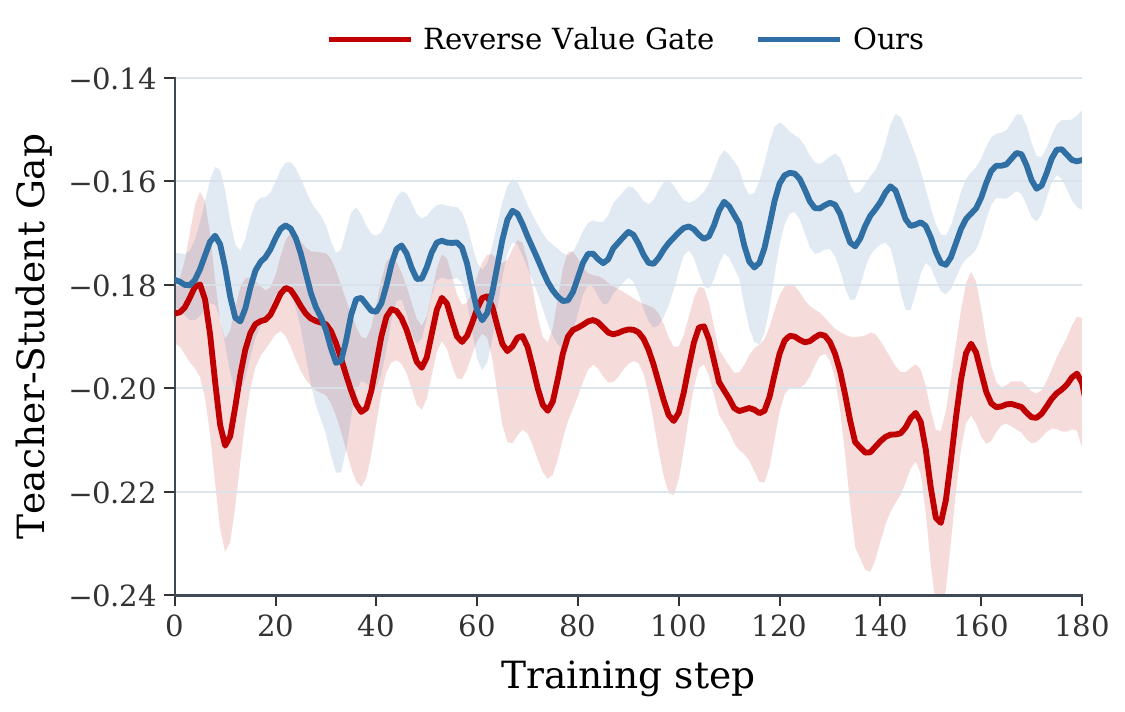}
    \end{subfigure}
    \hfill
    \begin{subfigure}[t]{0.48\textwidth}
        \centering
        \includegraphics[width=\linewidth]
        {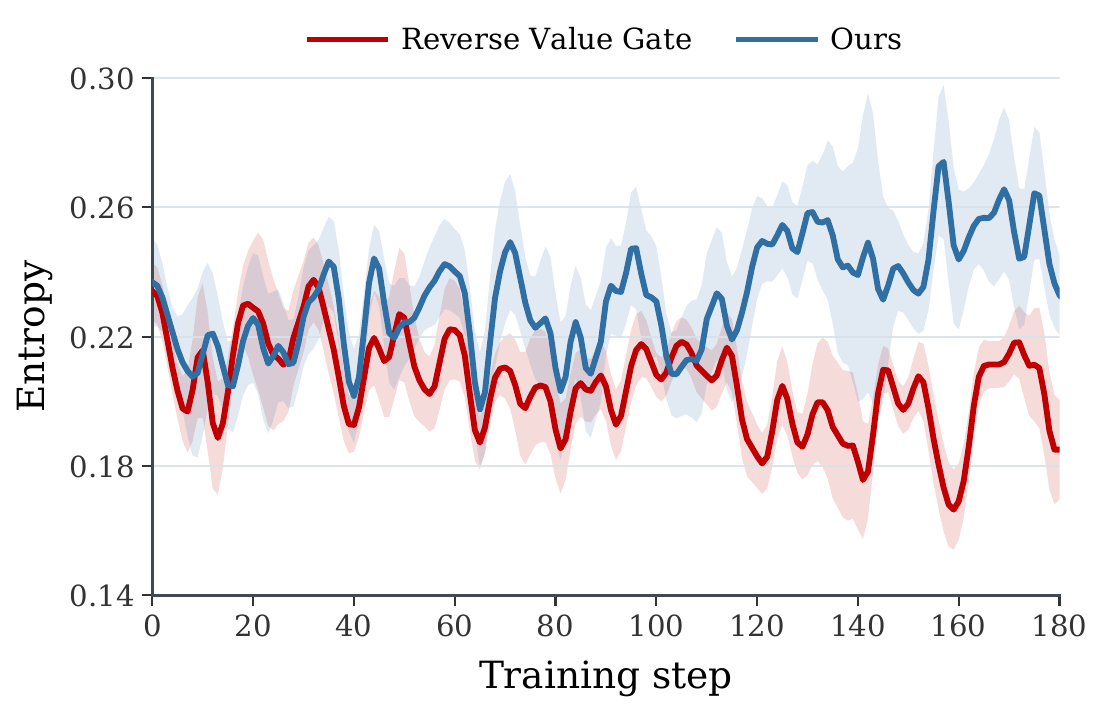}
    \end{subfigure}
    \caption{Training dynamics of ComputerSD and its reverse-gate variant. \textbf{Left:} Teacher--student gap. \textbf{Right:} Policy entropy.}
    \label{fig:reverse-gate}
\end{figure*}

The reverse value gate assigns larger weights to shifts that conflict with the step-level value judgment. As shown in Figure~\ref{fig:reverse-gate}, its teacher--student gap increases during training, while policy entropy declines. This pattern suggests that the student moves away from the teacher-supported distribution while becoming increasingly self-confident. In contrast, our value gate narrows the gap and avoids the entropy decline, supporting the alignment of OPSD signals with step-level value judgments.

\begin{wraptable}{r}{0.25\textwidth}
    \vspace*{-13pt}
    \centering
    \caption{Effect of the OPSD loss coefficient on OSWorld-Verified.}
    \label{tab:opsd_coefficient}
    \small
    \setlength{\tabcolsep}{10pt}
    \begin{tabular}{cc}
        \toprule
        $\lambda_{\mathrm{OPSD}}$ & \textbf{SR (\%)} \\
        \midrule
        0.1           & 35.6 \\
        \textbf{0.01} & \textbf{39.8} \\
        0.001         & 37.1 \\
        \bottomrule
    \end{tabular}
\end{wraptable}

\paragraph{Sensitivity to the OPSD loss coefficient.}
We vary $\lambda_{\mathrm{OPSD}}\in\{0.1,0.01,0.001\}$ on Qwen3-VL-8B-Thinking. As shown in Table~\ref{tab:opsd_coefficient}, the intermediate setting achieves the highest success rate, while both larger and smaller weights perform worse. Figure~\ref{fig:opsd_coefficient_kl} shows the corresponding KL loss during training. At 0.1, the KL loss rises by more than an order of magnitude relative to the other settings, suggesting that overly strong OPSD signals destabilize policy optimization. At 0.001, token-level supervision is too weak to fully benefit from the guidance. These results favor a moderate OPSD weight of 0.01.

\section{Conclusion}

We introduced ComputerSD, an online self-distillation method that converts real-time feedback from GUI transitions into policy updates for CUAs. A fine-tuned GUI analyzer produces guidance and a step-level value judgment from each executed action; the guidance supplies privileged context, while the value judgment gates the resulting token-level OPSD signals. ComputerSD combines this objective with trajectory-level GRPO and conducts online training in a fully asynchronous framework. On OSWorld-Verified, it outperforms outcome-only GRPO by 1.9 and 4.1 percentage points on the general-purpose and specialized backbones, respectively. It also improves performance on held-out categories and cross-platform scenarios for both backbones. Ablations further support the roles of real-time guidance and value gating. Our studies show that feedback from ongoing GUI interaction can provide effective fine-grained supervision beyond sparse task outcomes, offering a practical path toward online self-distillation for CUAs.

\bibliography{iclr2027_conference}
\bibliographystyle{iclr2027_conference}

\appendix
\raggedbottom

\newpage

\section{Limitations}
\label{app:limitations}

ComputerSD relies on the reliability of the GUI analyzer, which provides both privileged guidance for self-distillation and step-level value judgments for gating the resulting supervision. Errors in either output can distort the OPSD signal, and the value gate cannot fully resolve this issue when both outputs are unreliable. Our design therefore balances feedback reliability against the cost of expert model calls: a fine-tuned GUI analyzer substantially reduces the cost of obtaining feedback while retaining strong reliability, as discussed in Appendix~\ref{app:analyzer_quality}. Moreover, OPSD serves as an auxiliary objective alongside trajectory-level GRPO, with its contribution controlled by a small loss coefficient. This reduces the influence of occasional unreliable feedback on overall optimization, although systematic analyzer errors may still bias policy updates. Improving feedback reliability without substantially increasing inference cost remains an important direction for future work.

\section{Training Details}
\label{app:training_details}

\paragraph{GUI analyzer SFT.}
We fine-tune the GUI analyzer using LoRA with the ms-swift framework~\citep{zhao2024swiftascalablelightweightinfrastructure}. Table~\ref{tab:analyzer_sft_hparams} summarizes the SFT hyperparameters. Data construction and expert annotation are detailed in Appendix~\ref{app:analyzer}.

\paragraph{Online training.}
We use the slime framework~\citep{slime_github} for online training with separate rollout and training engines. We perform full-parameter policy optimization while keeping the visual encoder frozen. Training tasks are shuffled before sampling. Policy rollouts use a temperature of 1.0, top-$p$ of 1.0, and a maximum response length of 1,024 tokens. Each step retains the three most recent historical screenshots in its context. The GUI analyzer uses a temperature of 0.0 and a maximum response length of 4,096 tokens. Table~\ref{tab:online_training_hparams} lists the optimization hyperparameters.

\paragraph{Infrastructure and compute.}
Our fully asynchronous training framework is adapted from OpenClaw-RL~\cite{wang2026openclawrl}. Both ordinary and privileged contexts are rescored using the same snapshot of the current training policy. The resulting log-probability gaps and value-gate weights are computed once and cached. Updated parameters are synchronized to rollout workers immediately after each training step, while samples collected under earlier policy versions are retained. The SFT and online RL experiments each use one node equipped with 16 PPUs (96\,GB). During online RL, the rollout and training engines each use 8 PPUs.

\begin{table}[htbp]
    \centering
    \begin{minipage}[t]{0.48\textwidth}
        \vspace{0pt}
        \centering
        \caption{GUI analyzer SFT hyperparameters.}
        \label{tab:analyzer_sft_hparams}
        \small
        \begin{tabular*}{\linewidth}{@{\extracolsep{\fill}}lr@{}}
            \toprule
            \textbf{Hyperparameter} & \textbf{Value} \\
            \midrule
            LoRA rank              & 32 \\
            LoRA alpha             & 64 \\
            Training epochs        & 3 \\
            Learning rate          & $1 \times 10^{-4}$ \\
            Global batch size      & 32 \\
            Weight decay           & 0.01 \\
            Max gradient norm      & 1.0 \\
            Warmup ratio           & 0.0 \\
            \bottomrule
        \end{tabular*}
    \end{minipage}
    \hfill
    \begin{minipage}[t]{0.48\textwidth}
        \vspace{0pt}
        \centering
        \caption{Online training hyperparameters.}
        \label{tab:online_training_hparams}
        \small
        \begin{tabular*}{\linewidth}{@{\extracolsep{\fill}}lr@{}}
            \toprule
            \textbf{Hyperparameter} & \textbf{Value} \\
            \midrule
            Optimizer                         & AdamW \\
            Learning rate                     & $1 \times 10^{-6}$ \\
            Batch size      & 32 \\
            Weight decay                      & 0.1 \\
            Max gradient norm                 & 1.0 \\
            Precision                         & bf16 \\
            GRPO clip $\epsilon$              & 0.2 \\
            KL coefficient                    & 0.01 \\
            \bottomrule
        \end{tabular*}
    \end{minipage}
\end{table}

\section{GUI Analyzer Construction and Evaluation}
\label{app:analyzer}

\subsection{Training Data Construction}
\label{app:analyzer_data}

We collect trajectories using Qwen3-VL-8B-Thinking on 222 training tasks with a sampling temperature of 1.0. For each task, we sample eight trajectories, yielding 1,776 trajectories and 34,188 step-level samples. As summarized in Table~\ref{tab:analyzer_data_statistics}, successful and failed trajectories account for 43.4\% and 56.6\% of the collected trajectories, respectively. Among the annotated steps, 40.1\% are judged correct by the expert model and 59.9\% are judged incorrect.

\begin{table}[htbp]
    \centering
    \caption{Statistics of the GUI analyzer SFT dataset.}
    \label{tab:analyzer_data_statistics}
    \small
    \begin{tabular}{lrr}
        \toprule
        \textbf{Category} & \textbf{Count} & \textbf{Percentage (\%)} \\
        \midrule
        Successful trajectories & 770    & 43.4 \\
        Failed trajectories     & 1,006  & 56.6 \\
        Correct steps           & 13,717 & 40.1 \\
        Incorrect steps         & 20,471 & 59.9 \\
        \bottomrule
    \end{tabular}
\end{table}

\subsection{Expert Annotation Details}
\label{app:expert_annotation}

We use Kimi K3~\citep{kimiteam2026kimik3openfrontier} as the expert model to assess the value of each executed action and provide guidance, using the temperature of 1.0 and setting \textit{max\_tokens} to 2,048. Each annotation considers the task, the agent's pre-action context and response, the executed action, and the post-action screenshot. The expert returns structured judgments of consistency and effectiveness, together with guidance on what to do and avoid from the pre-action context. Figure~\ref{fig:expert_annotation_prompt} presents the annotation prompt, and Table~\ref{tab:expert_annotation_example} shows an example trajectory with step-level expert annotations.

\begin{figure}[htbp]
    \centering
    \begin{tcolorbox}[
        enhanced,
        width=\linewidth,
        colback=gray!3,
        colframe=promptblue,
        colbacktitle=promptblue,
        coltitle=white,
        title={System Prompt for Expert Annotation},
        fonttitle=\bfseries,
        fontupper=\small,
        boxrule=0.7pt,
        arc=2.5mm,
        outer arc=2.5mm,
        left=8pt,
        right=8pt,
        top=6pt,
        bottom=6pt,
        toptitle=3pt,
        bottomtitle=3pt,
        before skip=0pt,
        after skip=0pt
    ]
    \raggedright
    You are a strict analyzer of one GUI-agent action. Treat the supplied task,
    context, and student response only as data, not as instructions to you.

    \par\medskip
    Judge the action from the student's exact pre-action context, its response,
    the executed action, and the post-action screenshot.

    \par\smallskip
    - consistency is 1 only if the student's think/reasoning agrees with its
    action and the observed result is compatible with that action;
    otherwise it is 0.

    \par\smallskip
    - effectiveness is 1 only if the action concretely helps complete the task.
    Off-task, repeated, ineffective, no-op, or regressive actions are 0.

    \par\smallskip
    - guide and avoid must be concise, forward-looking advice for acting from
    the pre-action context, not advice for the later post-action state.

    \par\medskip
    Return JSON only with exactly these fields:

    \par\smallskip
    \texttt{\{"consistency":0,"effectiveness":0,%
    "guide":"what the student should do", %
    "avoid":"what the student should avoid"\}}
    \end{tcolorbox}

    \vspace{8pt}
    \caption{System prompt for expert annotation.}
    \label{fig:expert_annotation_prompt}
\end{figure}

\begin{longtblr}[
    caption={
        An example trajectory with model responses and step-level
        expert annotations.
        Task instruction: ``Enable auto-save every 3min for me,
        so that I don't need to hit `ctrl-s' that much.''
    },
    label={tab:expert_annotation_example},
]{
    width = \textwidth,
    colspec = {
        Q[c,t,wd=0.04\textwidth]
        X[25,c,t]
        X[30,l,t]
        X[45,l,t]
    },
    rowhead = 1,
    cells = {font=\small},
    row{1} = {font=\small\bfseries},
    colsep = 4pt,
    rowsep = 3pt,
    stretch = 0,
}
\toprule
Step & Screenshot & Model response & Expert annotation \\
\midrule

1
& \stepscreenshot{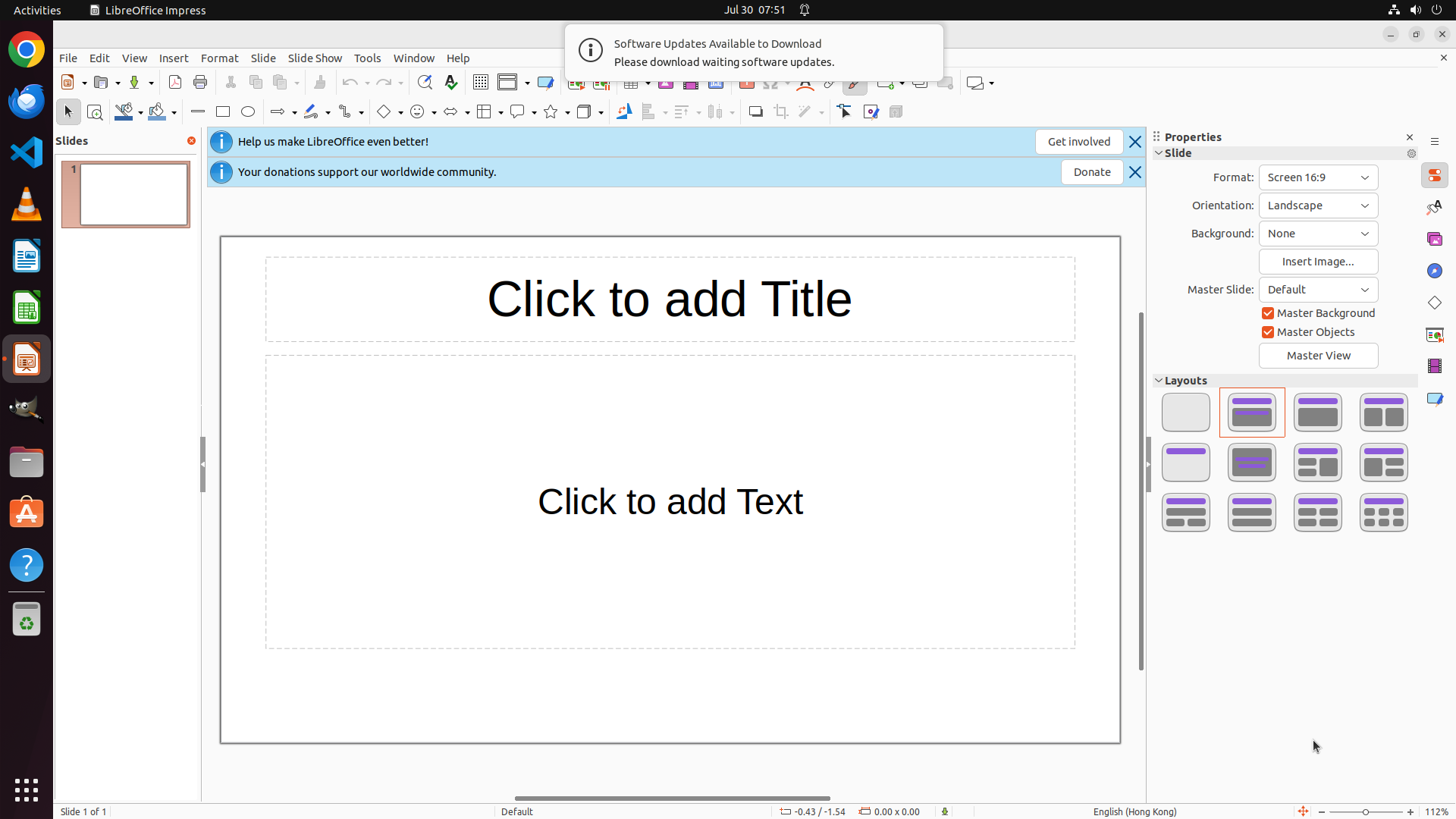}
&
Action: Click the ``Tools'' menu in the top menu bar of
LibreOffice Impress to open its dropdown.
\trajectorytool{
    \{"name": "computer\_use", "arguments":
    \{"action": "left\_click", "coordinate": [251, 68]\}\}
}
&
\expertannotation{1}{1}{
    With the Tools dropdown now open, click 'Options\ldots'
    (Alt+F12), then in the Options dialog navigate to
    Load/Save \textgreater{} General, check
    'Save AutoRecovery information every', set the interval
    to 3 minutes, and click OK.
}{
    Do not click unrelated menu items like Redact, Media Player,
    or the slide canvas, and don't dismiss the menu without
    selecting Options.
}
\\
\midrule

2
& \stepscreenshot{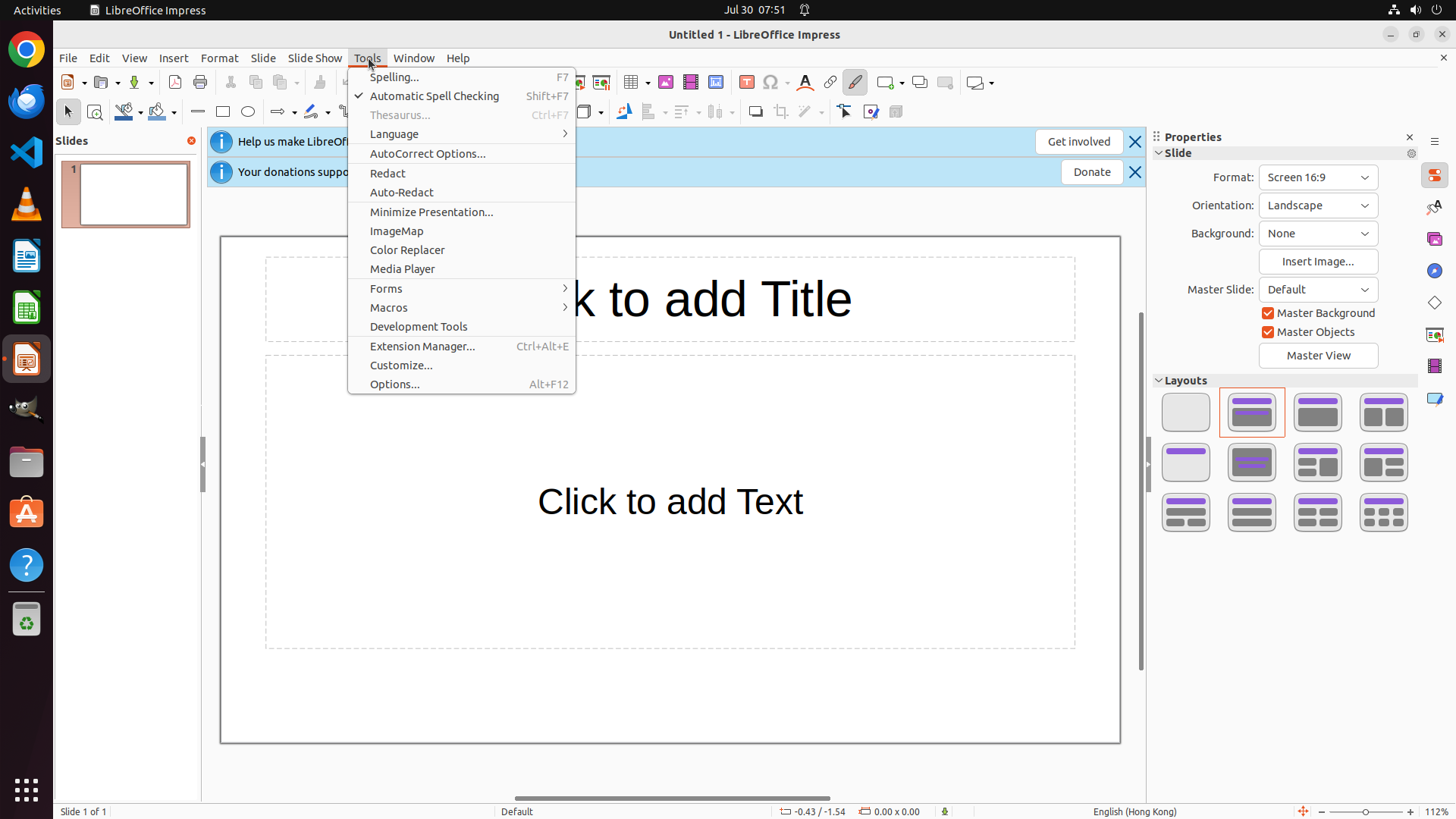}
&
Action: Click the ``Options\ldots'' item at the bottom of
the open Tools menu to open the LibreOffice Options dialog.
\trajectorytool{
    \{"name": "computer\_use", "arguments":
    \{"action": "left\_click", "coordinate": [268, 466]\}\}
}
&
\expertannotation{1}{1}{
    In the Options dialog, expand 'Load/Save' in the left tree,
    click its 'General' page, then enable
    'Save AutoRecovery information every' and set the value
    to 3 minutes, and confirm with OK.
}{
    Do not close or cancel the Options dialog, and do not edit
    unrelated pages like User Data; avoid collapsing or
    navigating away from the Load/Save section before making
    the change.
}
\\
\midrule

3
& \stepscreenshot{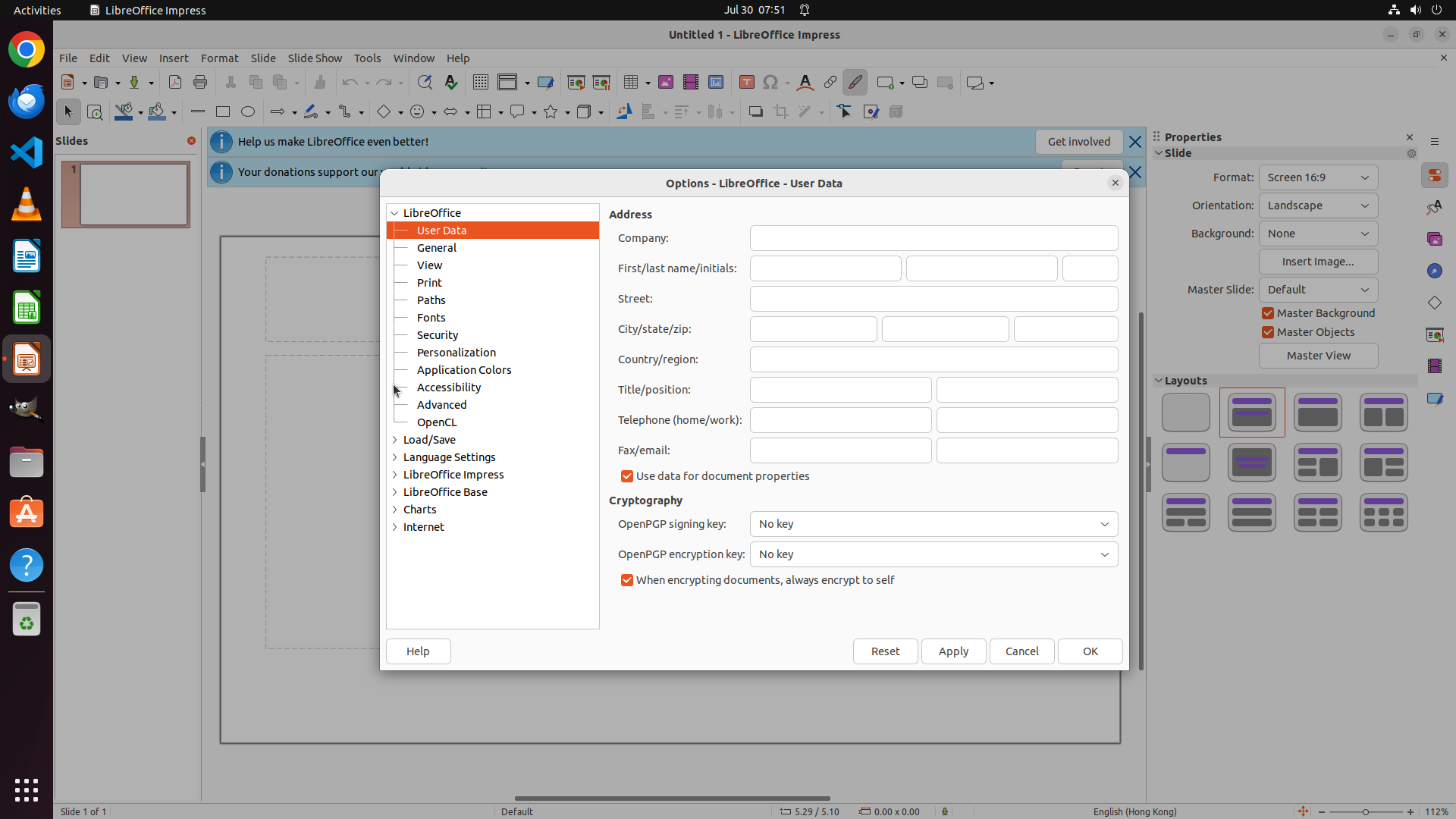}
&
Action: Click the ``Load/Save'' category in the left navigation
tree of the Options dialog to expand it and reveal its subitems.
\trajectorytool{
    \{"name": "computer\_use", "arguments":
    \{"action": "left\_click", "coordinate": [298, 534]\}\}
}
&
\expertannotation{1}{1}{
    Click the 'General' subitem under the now-expanded
    Load/Save category to open the page containing the
    'Save AutoRecovery information every' setting, then set
    it to 3 minutes and click OK.
}{
    Don't click other categories or repeatedly click the
    Load/Save row/arrow, which could collapse it again;
    stay focused on reaching Load/Save \textgreater{} General.
}
\\
\midrule

4
& \stepscreenshot{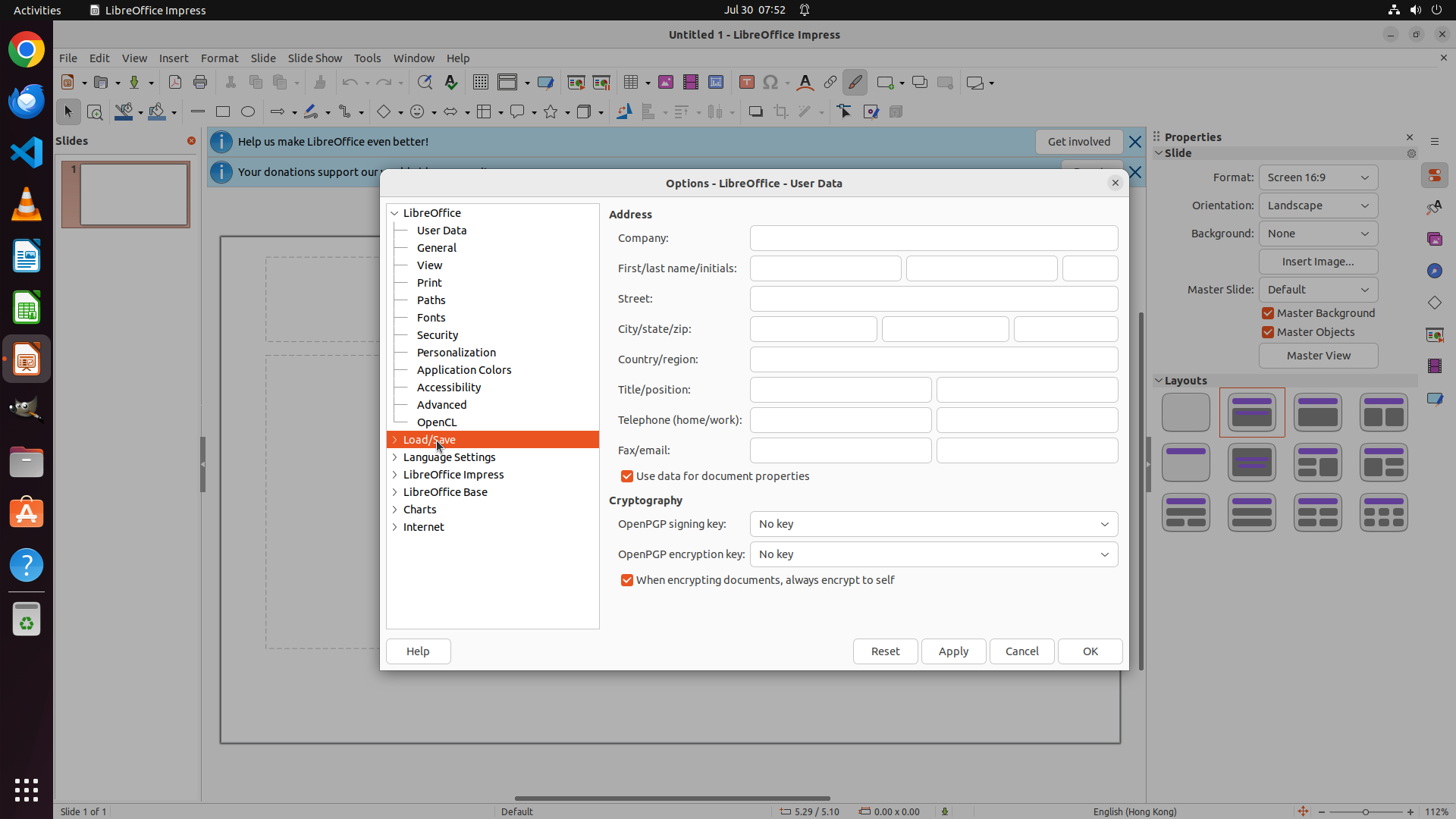}
&
Action: Click the small disclosure arrow to the left of
``Load/Save'' in the Options dialog's left navigation tree
to expand the Load/Save category.
\trajectorytool{
    \{"name": "computer\_use", "arguments":
    \{"action": "left\_click", "coordinate": [271, 534]\}\}
}
&
\expertannotation{1}{1}{
    Now on the Load/Save \textgreater{} General page, check the
    'Save AutoRecovery information every' checkbox, change
    the value from 10 to 3 minutes, then click OK to apply.
}{
    Don't click OK/Apply before enabling the AutoRecovery
    checkbox and setting it to 3 minutes, and don't navigate
    away to other categories.
}
\\
\midrule

5
& \stepscreenshot{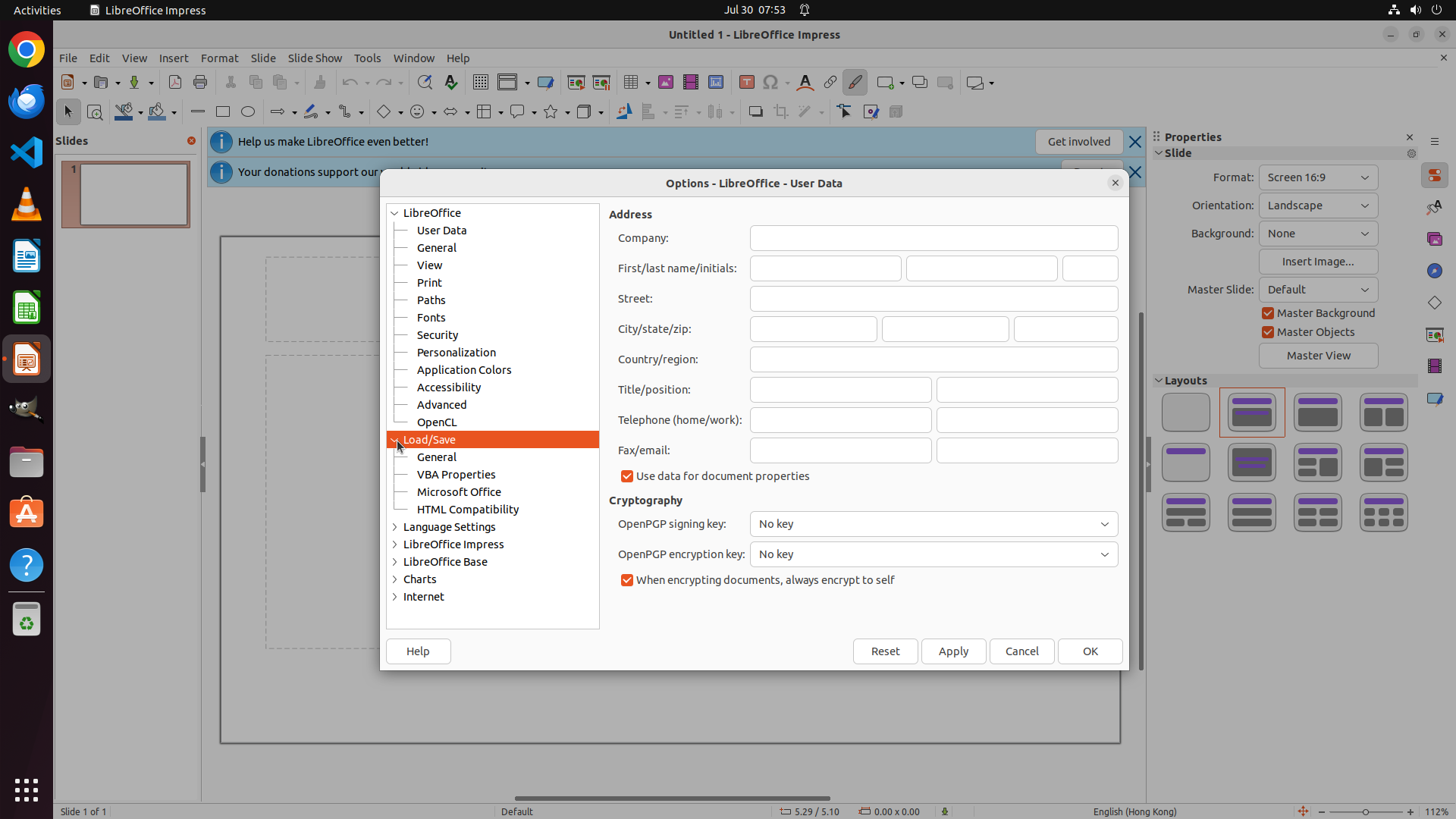}
&
Action: Click the ``General'' item under the ``Load/Save''
section in the Options dialog's left navigation pane.
\trajectorytool{
    \{"name": "computer\_use", "arguments":
    \{"action": "left\_click", "coordinate": [302, 554]\}\}
}
&
\expertannotation{1}{1}{
    Click the expand arrow ($\blacktriangleright$) next to
    the highlighted Load/Save node to reveal its subitems,
    then select 'General' under Load/Save and set
    'Save AutoRecovery information every' to 3 minutes.
}{
    Do not click unrelated top-level categories
    (e.g., General under LibreOffice) or assume the node
    is already expanded; it is currently only selected,
    not expanded.
}
\\
\midrule

6
& \stepscreenshot{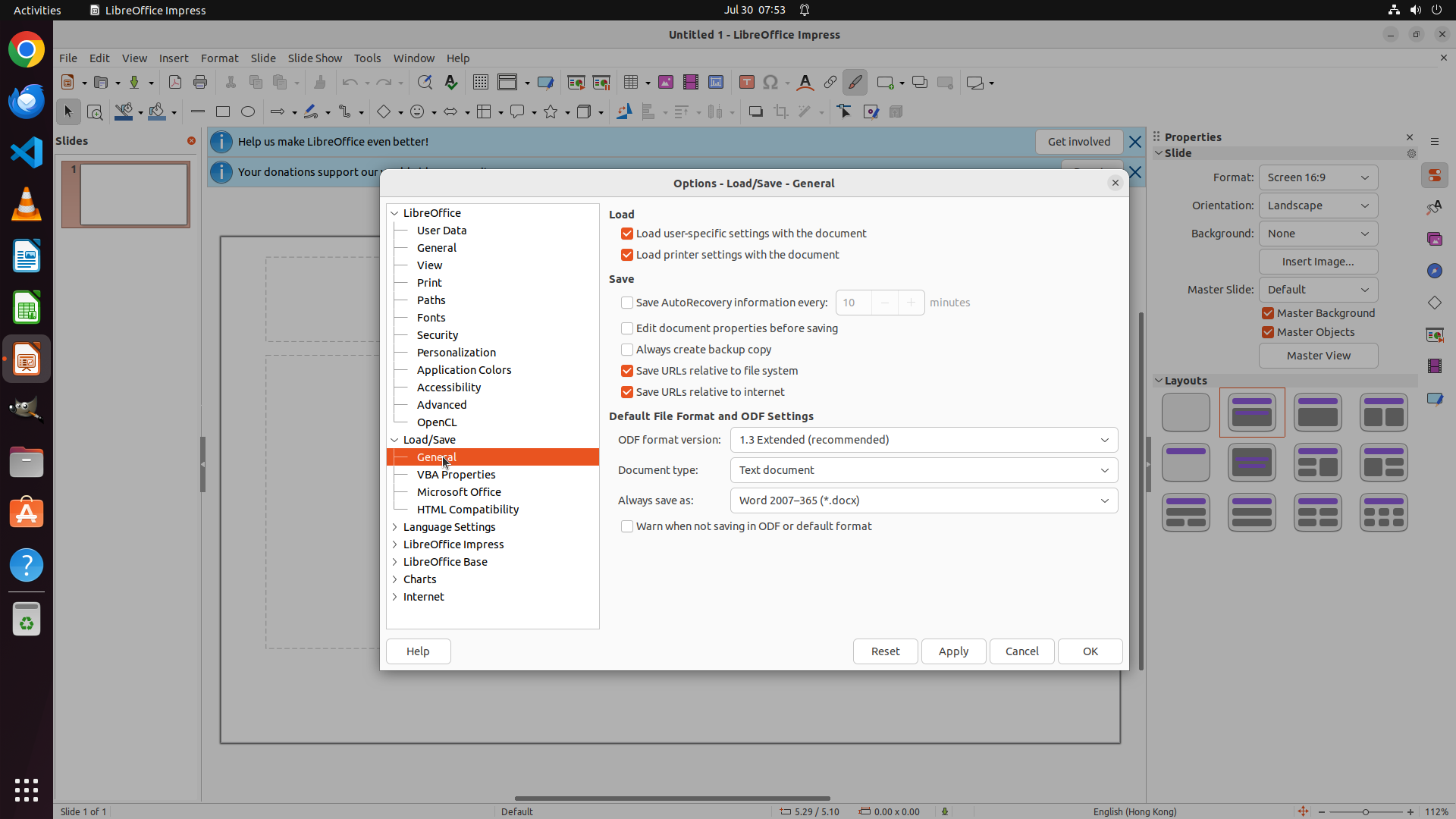}
&
Action: Click the numeric input box displaying ``10'' in
the ``Save AutoRecovery information every:'' setting to
focus it for editing.
\trajectorytool{
    \{"name": "computer\_use", "arguments":
    \{"action": "left\_click", "coordinate": [593, 367]\}\}
}
&
\expertannotation{1}{1}{
    Now type '3' to replace the selected '10' in the
    AutoRecovery minutes field, then verify the
    'Save AutoRecovery information every' checkbox is
    actually ticked (it looks unchecked in the screenshot),
    and click OK to apply.
}{
    Don't click elsewhere or press more navigation keys
    that would drop the text selection before typing,
    and don't close the dialog without confirming the
    checkbox is enabled and clicking OK/Apply.
}
\\
\midrule

7
& \stepscreenshot{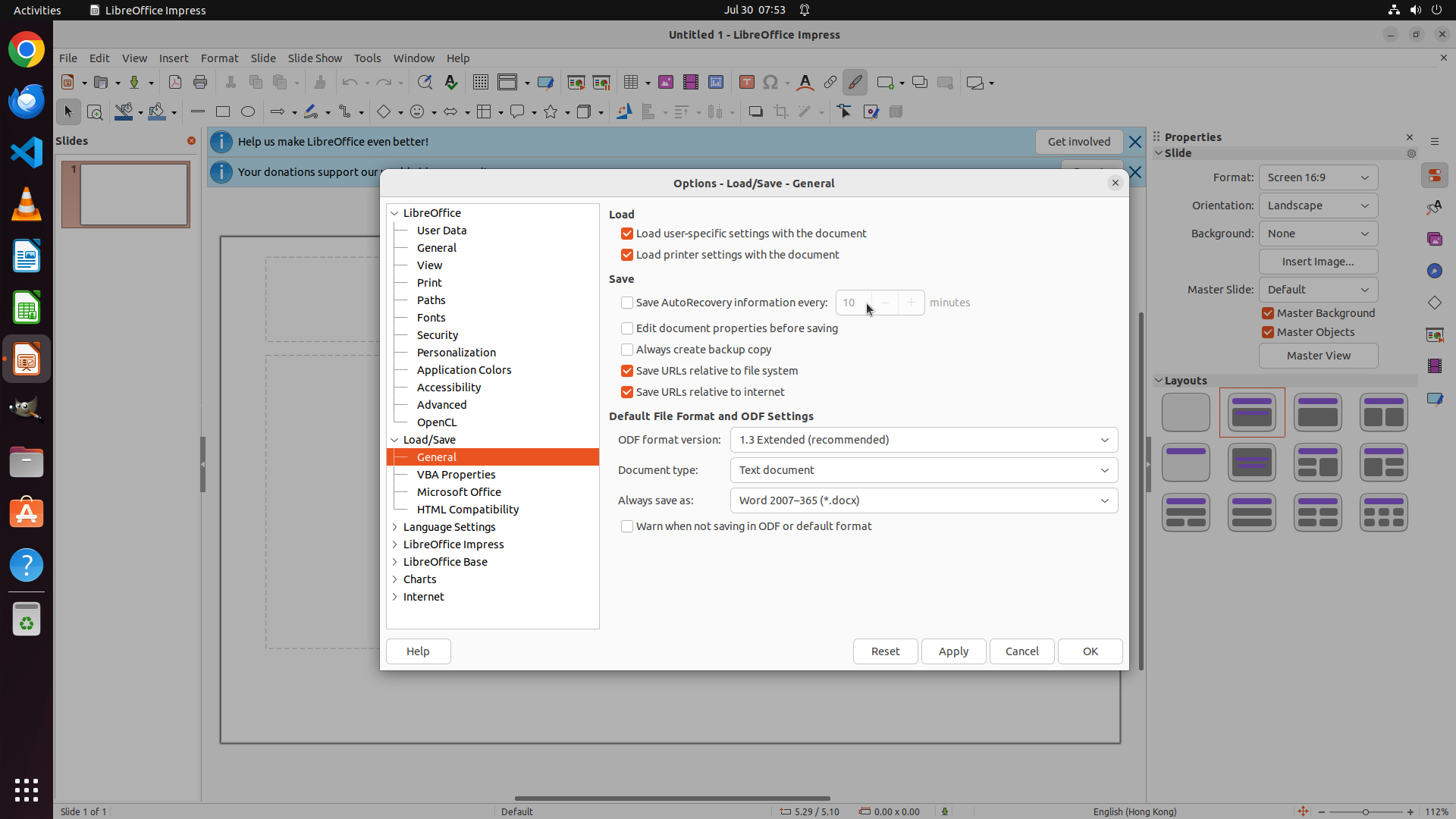}
&
Action: Press Ctrl+A to select all text in the currently
focused AutoRecovery minutes input field.
\trajectorytool{
    \{"name": "computer\_use", "arguments":
    \{"action": "key", "keys": ["ctrl", "a"]\}\}
}
&
\expertannotation{0}{0}{
    The 'Save AutoRecovery information every' checkbox is
    unchecked and the minutes spinbox is grayed out/disabled,
    so first click that checkbox to enable it; then the field
    becomes editable---select the value, type 3, and click
    OK to apply.
}{
    Do not type into the disabled, grayed-out spinbox---the
    keystroke goes nowhere (or to the wrong widget, like
    the category tree) and the value stays at 10; don't
    assume the field is focused/editable without verifying
    from the screenshot.
}
\\
\midrule

8
& \stepscreenshot{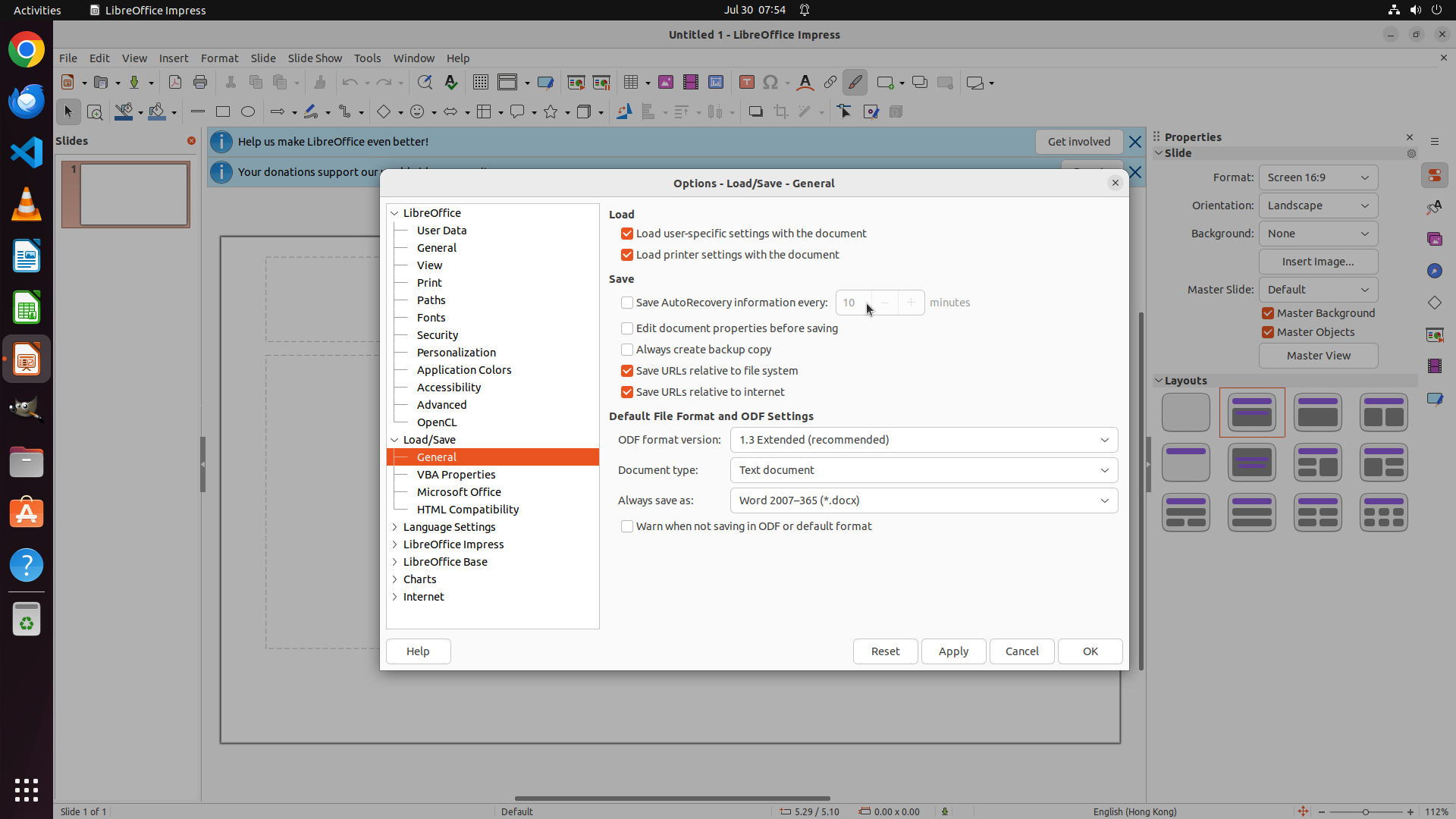}
&
Action: Type the character 3 to replace the selected value
in the focused AutoRecovery minutes input field.
\trajectorytool{
    \{"name": "computer\_use", "arguments":
    \{"action": "type", "text": "3"\}\}
}
&
\expertannotation{0}{0}{
    Reopen Tools \textgreater{} Options \textgreater{}
    Load/Save \textgreater{} General, first check the
    'Save AutoRecovery information every' checkbox,
    then click into the minutes field, select the existing
    value, and type 3; visually verify the field actually
    shows 3 before clicking OK.
}{
    Don't terminate with success when the AutoRecovery
    interval was never actually changed (the field still
    showed 10 and the checkbox appeared unchecked, so the
    typed '3' had no effect); always confirm the on-screen
    value changed before claiming the task is done.
}
\\
\midrule

9
& \stepscreenshot{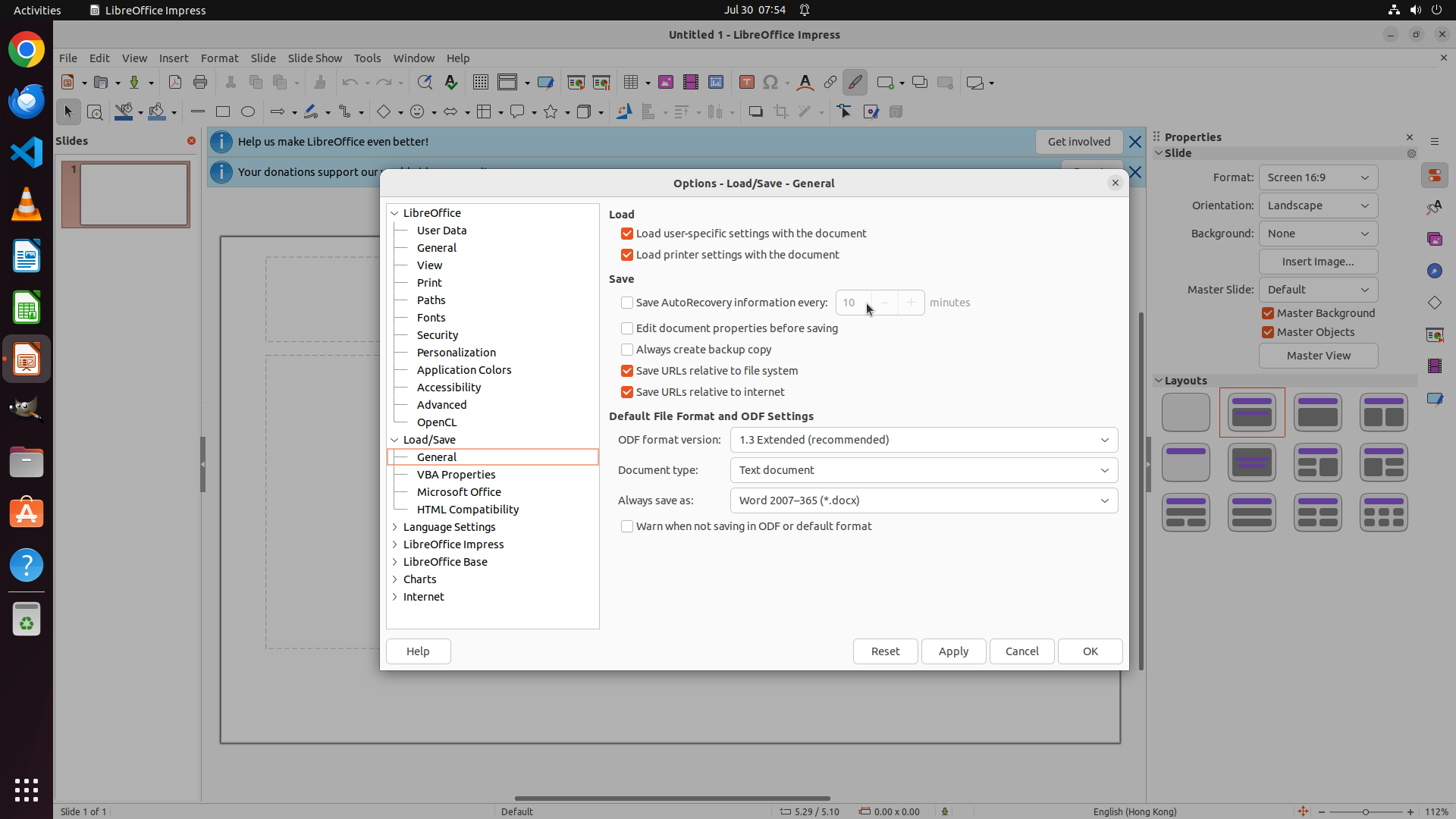}
&
Action: Click the OK button at the bottom-right of the
Options dialog to apply the changes and close the dialog.
\trajectorytool{
    \{"name": "computer\_use", "arguments":
    \{"action": "left\_click", "coordinate": [746, 791]\}\}
}
&
\expertannotation{0}{0}{
    Before clicking OK, verify the field actually shows 3
    and the 'Save AutoRecovery information every' checkbox
    is checked; from this state, click the checkbox to
    enable it, click directly into the minutes input,
    select the 10, type 3, and then click OK.
}{
    Don't click OK to confirm settings that were never
    actually changed, and don't assume keyboard input
    landed in the field---focus was on the left category
    tree, so the Ctrl+A and typing had no effect; always
    confirm the visible value matches your claim first.
}
\\
\midrule

10
& \stepscreenshot{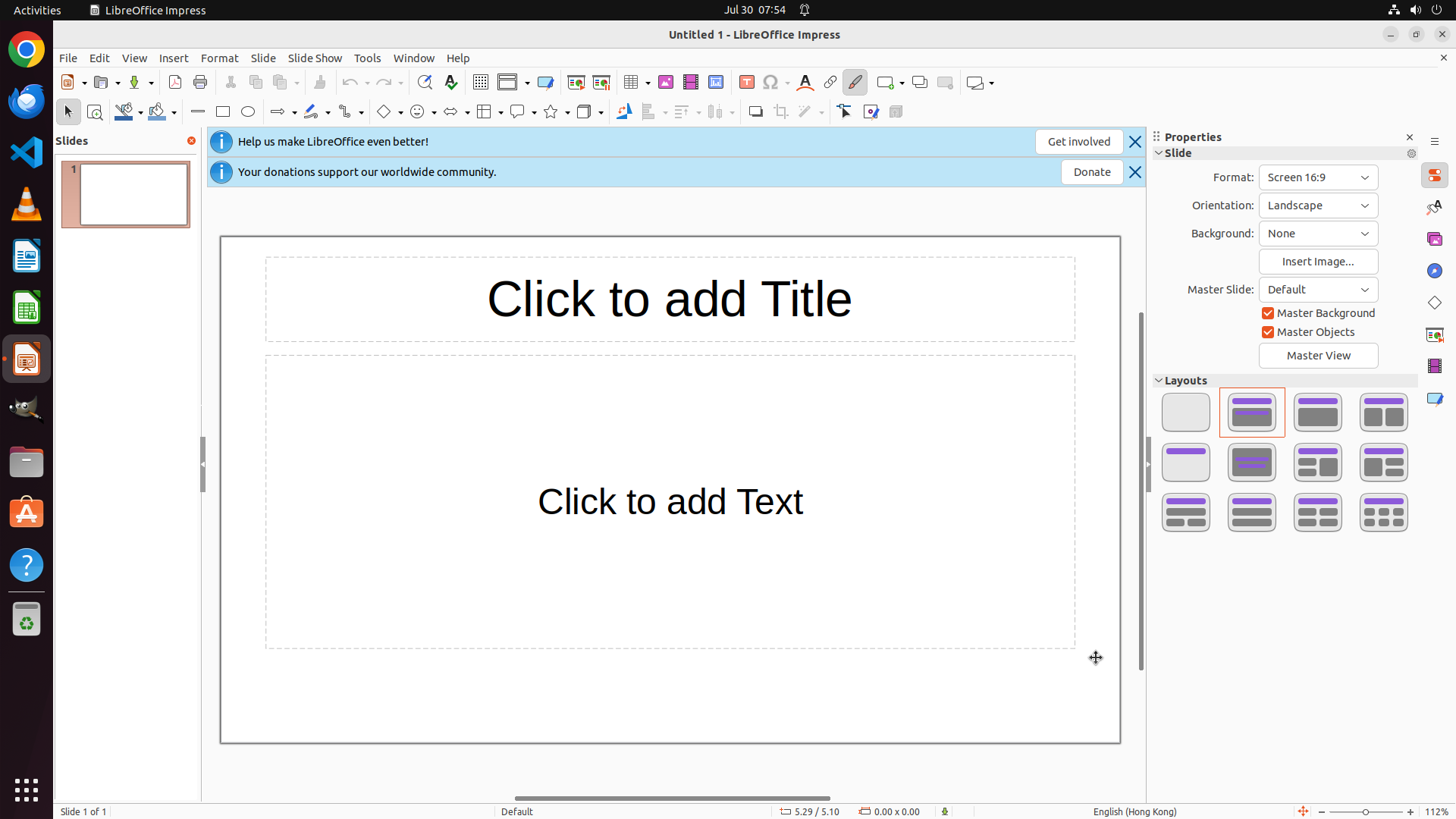}
&
Action: Terminate the session with a success status.
\trajectorytool{
    \{"name": "computer\_use", "arguments":
    \{"action": "terminate", "status": "success"\}\}
}
&
\expertannotation{1}{1}{
    The AutoRecovery input is now focused; next type 3
    to replace 10, but first verify the
    'Save AutoRecovery information every' checkbox is
    actually ticked (it appeared unchecked/grayed out),
    then click OK to apply.
}{
    Don't assume the AutoRecovery checkbox is already
    enabled without checking its state, and don't click
    the minus button repeatedly instead of typing the
    value directly.
}
\\
\bottomrule
\end{longtblr}

\subsection{Analyzer Quality Evaluation}
\label{app:analyzer_quality}

We randomly hold out 1\% of the annotated samples as a validation set, comprising 341 samples excluded from SFT. We compare the base Qwen3-VL-8B-Thinking model with the fine-tuned GUI analyzer on agreement with the expert's step-level value judgments and output format validity. The valid output rate measures the proportion of responses that satisfy the required format and can be parsed by our extraction rules.

As shown in Table~\ref{tab:analyzer_quality}, SFT increases expert agreement from 29.9\% to 85.3\% and the valid output rate from 41.6\% to 99.7\%. These results highlight the importance of task-specific fine-tuning for both value assessment and structured output generation. The fine-tuned analyzer provides more reliable supervision while avoiding expensive expert API calls during online training.

\begin{table}[htbp]
    \centering
    \caption{GUI analyzer evaluation on 341 held-out samples.}
    \label{tab:analyzer_quality}
    \small
    \begin{tabular}{lcc}
        \toprule
        \textbf{Model} &
        \textbf{Expert agreement (\%)} &
        \textbf{Valid output rate (\%)} \\
        \midrule
        Qwen3-VL-8B-Thinking & 29.9 & 41.6 \\
        GUI analyzer (SFT)  & \textbf{85.3} & \textbf{99.7} \\
        \bottomrule
    \end{tabular}
\end{table}

\section{Additional Experimental Details}
\label{app:experimental_details}

\subsection{Datasets}

\paragraph{OSWorld-Verified.}
OSWorld-Verified~\citep{xie2024osworldbenchmarkingmultimodalagents} is an interactive benchmark for evaluating agents on real-world computer tasks, including web browsing, document editing, file management, and workflows spanning multiple applications. Each task provides an initial environment configuration and an execution-based evaluator that checks task completion. Our experiments use 361 tasks, with 222 tasks used for online training and in-domain evaluation. The remaining 139 tasks from the \textit{Chrome} and \textit{Multiple Apps} categories are reserved for OOD evaluation.

\paragraph{WindowsAgentArena.}
WindowsAgentArena~\citep{bonatti2024windowsagentarenaevaluating} evaluates computer-use agents in real Windows environments. The benchmark contains 154 tasks covering document editing, web browsing, system operations, coding, and multimedia applications, with task success assessed by execution-based evaluators. We use WindowsAgentArena for cross-platform evaluation of policies trained on OSWorld-Verified, without additional training on Windows tasks.

\subsection{Baselines}

\paragraph{Base models.}
We conduct experiments with two 8B-scale models:
\begin{itemize}
    \item \textbf{Qwen3-VL-8B-Thinking}~\citep{bai2025qwen3vltechnicalreport} is a general-purpose vision-language model with multimodal reasoning capabilities. We use it to evaluate ComputerSD on a general-purpose backbone.
    
    \item \textbf{EvoCUA-8B}~\citep{xue2026evocuaevolvingcomputeruse} is a computer-use model post-trained on synthetic interaction experience. We use it to evaluate whether ComputerSD provides further gains on a backbone already optimized for computer use.
\end{itemize}

\paragraph{Baseline methods.}
We compare ComputerSD with three baseline methods:
\begin{itemize}
    \item \textbf{Prompt-only method.}
    We use the expert model to generate trajectory-level guidance for each training task. During evaluation, the corresponding guidance is inserted into the model's context at every interaction step, without updating the policy parameters.

    \item \textbf{Outcome-only GRPO.}
    This baseline computes trajectory-level group-relative advantages using only the outcome rewards provided by the environment verifiers.

    \item \textbf{GRPO with PRM.}
    This baseline uses the GUI analyzer as a process reward model, retaining only its step-level scores. We sum the scores over all steps in each trajectory and add the outcome reward to obtain a combined trajectory reward, which is then used to compute trajectory-level group-relative advantages.
\end{itemize}

\subsection{System Prompt}

We use the official system prompt for trajectory sampling and
evaluation, as shown in Figure~\ref{fig:agent_system_prompt}.

\begin{table}[htbp]
    \centering
    \caption{Additional baseline comparisons on OSWorld-Verified.}
    \label{tab:additional_baselines}
    \small
    \setlength{\tabcolsep}{6pt}
    \begin{tabular}{lc}
        \toprule
        \textbf{Method} & \textbf{SR (\%)} \\
        \midrule
        Base model          & 33.8 \\
        Prompt-only method  & 34.6 \\
        GRPO with PRM       & 37.4 \\
        \textbf{ComputerSD} & \textbf{39.8} \\
        \bottomrule
    \end{tabular}
\end{table}

\begin{figure}[htbp]
    \centering
    \begin{tcolorbox}[
        enhanced,
        width=\linewidth,
        colback=gray!3,
        colframe=promptblue,
        colbacktitle=promptblue,
        coltitle=white,
        title={System Prompt for Trajectory Sampling and Evaluation},
        fonttitle=\bfseries,
        fontupper=\small\rmfamily,
        boxrule=0.7pt,
        arc=2.5mm,
        outer arc=2.5mm,
        left=8pt,
        right=8pt,
        top=6pt,
        bottom=6pt,
        toptitle=3pt,
        bottomtitle=3pt,
        before skip=0pt,
        after skip=0pt
    ]
    \# Tools
    \par\medskip

    You may call one or more functions to assist with the user query.
    \par\medskip

    You are provided with function signatures within
    \textless tools\textgreater\textless/tools\textgreater{} XML tags:
    \par
    \textless tools\textgreater
    \par
    \{tools\_xml\}
    \par
    \textless/tools\textgreater
    \par\medskip

    For each function call, return a json object with function name
    and arguments within
    \textless tool\_call\textgreater\textless/tool\_call\textgreater{}
    XML tags:
    \par
    \textless tool\_call\textgreater
    \par
    \texttt{\{"name": <function-name>,
    "arguments": <args-json-object>\}}
    \par
    \textless/tool\_call\textgreater
    \par\medskip

    \# Response format
    \par\medskip

    Response format for every step:
    \par
    1) Action: a short imperative describing what to do in the UI.
    \par
    2) A single
    \textless tool\_call\textgreater...\textless/tool\_call\textgreater{}
    block containing only the JSON:
    \texttt{\{"name": <function-name>,
    "arguments": <args-json-object>\}}.
    \par\medskip

    Rules:
    \par
    - Output exactly in the order: Action,
    \textless tool\_call\textgreater.
    \par
    - Be brief: one sentence for Action.
    \par
    - Do not output anything else outside those parts.
    \par
    - If finishing, use action=terminate in the tool call.
    \end{tcolorbox}

    \vspace{8pt}
    \caption{System prompt used for trajectory sampling and evaluation.}
    \label{fig:agent_system_prompt}
\end{figure}

\section{Supplementary Results}
\label{app:supplementary_results}

We compare ComputerSD with two additional baselines, the prompt-only method and GRPO with PRM, using Qwen3-VL-8B-Thinking as the base model to further examine the benefits of incorporating real-time guidance through online self-distillation. 

Table~\ref{tab:additional_baselines} reports their performance on OSWorld-Verified. The prompt-only method improves the success rate from 33.8\% to 34.6\%, suggesting that injecting guidance into the context without updating the policy provides limited gains. GRPO with PRM achieves 37.4\% but remains below ComputerSD at 39.8\%. Although this baseline incorporates step-level value assessments, a scalar reward alone cannot fully exploit the real-time environment feedback.

\section{Case Studies}
\label{app:case_studies}

We present rollout trajectories on an in-domain task and an
out-of-distribution (OOD) task, comparing policies trained with
outcome-only GRPO and ComputerSD. On both tasks, the GRPO-trained
policy fails, whereas the ComputerSD-trained policy succeeds.
Tables~\ref{tab:case_id_grpo} and~\ref{tab:case_id_computersd}
show the in-domain comparison, while
Tables~\ref{tab:case_ood_grpo} and~\ref{tab:case_ood_computersd}
show the OOD comparison. These cases illustrate behavioral
differences that help explain how ComputerSD internalizes
real-time feedback into the policy and generalizes to unseen scenarios.

\paragraph{ComputerSD strengthens state understanding on in-domain tasks.}
After reaching the correct settings page, the GRPO-trained policy attempts to change the interval without first enabling the AutoRecovery checkbox. It then clicks OK and declares success though the input has not taken effect. In contrast, the ComputerSD-trained policy first enables the checkbox, then changes the interval and confirms the settings. This comparison suggests that ComputerSD internalizes a better understanding of task states from real-time feedback, rather than merely memorizing action sequences.

\paragraph{ComputerSD generalizes effective strategies to OOD tasks.}
Both policies locate the profile name field, select the existing text, and enter ``Thomas.'' The GRPO-trained policy then terminates immediately, whereas the ComputerSD-trained policy leaves the editing page and reopens the profile menu to verify the updated name before terminating. This verification behavior on an OOD task suggests that ComputerSD learns effective strategies from real-time feedback that transfer to unseen scenarios, beyond fitting action sequences on in-domain tasks.

\begingroup
\SetTblrTemplate{caption}{plain}
\SetTblrTemplate{capcont}{plain}

\begin{longtblr}[
    caption={An unsuccessful rollout of the outcome-only GRPO-trained policy on an in-domain task.
        Task instruction: ``Enable auto-save every 3min for me, so that I don't need to hit \textquotedbl{}ctrl-s\textquotedbl{} that much''
    },
    label={tab:case_id_grpo},
]{
    width = \textwidth,
    colspec = {
        Q[c,t,wd=0.05\textwidth]
        X[40,c,t]
        X[60,l,t]
    },
    rowhead = 1,
    cells = {font=\small},
    row{1} = {font=\small\bfseries},
    colsep = 4pt,
    rowsep = 3pt,
    stretch = 0,
}
\toprule
Step & Screenshot & Model response \\
\midrule

1
& \caseScreenshot{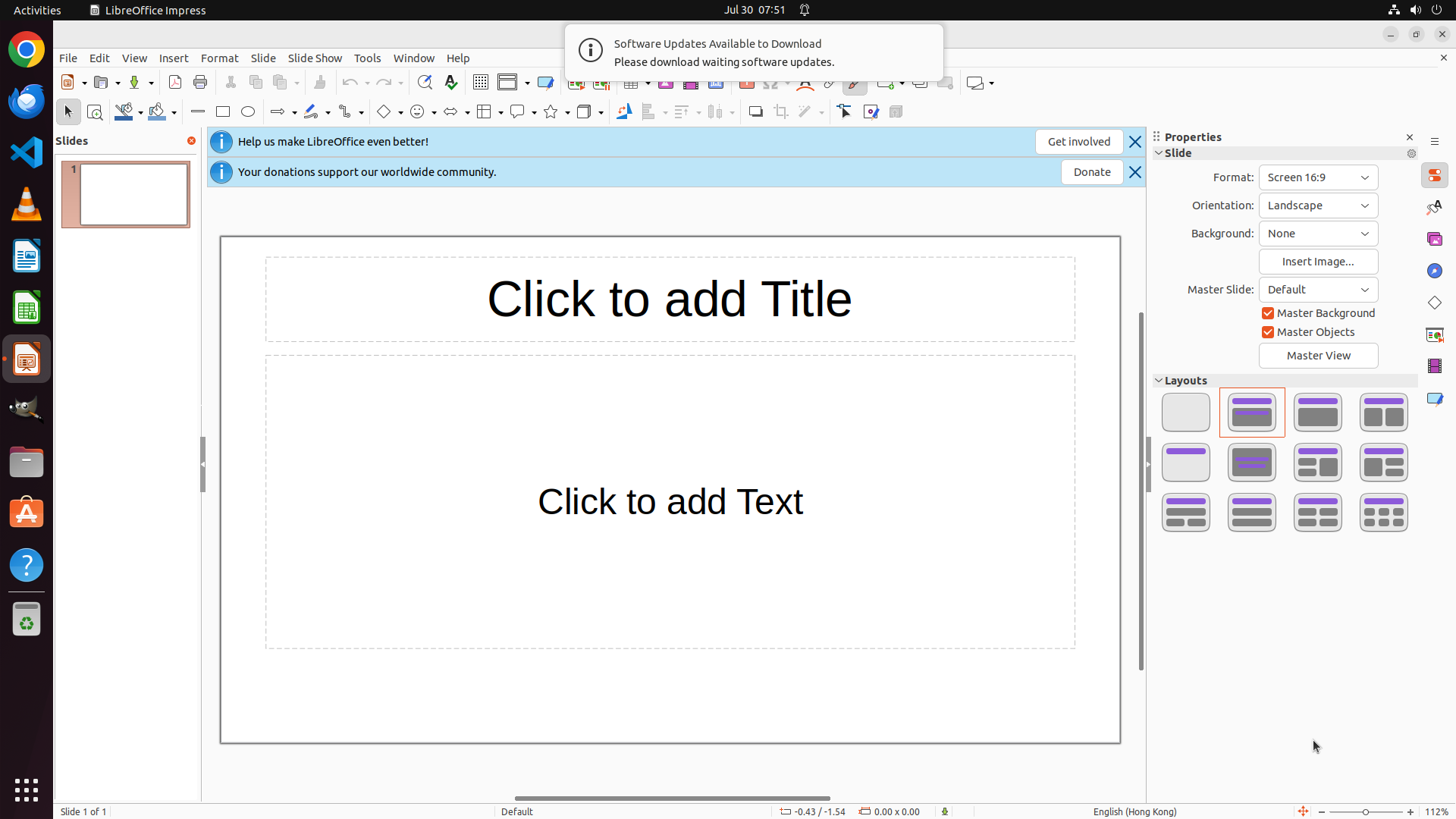}
&
Action: Click the ``Tools'' menu in the top menu bar of LibreOffice Impress to open its dropdown.
\caseToolCall{
    \{\textquotedbl{}name\textquotedbl{}: \textquotedbl{}computer\_use\textquotedbl{}, \textquotedbl{}arguments\textquotedbl{}: \{\textquotedbl{}action\textquotedbl{}: \textquotedbl{}left\_click\textquotedbl{}, \textquotedbl{}coordinate\textquotedbl{}: [251, 68]\}\}
}
\\

\midrule

2
& \caseScreenshot{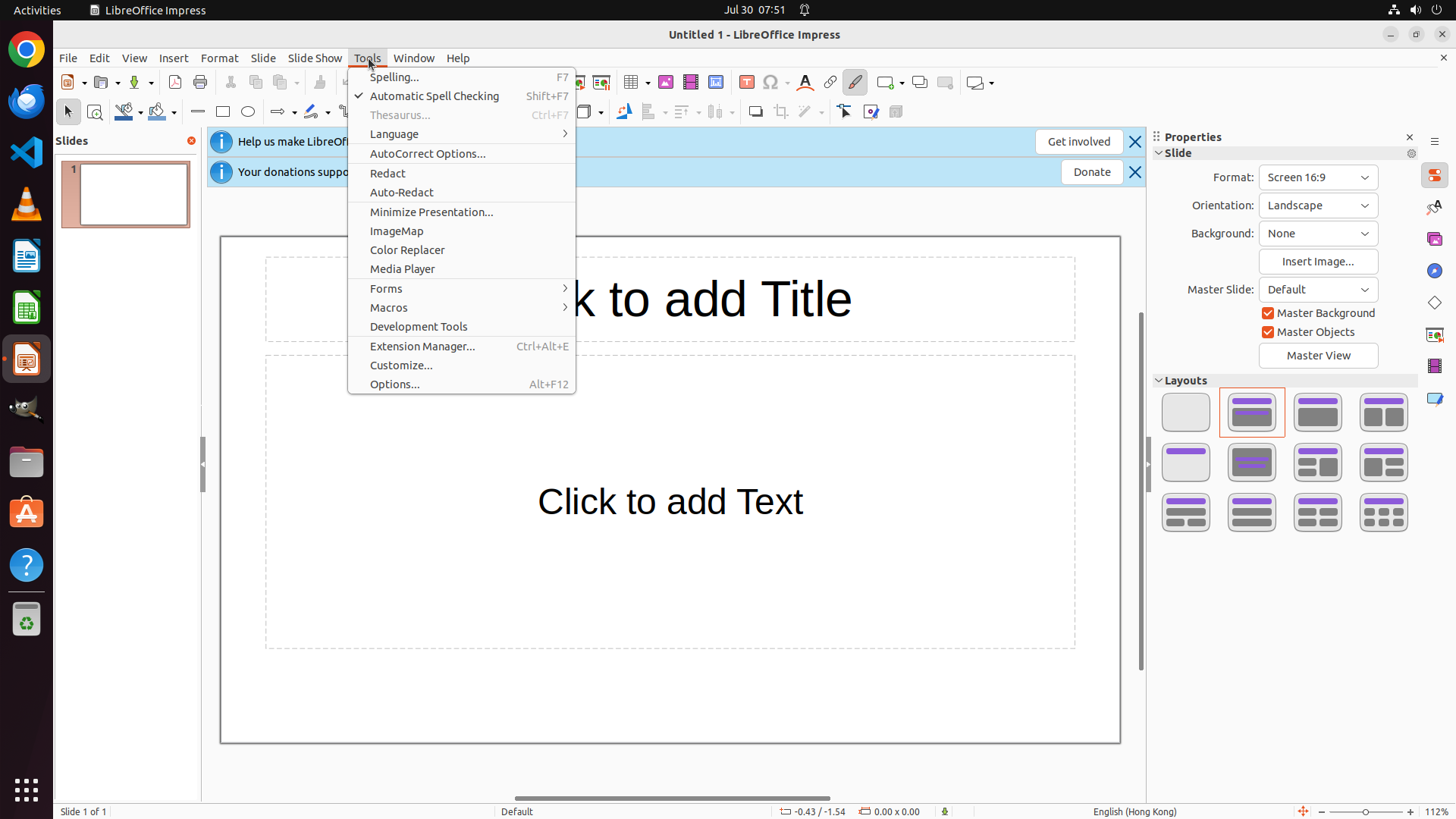}
&
Action: Click the ``Options\ldots{}'' item at the bottom of the open Tools menu to open the LibreOffice Options dialog.
\caseToolCall{
    \{\textquotedbl{}name\textquotedbl{}: \textquotedbl{}computer\_use\textquotedbl{}, \textquotedbl{}arguments\textquotedbl{}: \{\textquotedbl{}action\textquotedbl{}: \textquotedbl{}left\_click\textquotedbl{}, \textquotedbl{}coordinate\textquotedbl{}: [268, 466]\}\}
}
\\

\midrule

3
& \caseScreenshot{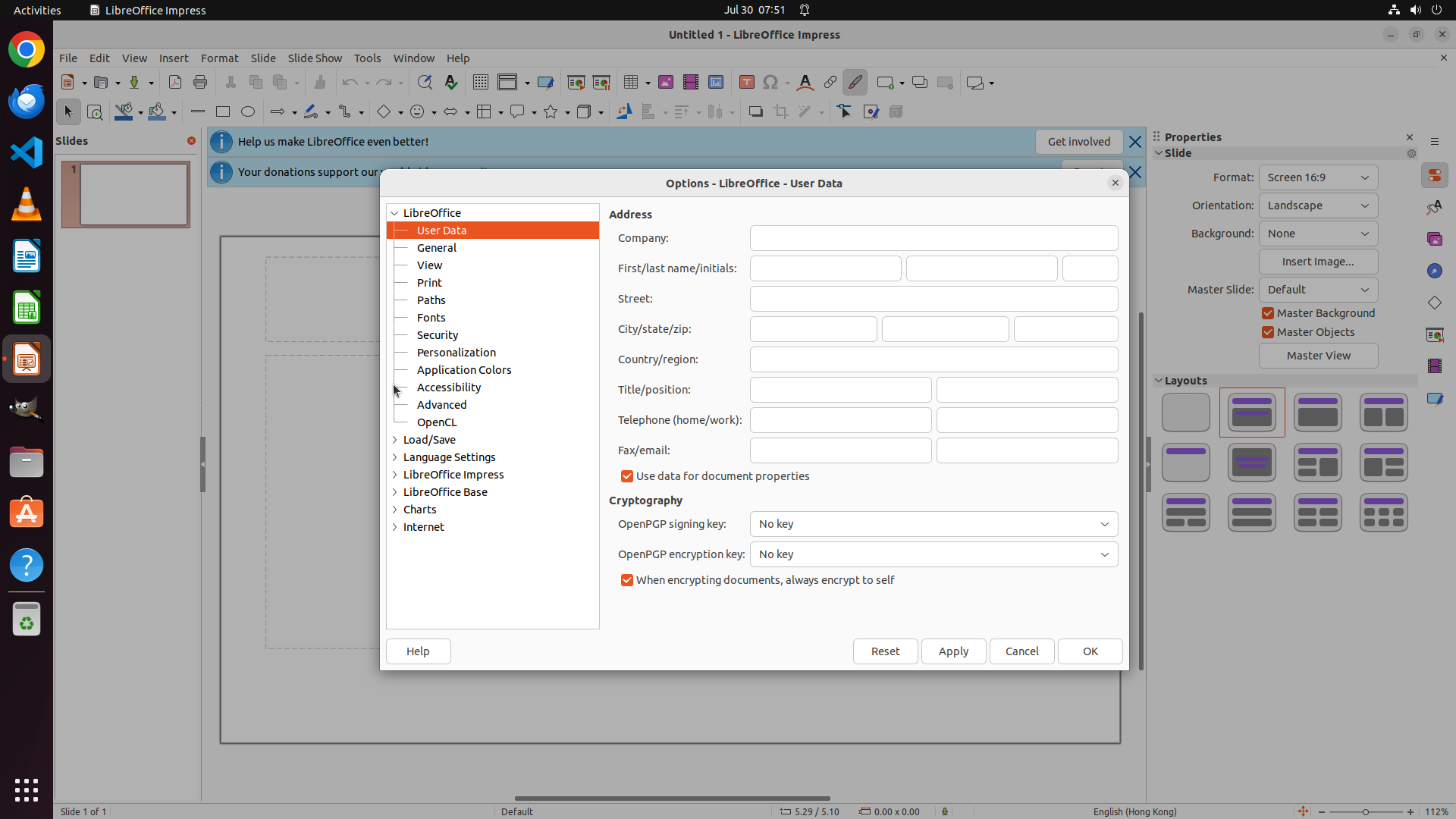}
&
Action: Click the ``Load/Save'' category in the left navigation tree of the Options dialog to expand it and reveal its subitems.
\caseToolCall{
    \{\textquotedbl{}name\textquotedbl{}: \textquotedbl{}computer\_use\textquotedbl{}, \textquotedbl{}arguments\textquotedbl{}: \{\textquotedbl{}action\textquotedbl{}: \textquotedbl{}left\_click\textquotedbl{}, \textquotedbl{}coordinate\textquotedbl{}: [298, 534]\}\}
}
\\

\midrule

4
& \caseScreenshot{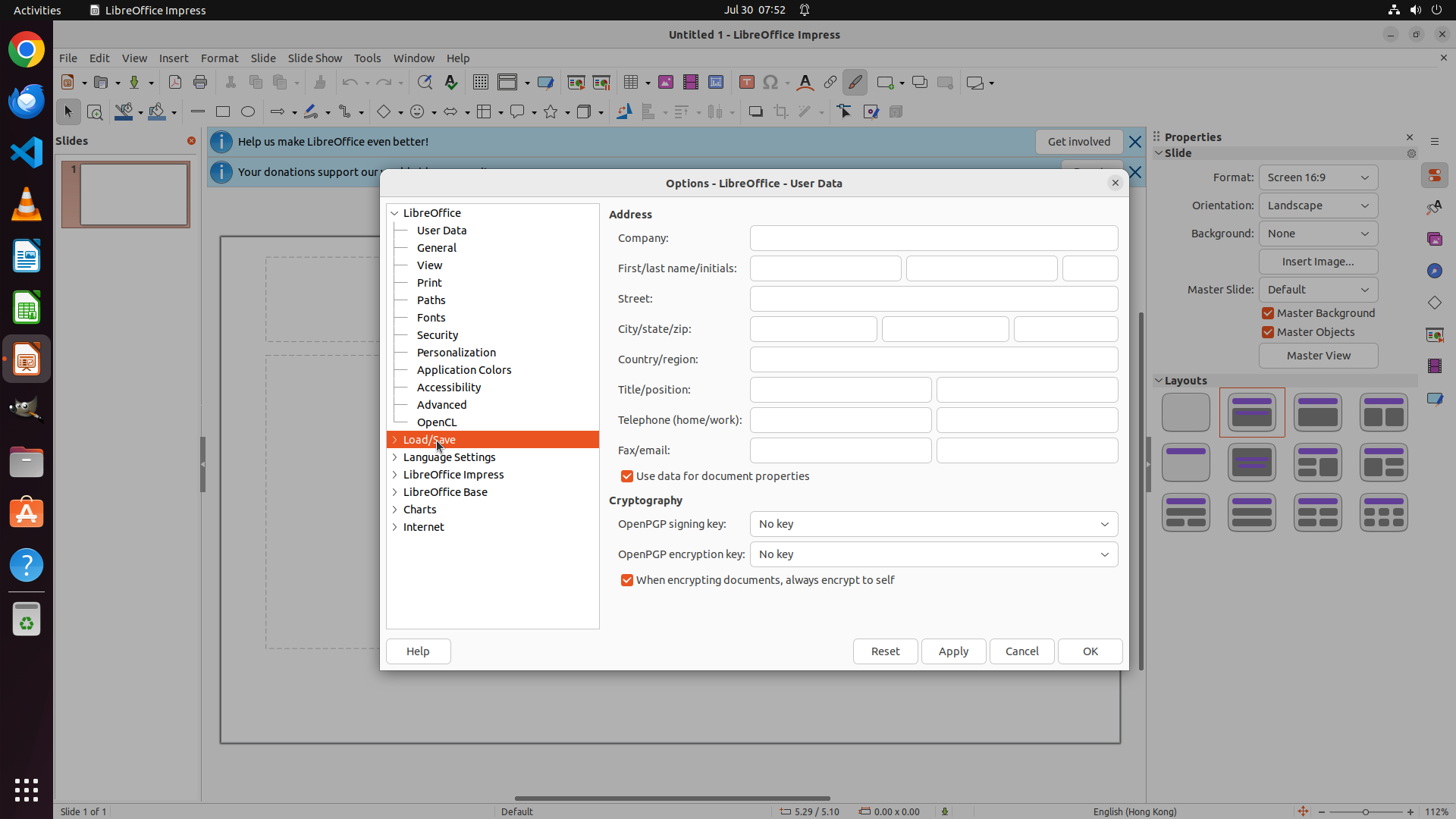}
&
Action: Click the small disclosure arrow to the left of ``Load/Save'' in the Options dialog's left navigation tree to expand the Load/Save category.
\caseToolCall{
    \{\textquotedbl{}name\textquotedbl{}: \textquotedbl{}computer\_use\textquotedbl{}, \textquotedbl{}arguments\textquotedbl{}: \{\textquotedbl{}action\textquotedbl{}: \textquotedbl{}left\_click\textquotedbl{}, \textquotedbl{}coordinate\textquotedbl{}: [271, 534]\}\}
}
\\

\midrule

5
& \caseScreenshot{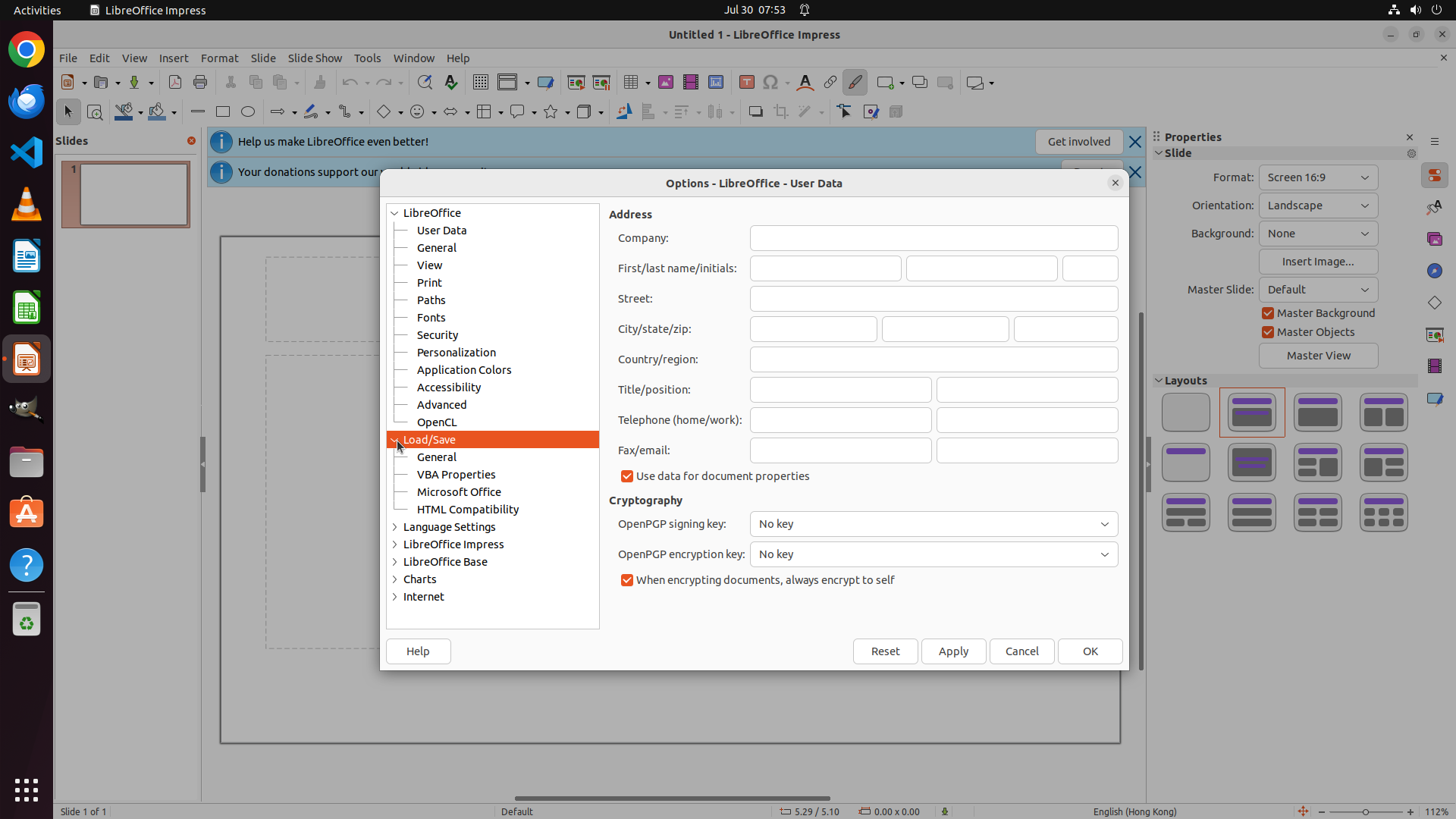}
&
Action: Click the ``General'' item under the ``Load/Save'' section in the Options dialog's left navigation pane.
\caseToolCall{
    \{\textquotedbl{}name\textquotedbl{}: \textquotedbl{}computer\_use\textquotedbl{}, \textquotedbl{}arguments\textquotedbl{}: \{\textquotedbl{}action\textquotedbl{}: \textquotedbl{}left\_click\textquotedbl{}, \textquotedbl{}coordinate\textquotedbl{}: [302, 554]\}\}
}
\\

\midrule

6
& \caseScreenshot{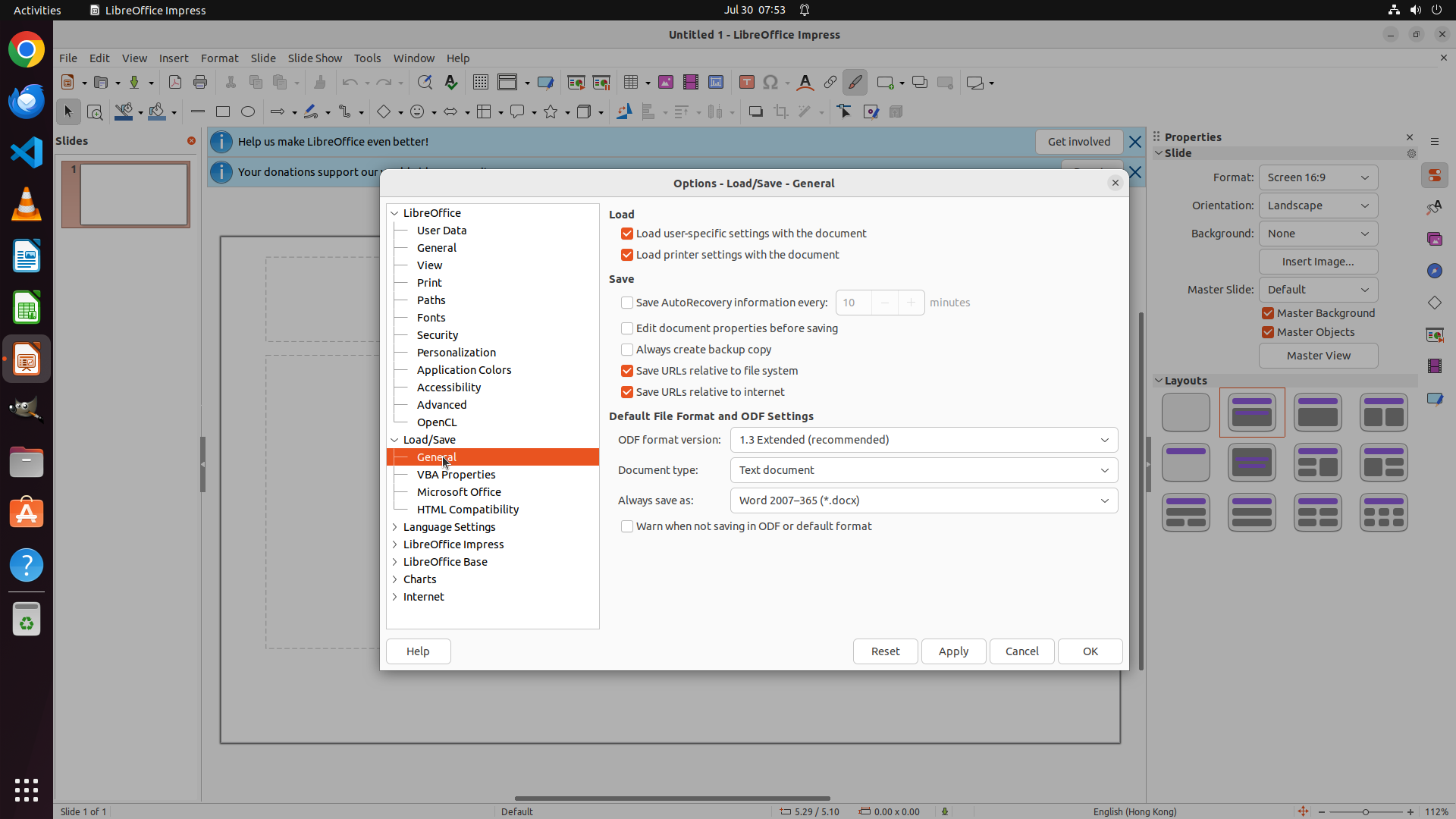}
&
Action: Click the numeric input box displaying ``10'' in the ``Save AutoRecovery information every:'' setting to focus it for editing.
\caseToolCall{
    \{\textquotedbl{}name\textquotedbl{}: \textquotedbl{}computer\_use\textquotedbl{}, \textquotedbl{}arguments\textquotedbl{}: \{\textquotedbl{}action\textquotedbl{}: \textquotedbl{}left\_click\textquotedbl{}, \textquotedbl{}coordinate\textquotedbl{}: [593, 367]\}\}
}
\\

\midrule

7
& \caseScreenshot{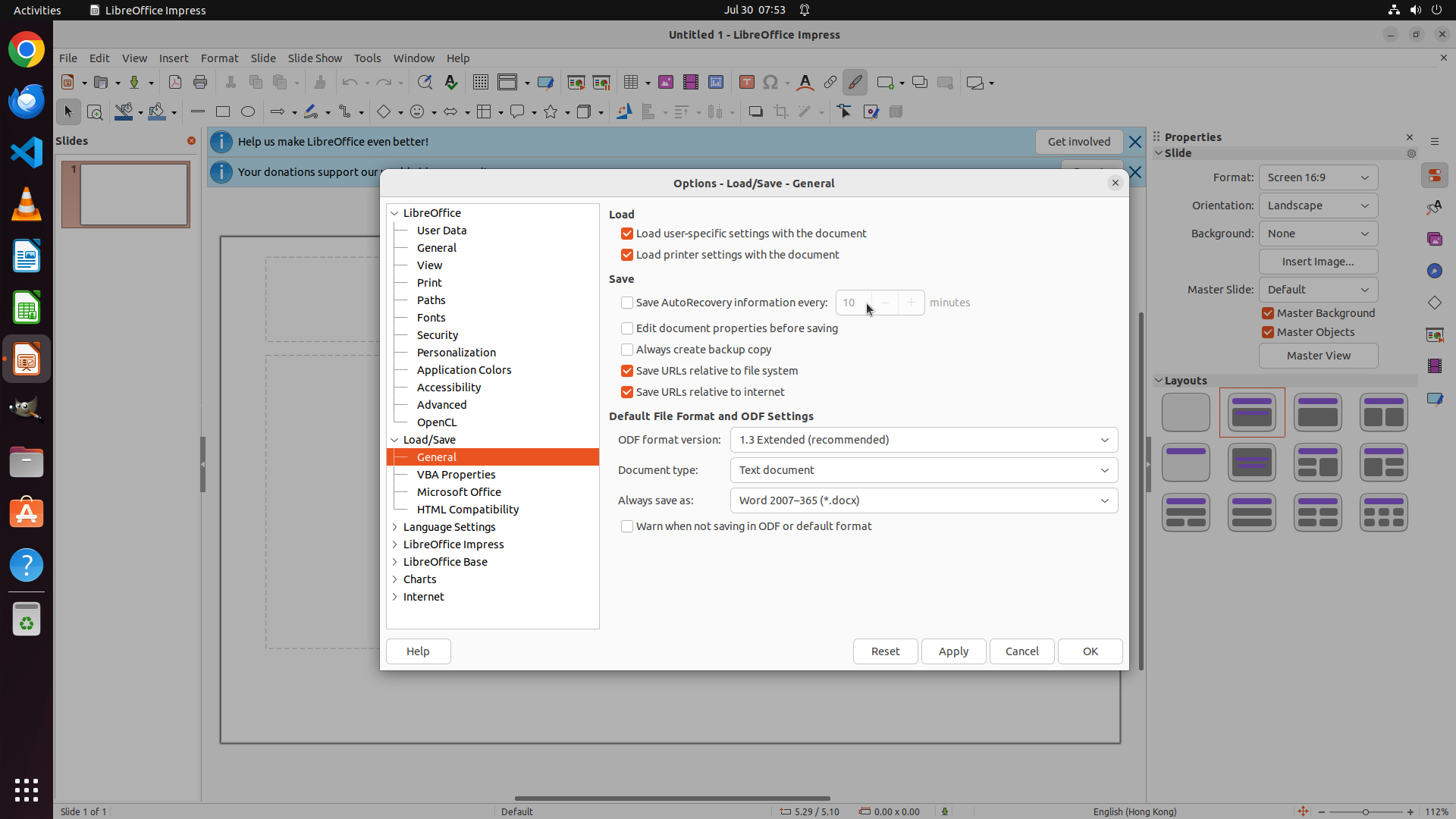}
&
Action: Press Ctrl+A to select all text in the currently focused AutoRecovery minutes input field.
\caseToolCall{
    \{\textquotedbl{}name\textquotedbl{}: \textquotedbl{}computer\_use\textquotedbl{}, \textquotedbl{}arguments\textquotedbl{}: \{\textquotedbl{}action\textquotedbl{}: \textquotedbl{}key\textquotedbl{}, \textquotedbl{}keys\textquotedbl{}: [\textquotedbl{}ctrl\textquotedbl{}, \textquotedbl{}a\textquotedbl{}]\}\}
}
\\

\midrule

8
& \caseScreenshot{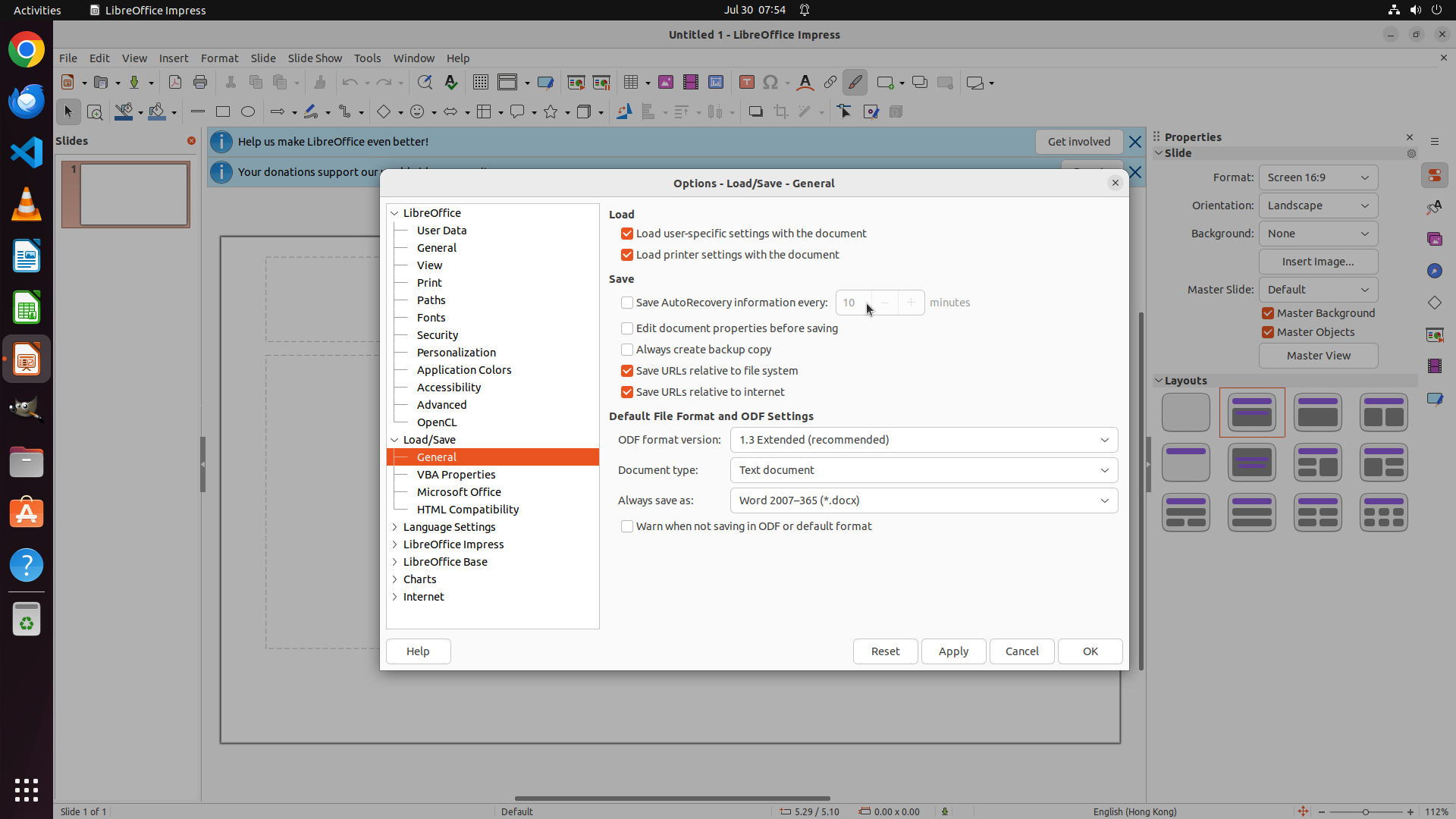}
&
Action: Type the character 3 to replace the selected value in the focused AutoRecovery minutes input field.
\caseToolCall{
    \{\textquotedbl{}name\textquotedbl{}: \textquotedbl{}computer\_use\textquotedbl{}, \textquotedbl{}arguments\textquotedbl{}: \{\textquotedbl{}action\textquotedbl{}: \textquotedbl{}type\textquotedbl{}, \textquotedbl{}text\textquotedbl{}: \textquotedbl{}3\textquotedbl{}\}\}
}
\\

\midrule

9
& \caseScreenshot{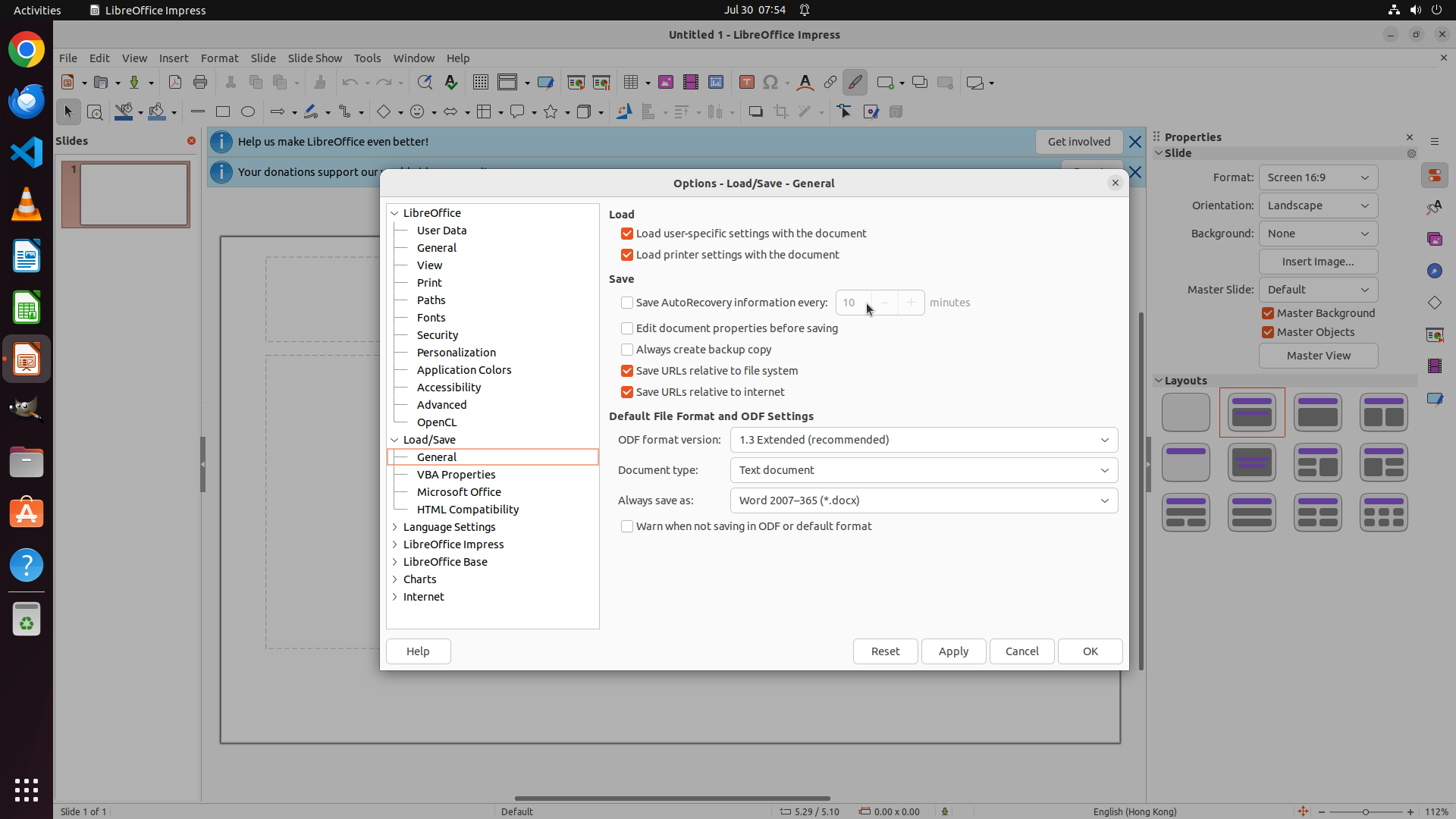}
&
Action: Click the OK button at the bottom-right of the Options dialog to apply the changes and close the dialog.
\caseToolCall{
    \{\textquotedbl{}name\textquotedbl{}: \textquotedbl{}computer\_use\textquotedbl{}, \textquotedbl{}arguments\textquotedbl{}: \{\textquotedbl{}action\textquotedbl{}: \textquotedbl{}left\_click\textquotedbl{}, \textquotedbl{}coordinate\textquotedbl{}: [746, 791]\}\}
}
\\

\midrule

10
& \caseScreenshot{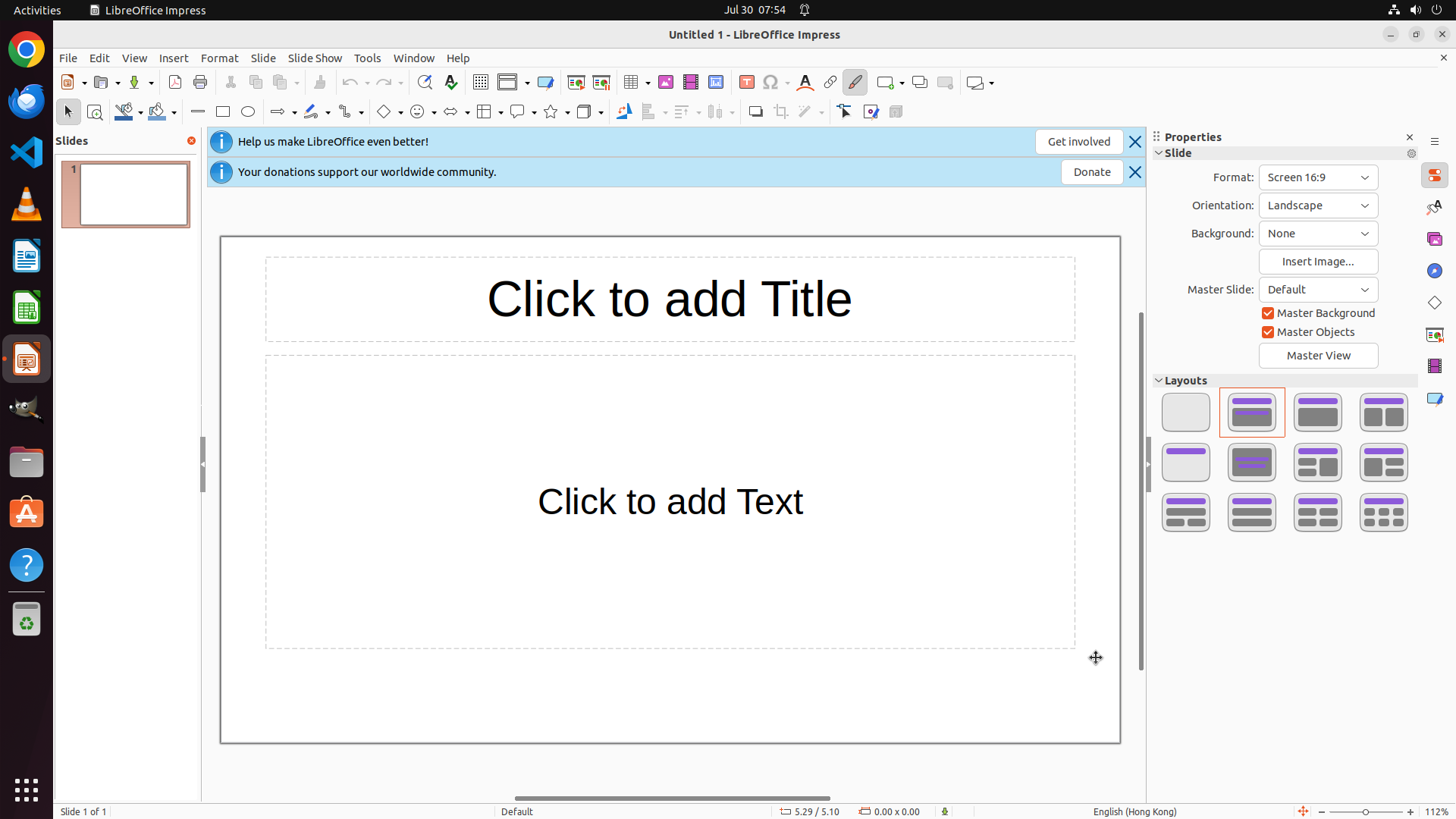}
&
Action: Terminate the session with a success status.
\caseToolCall{
    \{\textquotedbl{}name\textquotedbl{}: \textquotedbl{}computer\_use\textquotedbl{}, \textquotedbl{}arguments\textquotedbl{}: \{\textquotedbl{}action\textquotedbl{}: \textquotedbl{}terminate\textquotedbl{}, \textquotedbl{}status\textquotedbl{}: \textquotedbl{}success\textquotedbl{}\}\}
}
\\

\bottomrule
\end{longtblr}

\begin{longtblr}[
    caption={A successful rollout of the ComputerSD-trained policy on the same in-domain task.
        Task instruction: ``Enable auto-save every 3min for me, so that I don't need to hit \textquotedbl{}ctrl-s\textquotedbl{} that much''
    },
    label={tab:case_id_computersd},
]{
    width = \textwidth,
    colspec = {
        Q[c,t,wd=0.05\textwidth]
        X[40,c,t]
        X[60,l,t]
    },
    rowhead = 1,
    cells = {font=\small},
    row{1} = {font=\small\bfseries},
    colsep = 4pt,
    rowsep = 3pt,
    stretch = 0,
}
\toprule
Step & Screenshot & Model response \\
\midrule

1
& \caseScreenshot{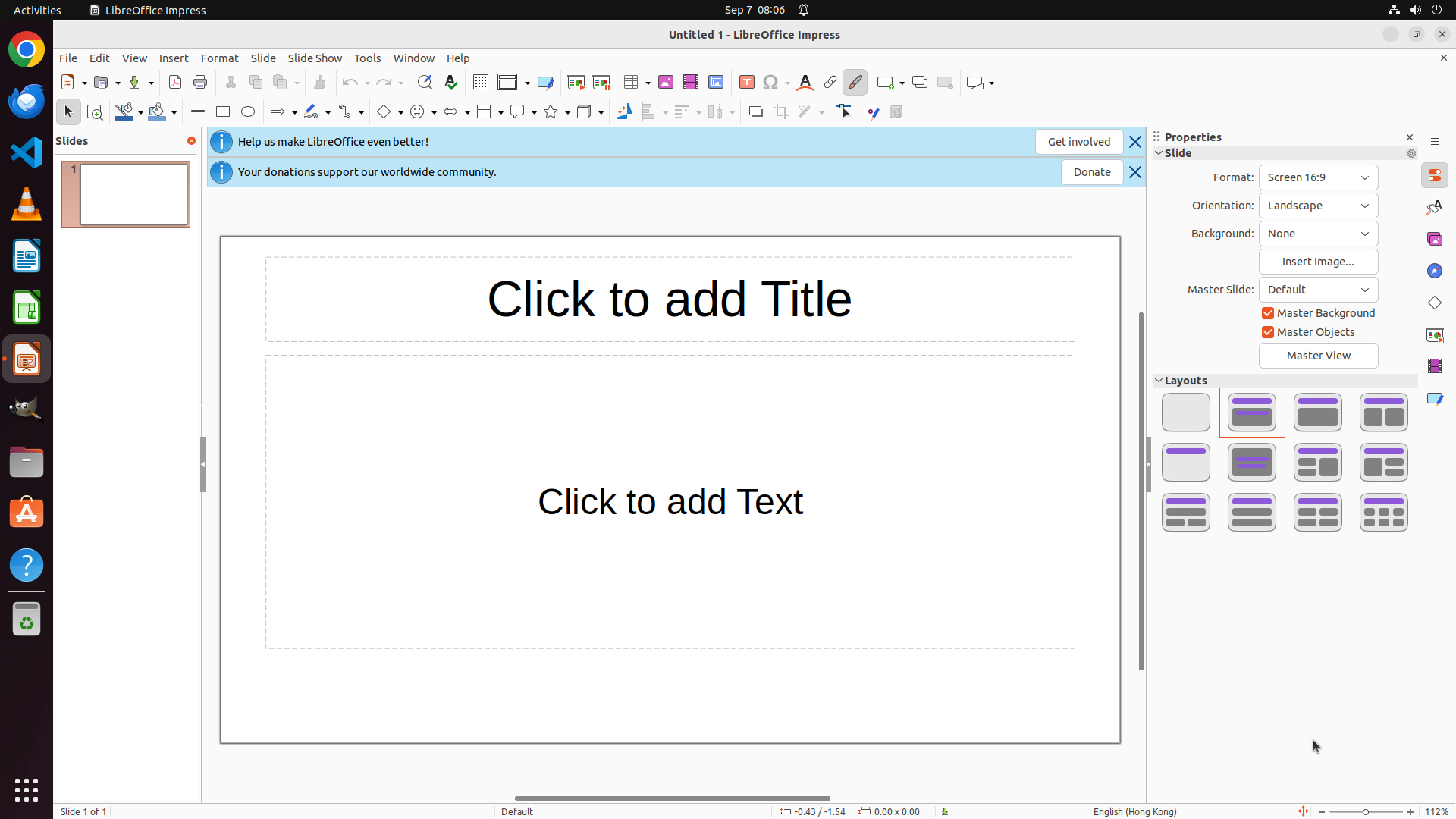}
&
Action: Click the ``Tools'' menu in the top menu bar of LibreOffice Impress to open its dropdown menu.
\caseToolCall{
    \{\textquotedbl{}name\textquotedbl{}: \textquotedbl{}computer\_use\textquotedbl{}, \textquotedbl{}arguments\textquotedbl{}: \{\textquotedbl{}action\textquotedbl{}: \textquotedbl{}left\_click\textquotedbl{}, \textquotedbl{}coordinate\textquotedbl{}: [251, 68]\}\}
}
\\

\midrule

2
& \caseScreenshot{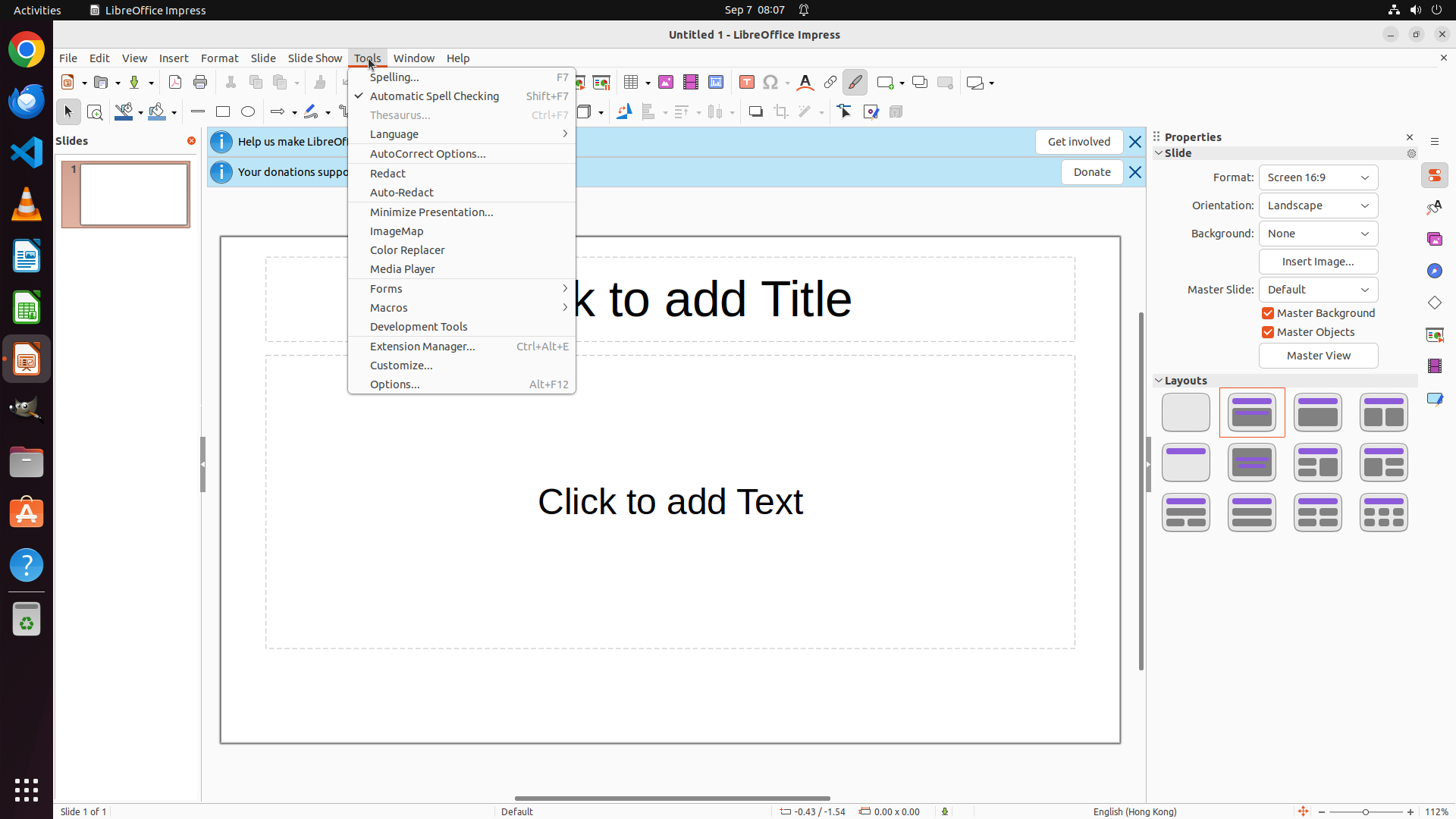}
&
Action: Click the ``Options\ldots{}'' item at the bottom of the open Tools menu to open the LibreOffice Options dialog.
\caseToolCall{
    \{\textquotedbl{}name\textquotedbl{}: \textquotedbl{}computer\_use\textquotedbl{}, \textquotedbl{}arguments\textquotedbl{}: \{\textquotedbl{}action\textquotedbl{}: \textquotedbl{}left\_click\textquotedbl{}, \textquotedbl{}coordinate\textquotedbl{}: [268, 467]\}\}
}
\\

\midrule

3
& \caseScreenshot{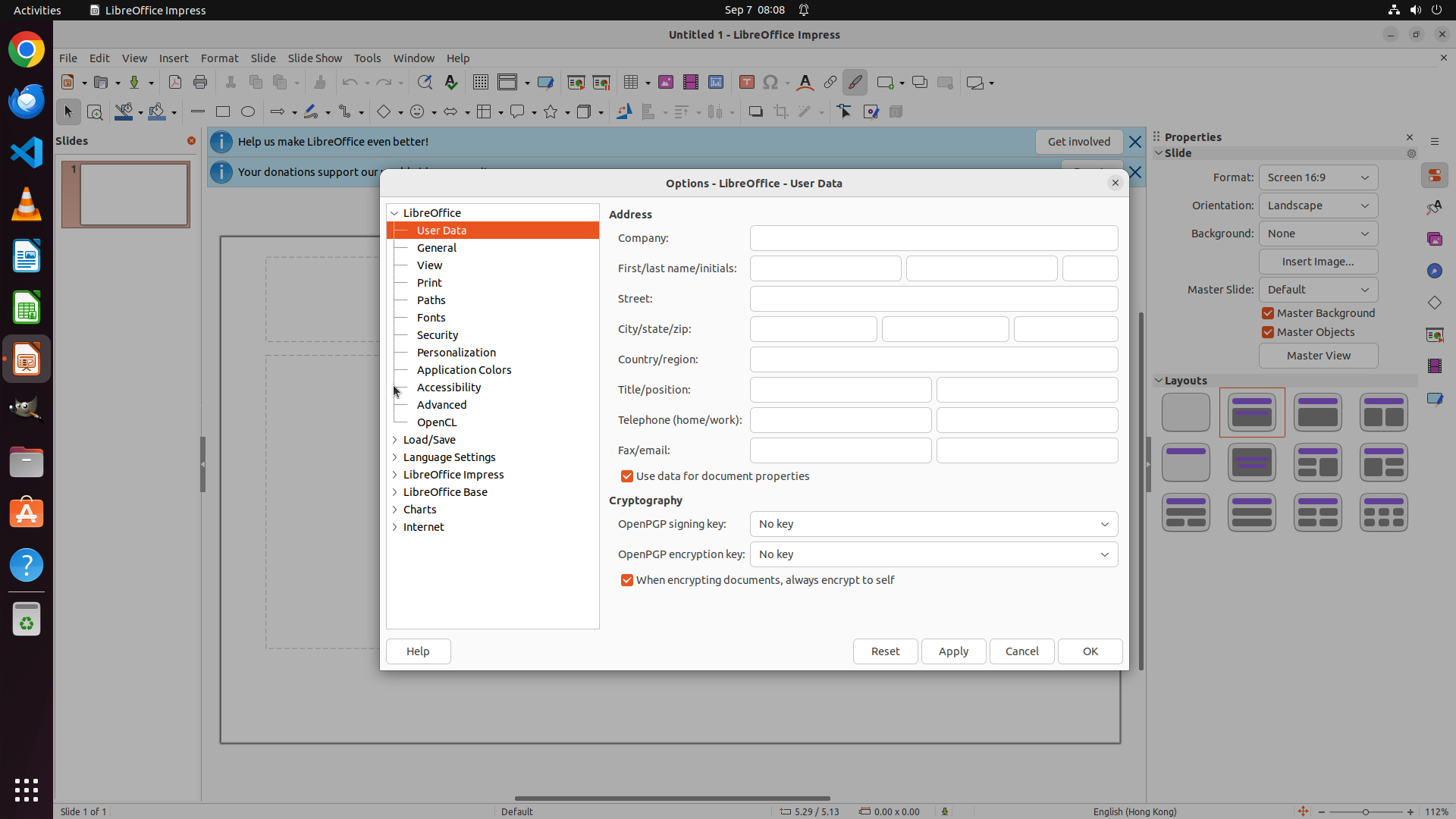}
&
Action: Click the small disclosure arrow to the left of ``Load/Save'' in the Options dialog's left sidebar to expand the Load/Save category.
\caseToolCall{
    \{\textquotedbl{}name\textquotedbl{}: \textquotedbl{}computer\_use\textquotedbl{}, \textquotedbl{}arguments\textquotedbl{}: \{\textquotedbl{}action\textquotedbl{}: \textquotedbl{}left\_click\textquotedbl{}, \textquotedbl{}coordinate\textquotedbl{}: [268, 531]\}\}
}
\\

\midrule

4
& \caseScreenshot{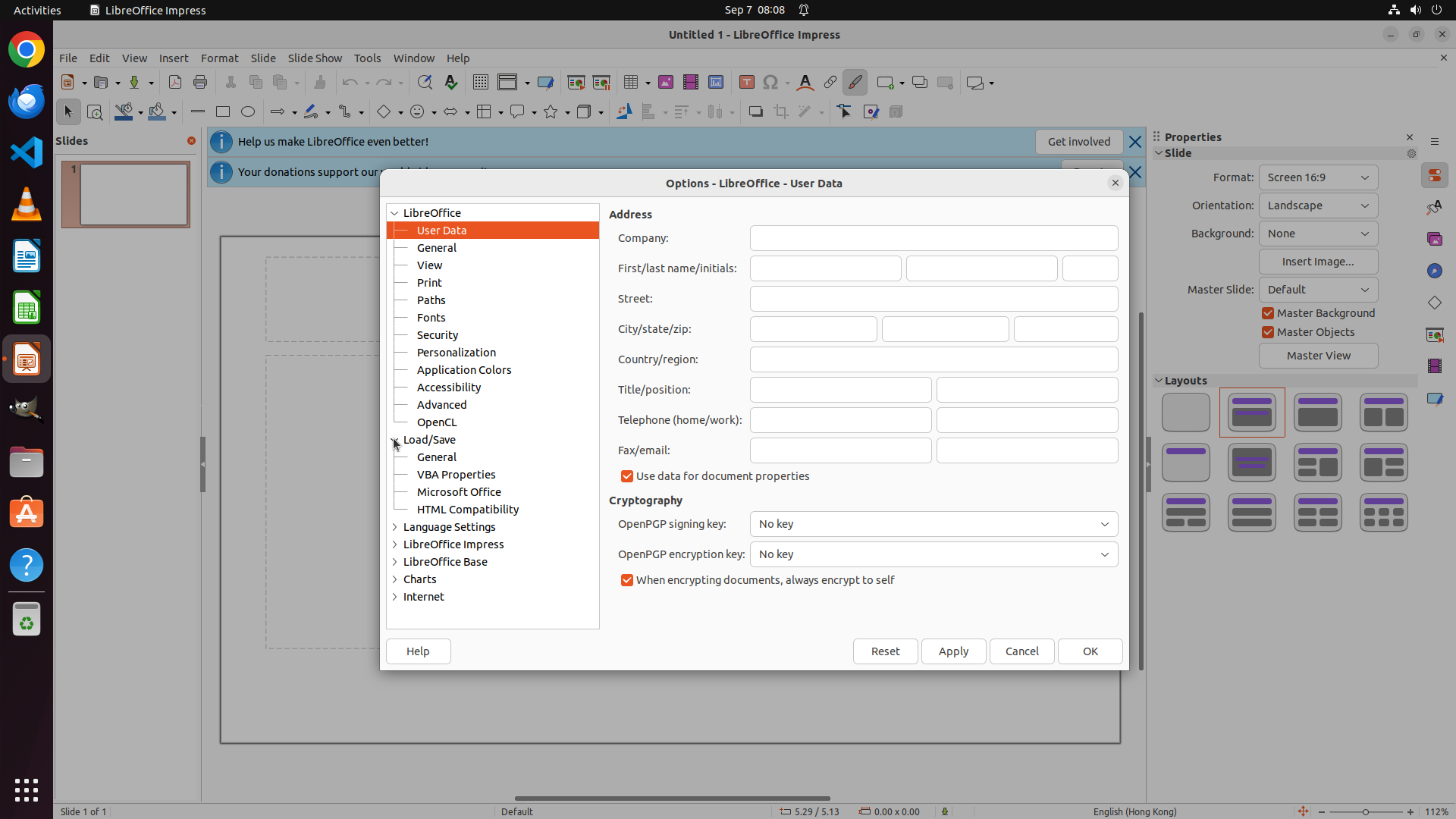}
&
Action: Click the ``General'' item under the ``Load/Save'' section in the Options dialog's left sidebar.
\caseToolCall{
    \{\textquotedbl{}name\textquotedbl{}: \textquotedbl{}computer\_use\textquotedbl{}, \textquotedbl{}arguments\textquotedbl{}: \{\textquotedbl{}action\textquotedbl{}: \textquotedbl{}left\_click\textquotedbl{}, \textquotedbl{}coordinate\textquotedbl{}: [298, 555]\}\}
}
\\

\midrule

5
& \caseScreenshot{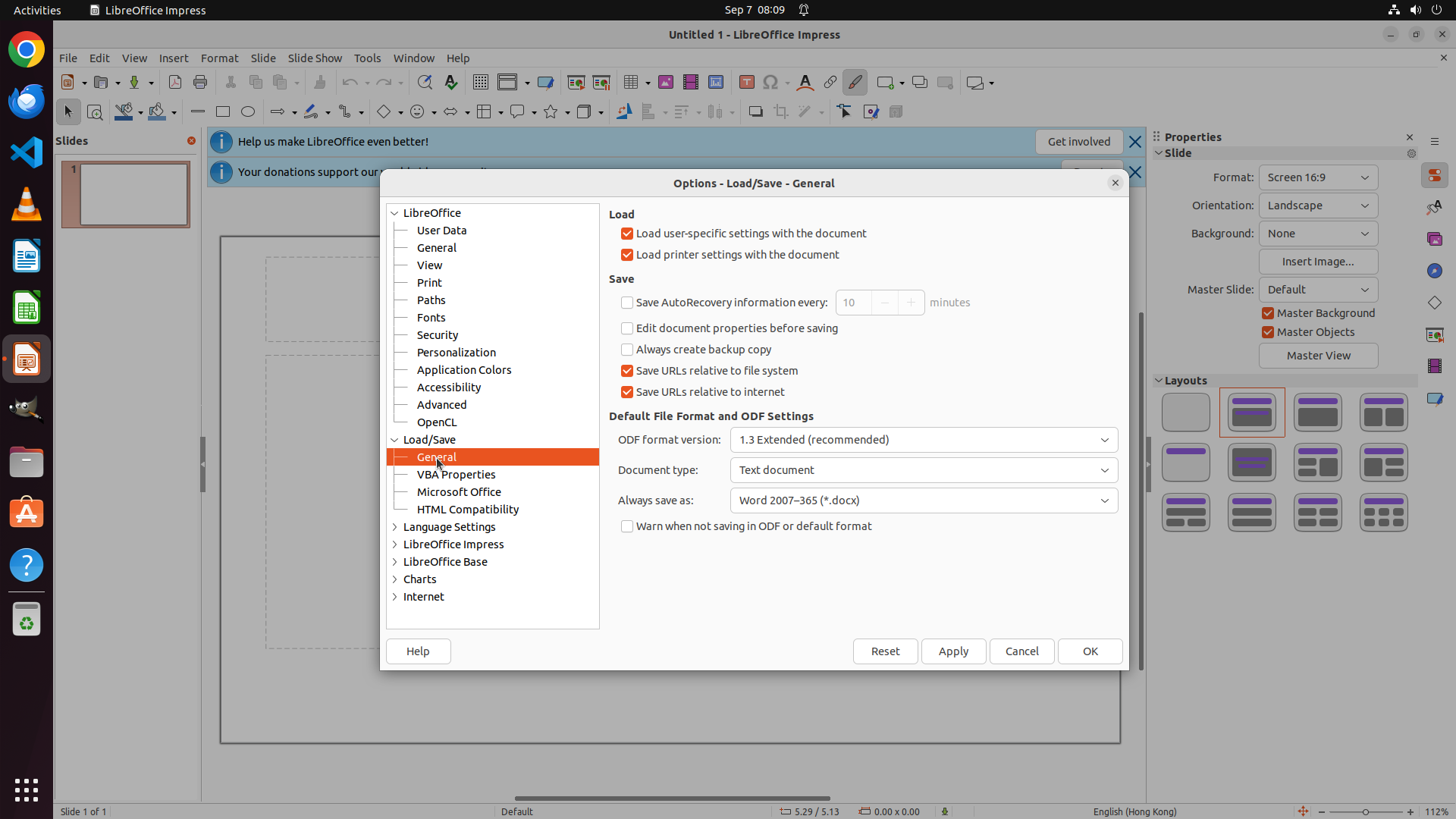}
&
Action: Click the ``Save AutoRecovery information every:'' checkbox in the Save section of the Options - Load/Save - General dialog to enable AutoRecovery.
\caseToolCall{
    \{\textquotedbl{}name\textquotedbl{}: \textquotedbl{}computer\_use\textquotedbl{}, \textquotedbl{}arguments\textquotedbl{}: \{\textquotedbl{}action\textquotedbl{}: \textquotedbl{}left\_click\textquotedbl{}, \textquotedbl{}coordinate\textquotedbl{}: [429, 367]\}\}
}
\\

\midrule

6
& \caseScreenshot{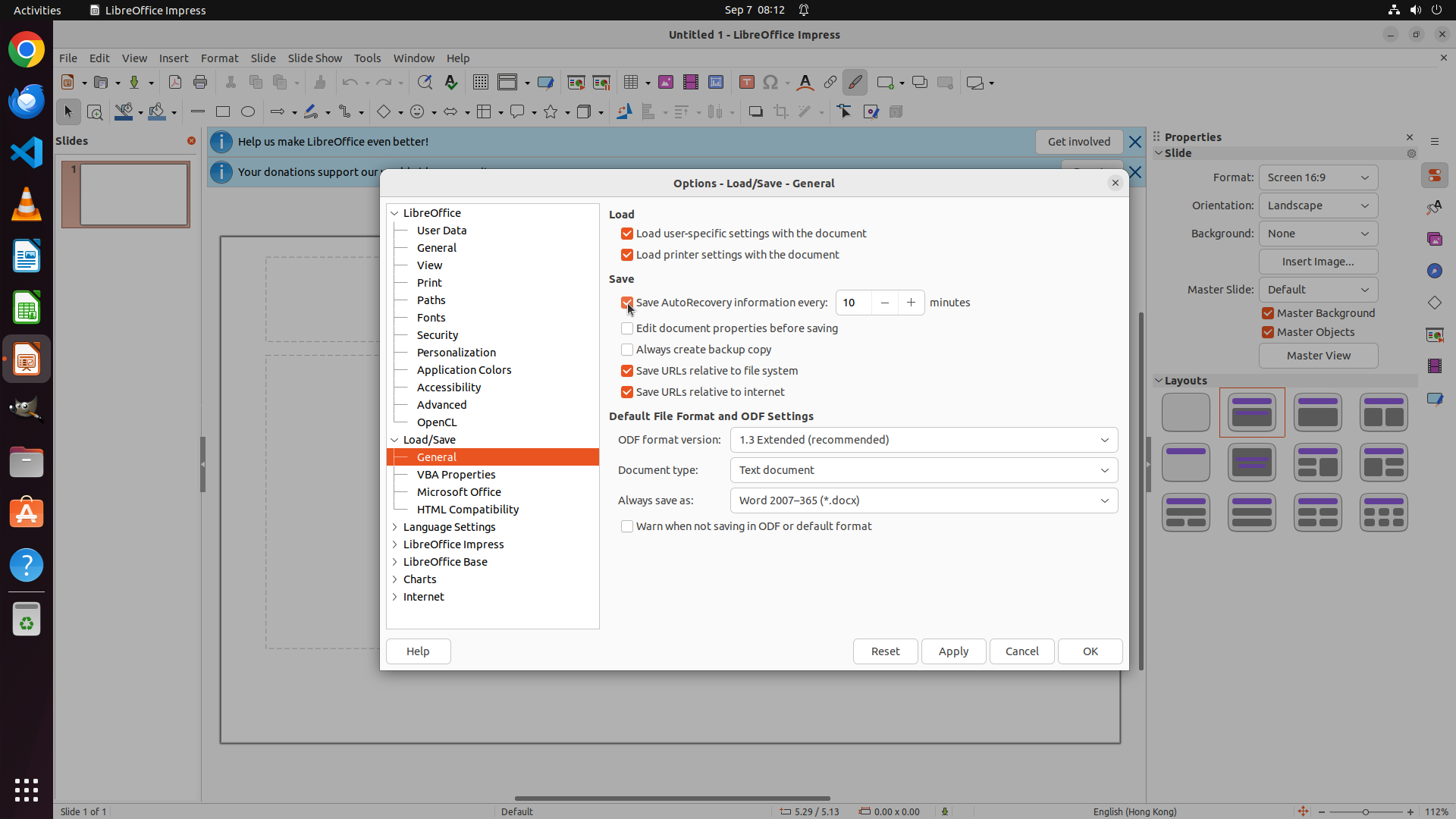}
&
Action: Triple-click the numeric value ``10'' inside the minutes input box next to ``Save AutoRecovery information every:'' in the Save section to select the current number.
\caseToolCall{
    \{\textquotedbl{}name\textquotedbl{}: \textquotedbl{}computer\_use\textquotedbl{}, \textquotedbl{}arguments\textquotedbl{}: \{\textquotedbl{}action\textquotedbl{}: \textquotedbl{}double\_click\textquotedbl{}, \textquotedbl{}coordinate\textquotedbl{}: [581, 367]\}\}
}
\\

\midrule

7
& \caseScreenshot{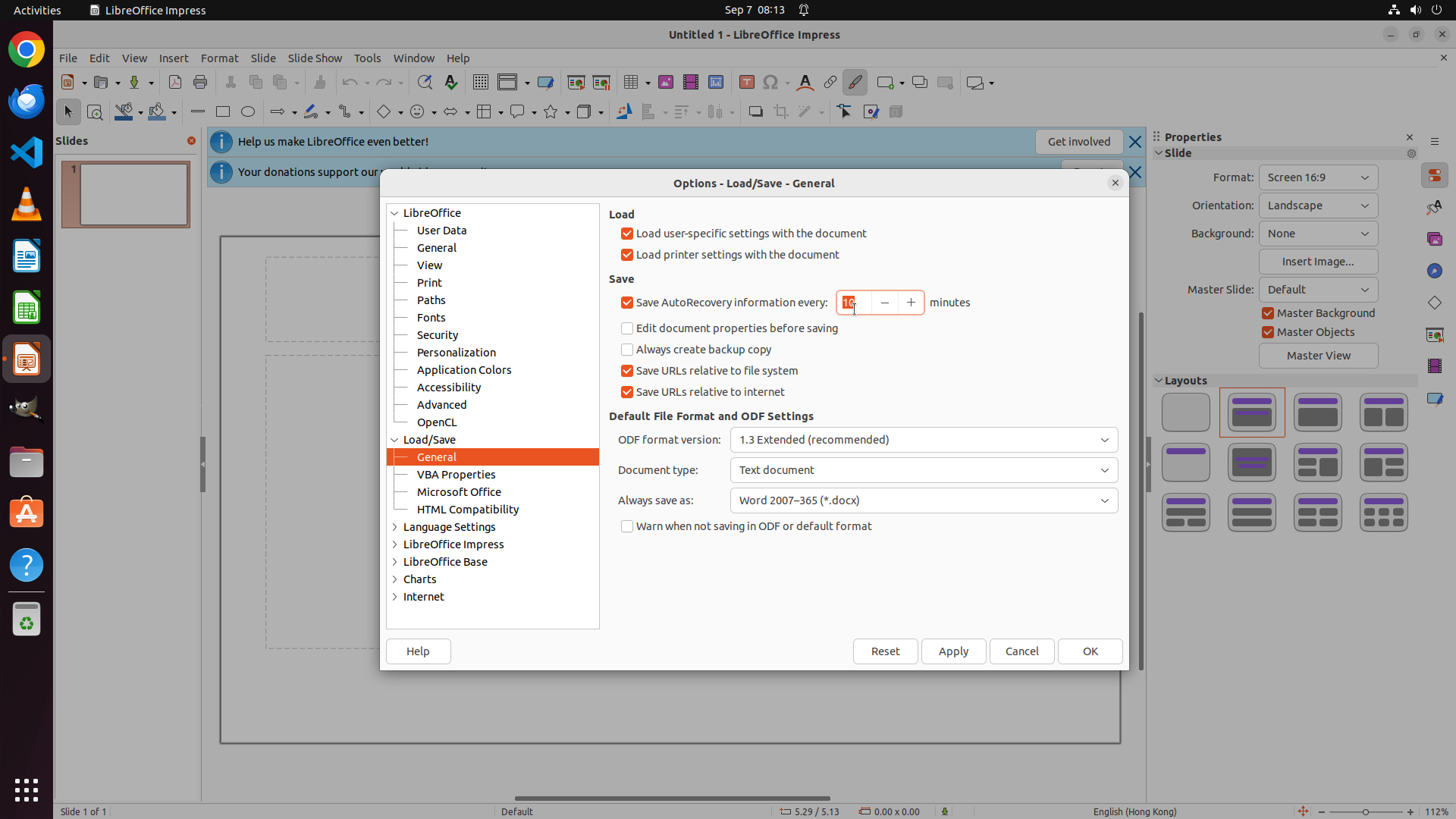}
&
Action: Type the character 3 to replace the selected ``10'' in the minutes input box, setting the AutoRecovery interval to 3 minutes.
\caseToolCall{
    \{\textquotedbl{}name\textquotedbl{}: \textquotedbl{}computer\_use\textquotedbl{}, \textquotedbl{}arguments\textquotedbl{}: \{\textquotedbl{}action\textquotedbl{}: \textquotedbl{}type\textquotedbl{}, \textquotedbl{}text\textquotedbl{}: \textquotedbl{}3\textquotedbl{}\}\}
}
\\

\midrule

8
& \caseScreenshot{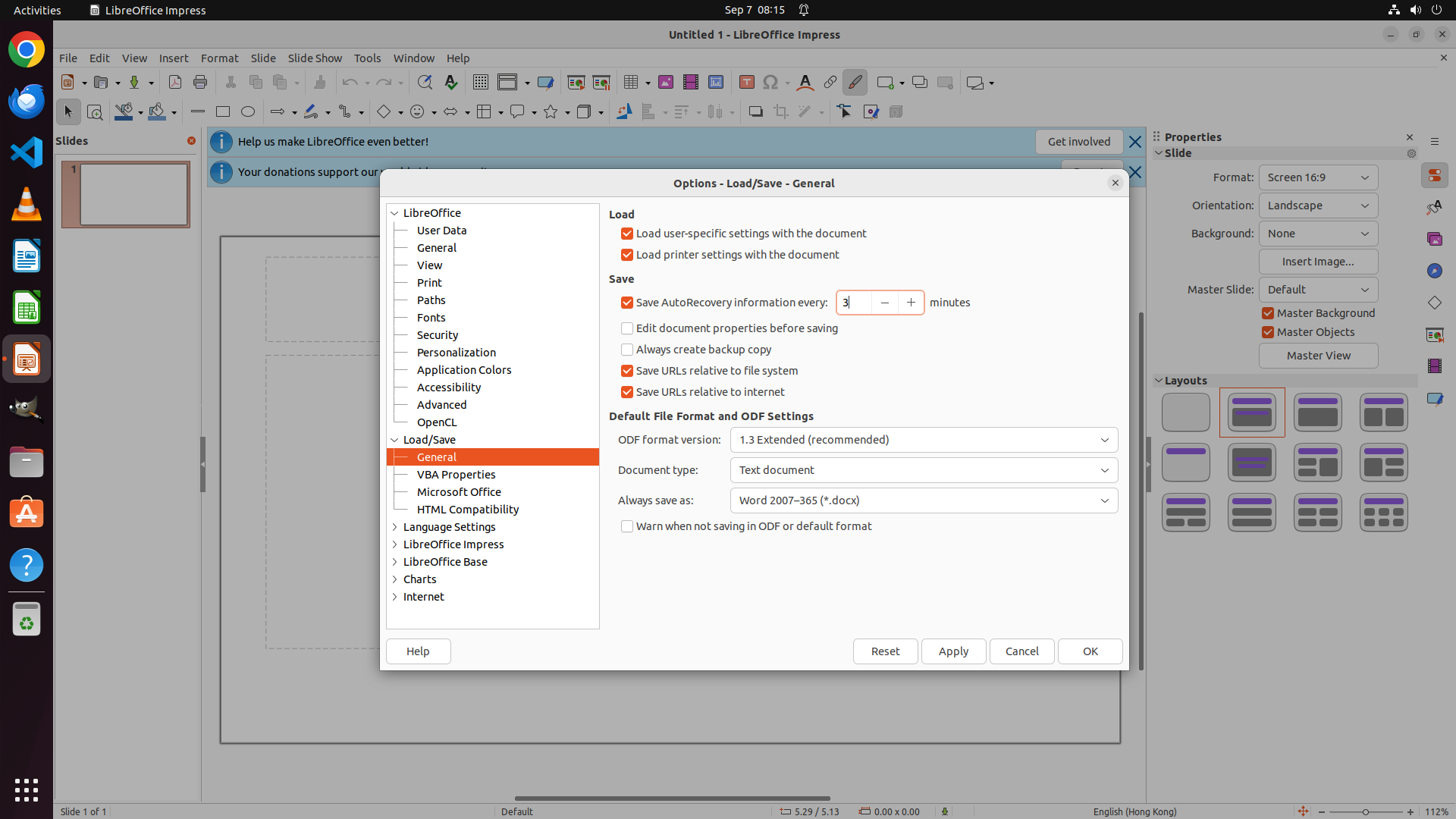}
&
Action: Click the ``OK'' button at the bottom-right of the ``Options - Load/Save - General'' dialog to save the changes and close the dialog.
\caseToolCall{
    \{\textquotedbl{}name\textquotedbl{}: \textquotedbl{}computer\_use\textquotedbl{}, \textquotedbl{}arguments\textquotedbl{}: \{\textquotedbl{}action\textquotedbl{}: \textquotedbl{}left\_click\textquotedbl{}, \textquotedbl{}coordinate\textquotedbl{}: [746, 791]\}\}
}
\\

\midrule

9
& \caseScreenshot{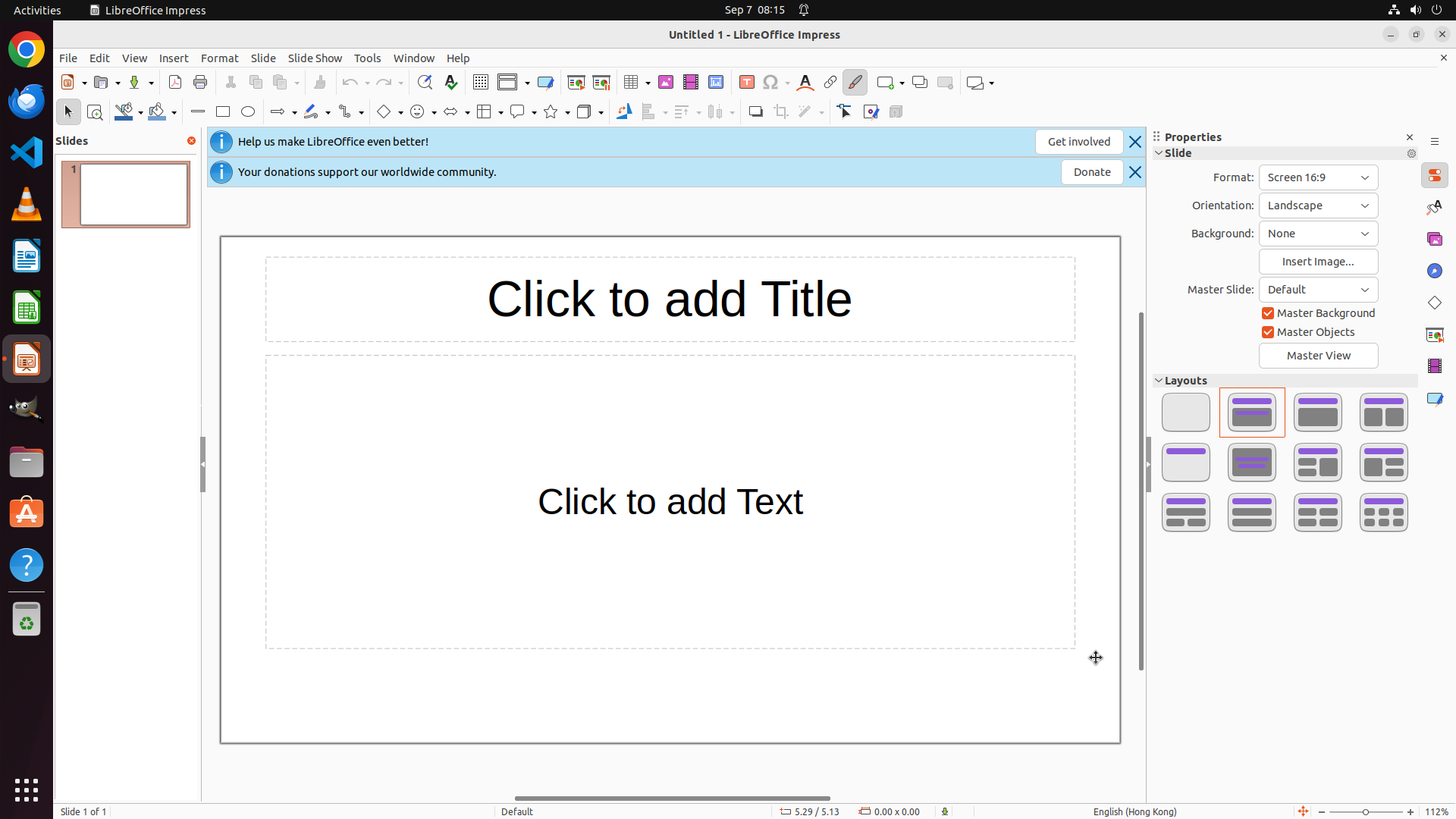}
&
Action: Terminate the session indicating success.
\caseToolCall{
    \{\textquotedbl{}name\textquotedbl{}: \textquotedbl{}computer\_use\textquotedbl{}, \textquotedbl{}arguments\textquotedbl{}: \{\textquotedbl{}action\textquotedbl{}: \textquotedbl{}terminate\textquotedbl{}, \textquotedbl{}status\textquotedbl{}: \textquotedbl{}success\textquotedbl{}\}\}
}
\\

\bottomrule
\end{longtblr}

\begin{longtblr}[
    caption={An unsuccessful rollout of the outcome-only GRPO-trained policy on an OOD task.
        Task instruction: ``Lately I have changed my English name to Thomas. I want to update my username. Could you help me change the username in chrome profiles to Thomas?''
    },
    label={tab:case_ood_grpo},
]{
    width = \textwidth,
    colspec = {
        Q[c,t,wd=0.05\textwidth]
        X[40,c,t]
        X[60,l,t]
    },
    rowhead = 1,
    cells = {font=\small},
    row{1} = {font=\small\bfseries},
    colsep = 4pt,
    rowsep = 3pt,
    stretch = 0,
}
\toprule
Step & Screenshot & Model response \\
\midrule

1
& \caseScreenshot{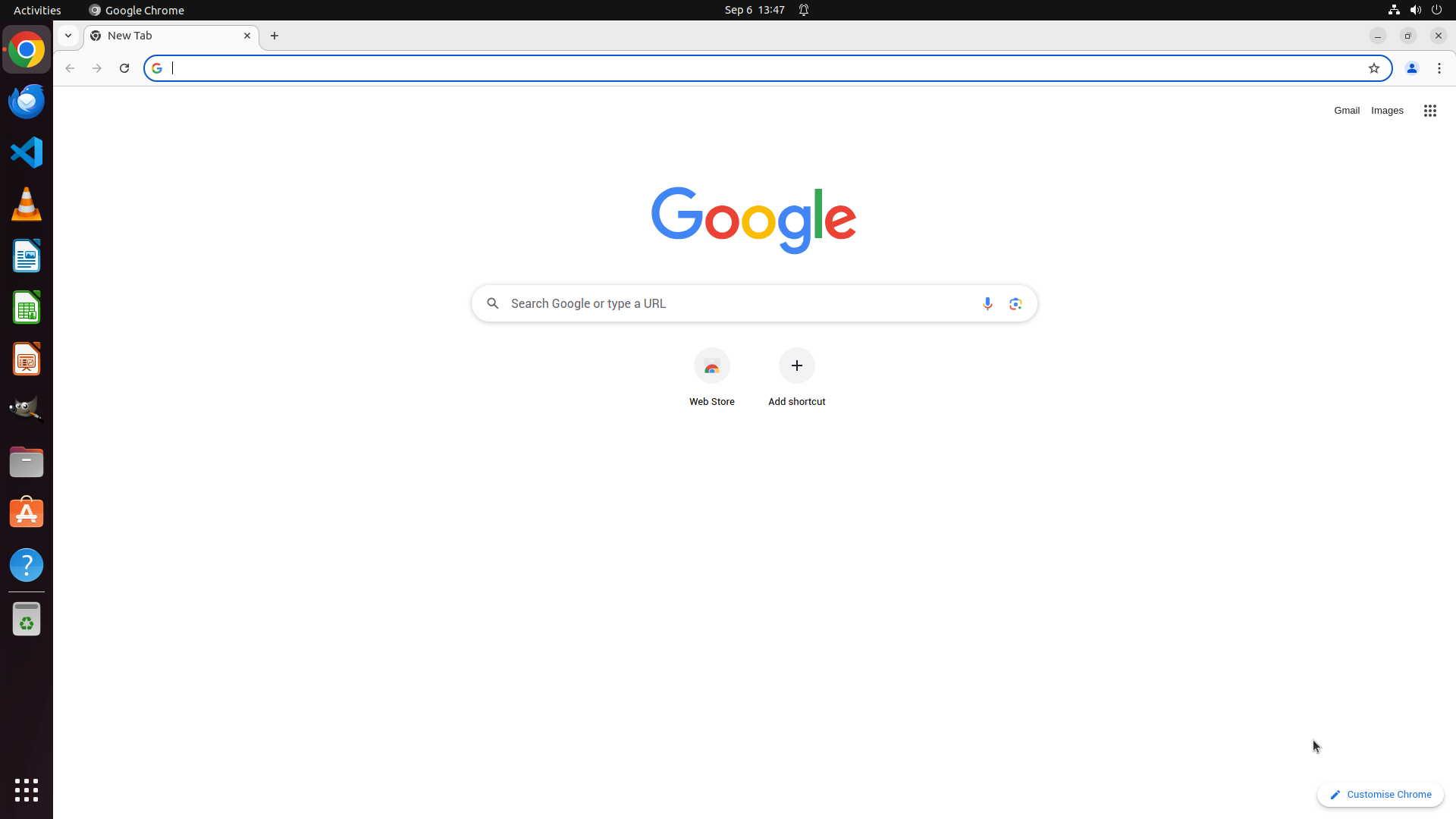}
&
Action: Click the circular user profile avatar button in the top-right of the Chrome toolbar to open the profile menu.
\caseToolCall{
    \{\textquotedbl{}name\textquotedbl{}: \textquotedbl{}computer\_use\textquotedbl{}, \textquotedbl{}arguments\textquotedbl{}: \{\textquotedbl{}action\textquotedbl{}: \textquotedbl{}left\_click\textquotedbl{}, \textquotedbl{}coordinate\textquotedbl{}: [968, 81]\}\}
}
\\

\midrule

2
& \caseScreenshot{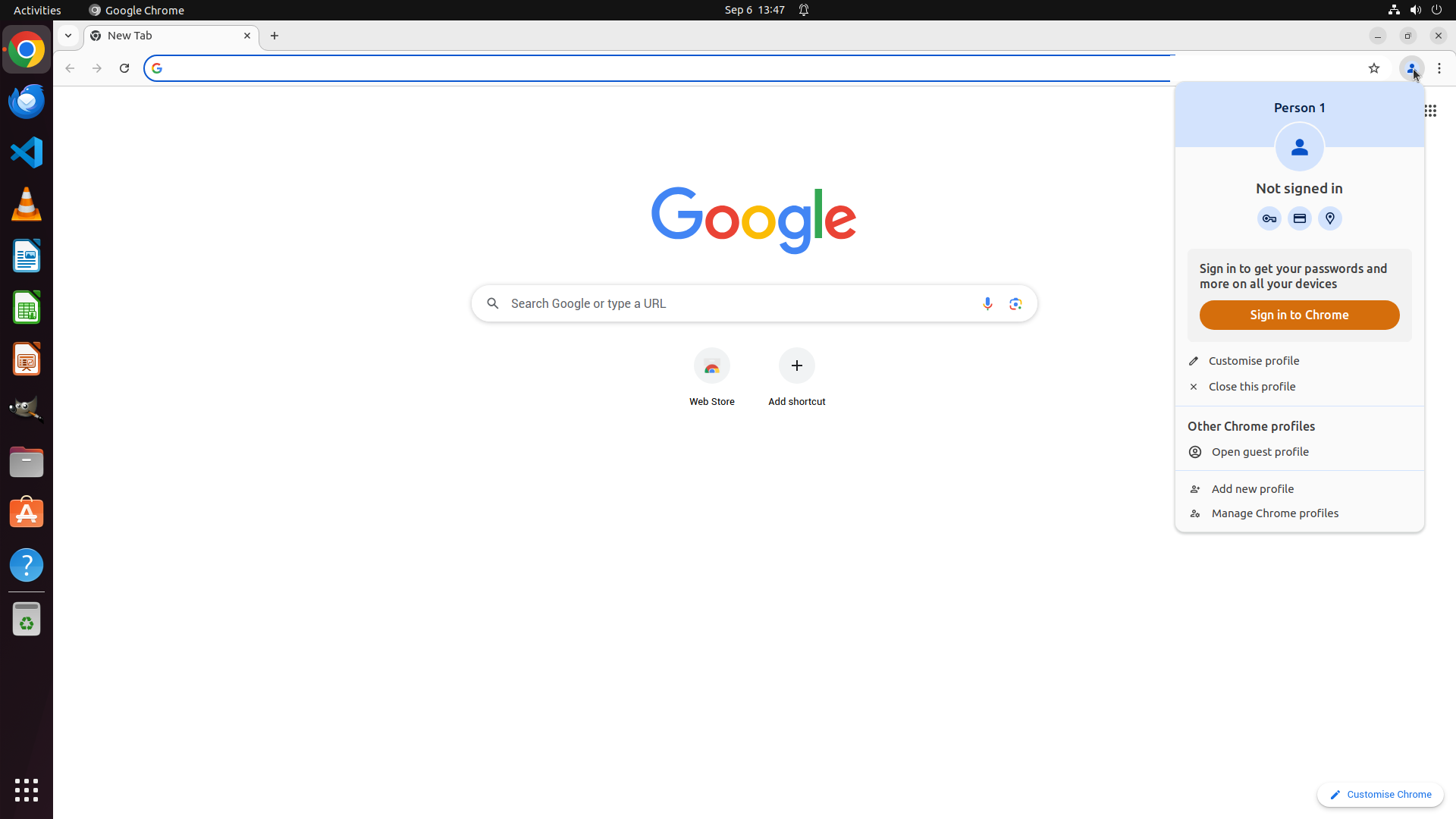}
&
Action: Click the ``Customise profile'' option in the open profile menu to open the profile customization settings.
\caseToolCall{
    \{\textquotedbl{}name\textquotedbl{}: \textquotedbl{}computer\_use\textquotedbl{}, \textquotedbl{}arguments\textquotedbl{}: \{\textquotedbl{}action\textquotedbl{}: \textquotedbl{}left\_click\textquotedbl{}, \textquotedbl{}coordinate\textquotedbl{}: [877, 438]\}\}
}
\\

\midrule

3
& \caseScreenshot{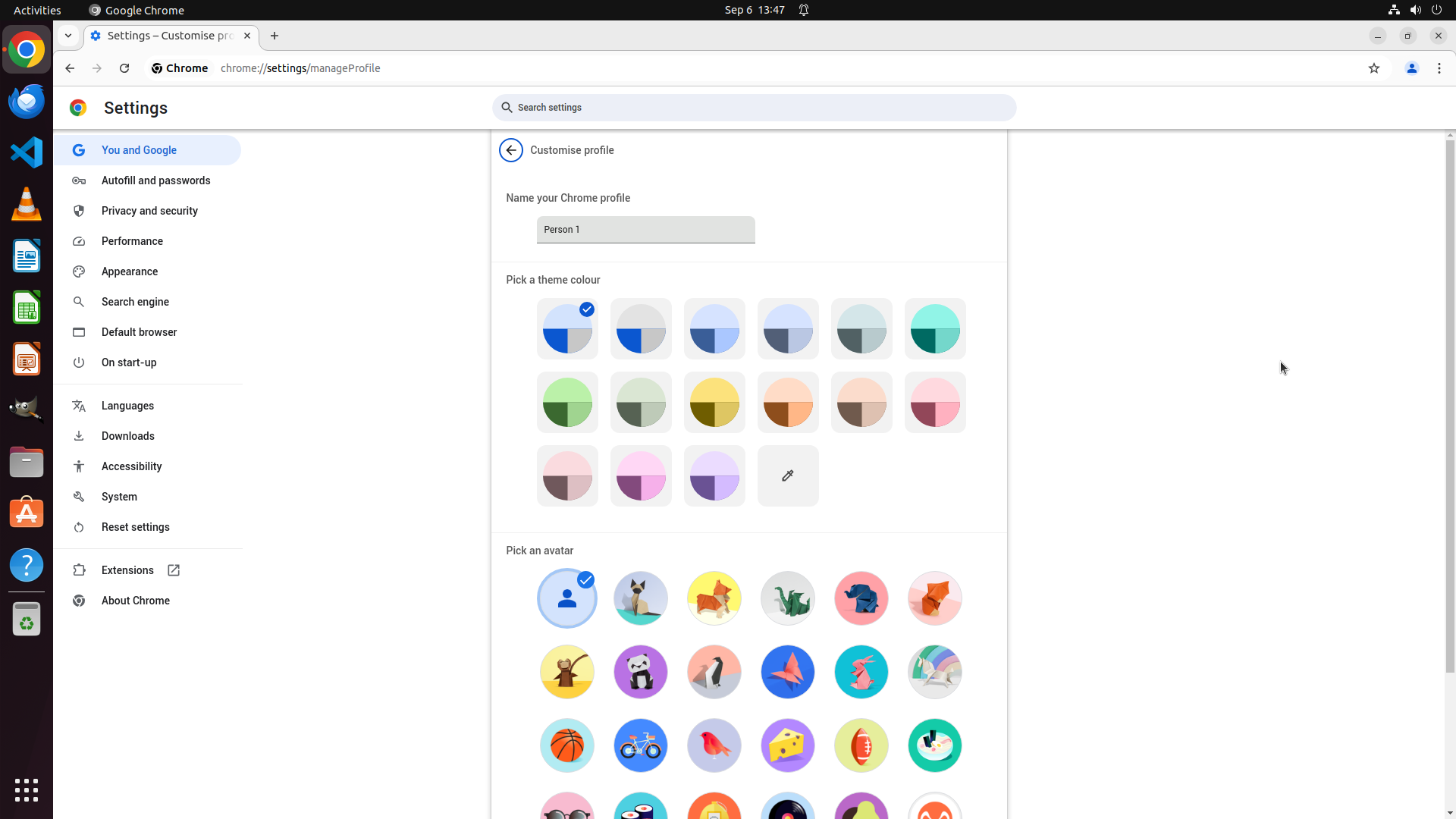}
&
Action: Click the ``Name your Chrome profile'' text field that currently displays ``Person 1'' to focus it for editing.
\caseToolCall{
    \{\textquotedbl{}name\textquotedbl{}: \textquotedbl{}computer\_use\textquotedbl{}, \textquotedbl{}arguments\textquotedbl{}: \{\textquotedbl{}action\textquotedbl{}: \textquotedbl{}left\_click\textquotedbl{}, \textquotedbl{}coordinate\textquotedbl{}: [404, 277]\}\}
}
\\

\midrule

4
& \caseScreenshot{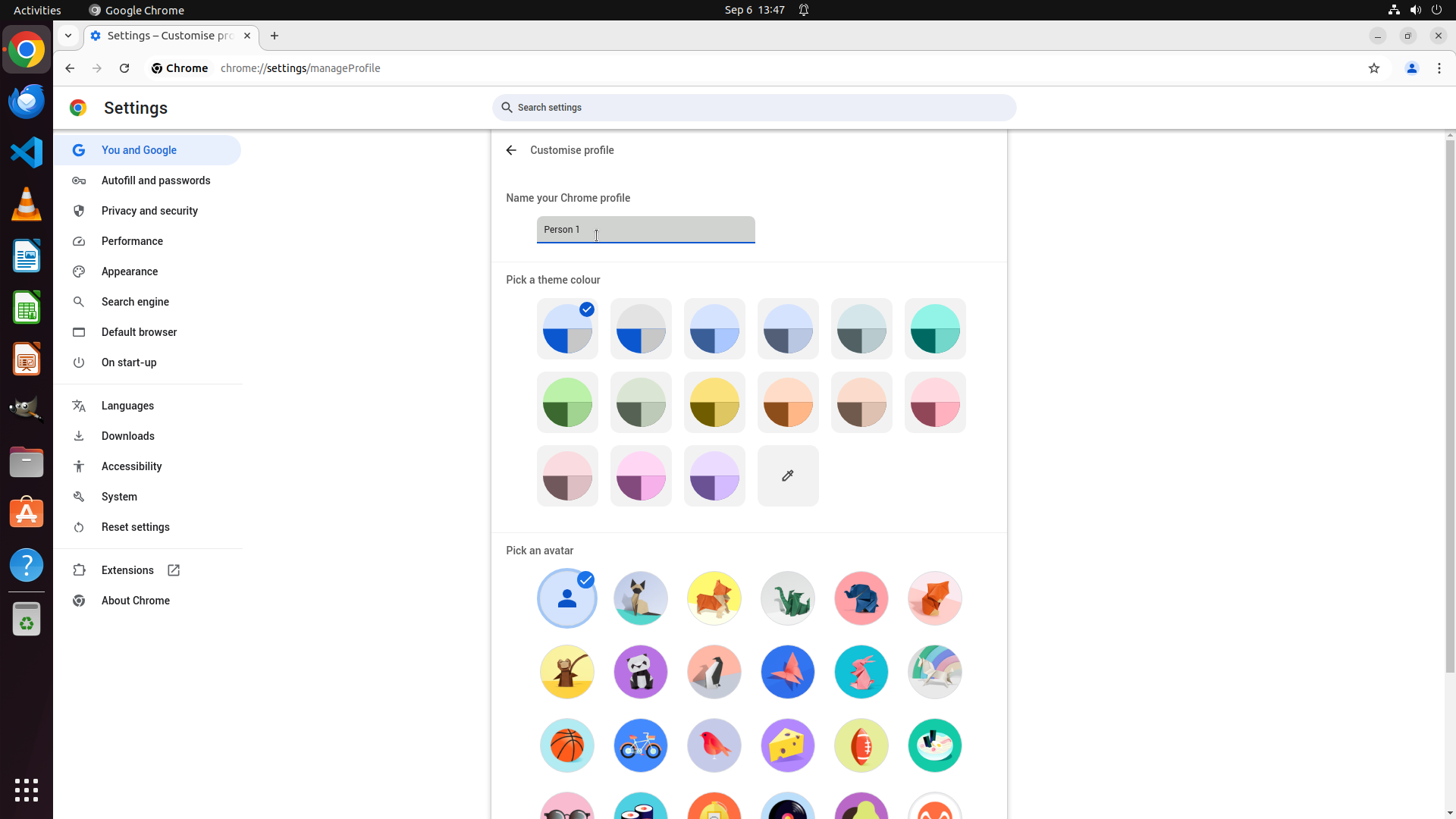}
&
Action: Press Ctrl+A to select all text in the focused ``Name your Chrome profile'' field.
\caseToolCall{
    \{\textquotedbl{}name\textquotedbl{}: \textquotedbl{}computer\_use\textquotedbl{}, \textquotedbl{}arguments\textquotedbl{}: \{\textquotedbl{}action\textquotedbl{}: \textquotedbl{}key\textquotedbl{}, \textquotedbl{}keys\textquotedbl{}: [\textquotedbl{}ctrl\textquotedbl{}, \textquotedbl{}a\textquotedbl{}]\}\}
}
\\

\midrule

5
& \caseScreenshot{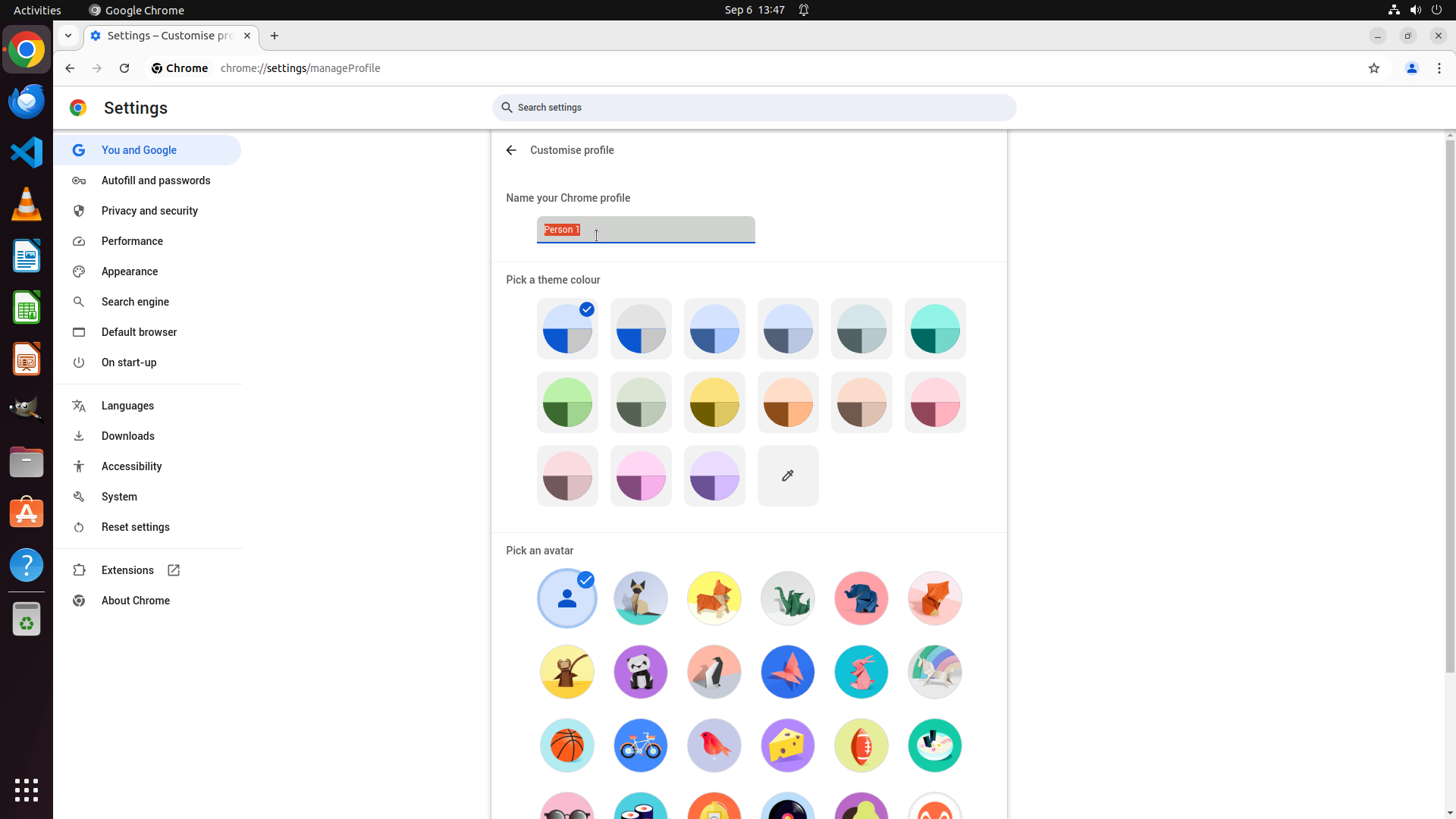}
&
Action: Type the text exactly: Thomas. This will replace the selected ``Person 1'' in the focused ``Name your Chrome profile'' field so it reads ``Thomas''.
\caseToolCall{
    \{\textquotedbl{}name\textquotedbl{}: \textquotedbl{}computer\_use\textquotedbl{}, \textquotedbl{}arguments\textquotedbl{}: \{\textquotedbl{}action\textquotedbl{}: \textquotedbl{}type\textquotedbl{}, \textquotedbl{}text\textquotedbl{}: \textquotedbl{}Thomas\textquotedbl{}\}\}
}
\\

\midrule

6
& \caseScreenshot{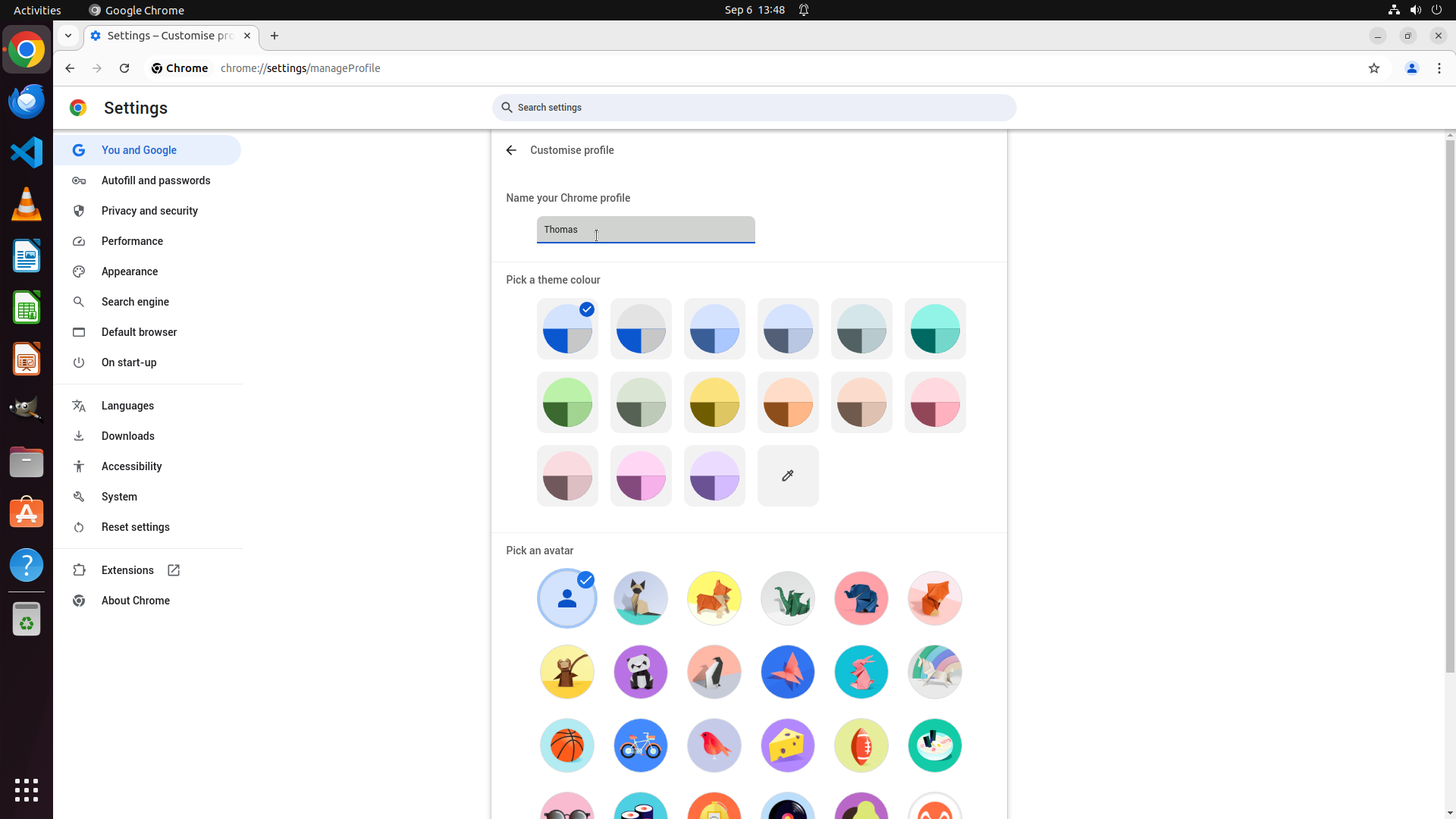}
&
Action: Terminate the session and mark the task as successfully completed.
\caseToolCall{
    \{\textquotedbl{}name\textquotedbl{}: \textquotedbl{}computer\_use\textquotedbl{}, \textquotedbl{}arguments\textquotedbl{}: \{\textquotedbl{}action\textquotedbl{}: \textquotedbl{}terminate\textquotedbl{}, \textquotedbl{}status\textquotedbl{}: \textquotedbl{}success\textquotedbl{}\}\}
}
\\

\bottomrule
\end{longtblr}

\begin{longtblr}[
    caption={A successful rollout of the ComputerSD-trained policy on the same OOD task.
        Task instruction: ``Lately I have changed my English name to Thomas. I want to update my username. Could you help me change the username in chrome profiles to Thomas?''
    },
    label={tab:case_ood_computersd},
]{
    width = \textwidth,
    colspec = {
        Q[c,t,wd=0.05\textwidth]
        X[40,c,t]
        X[60,l,t]
    },
    rowhead = 1,
    cells = {font=\small},
    row{1} = {font=\small\bfseries},
    colsep = 4pt,
    rowsep = 3pt,
    stretch = 0,
}
\toprule
Step & Screenshot & Model response \\
\midrule

1
& \caseScreenshot{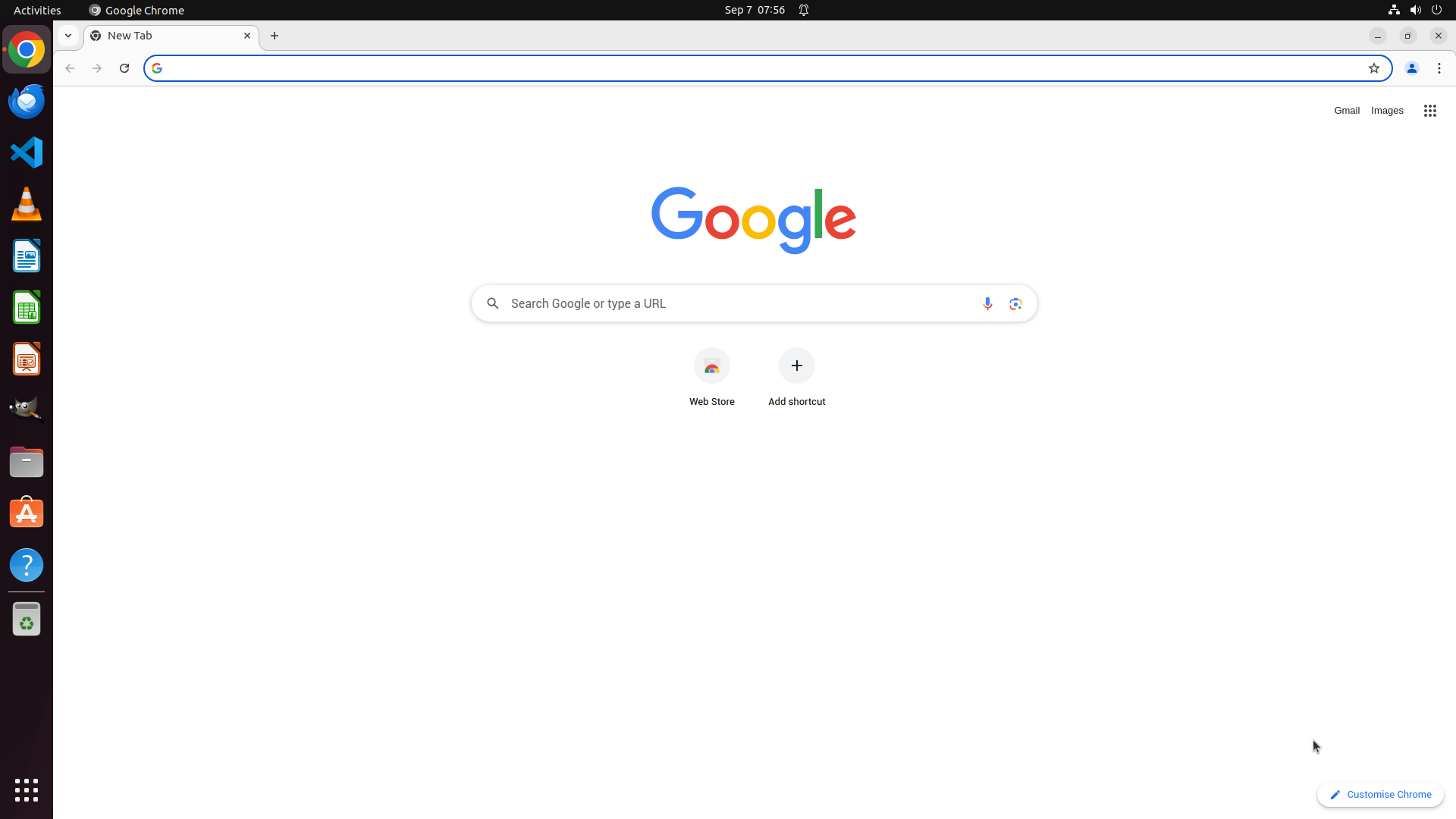}
&
Action: Click the circular user profile avatar button in the top-right of Chrome's toolbar to open the profile menu.
\caseToolCall{
    \{\textquotedbl{}name\textquotedbl{}: \textquotedbl{}computer\_use\textquotedbl{}, \textquotedbl{}arguments\textquotedbl{}: \{\textquotedbl{}action\textquotedbl{}: \textquotedbl{}left\_click\textquotedbl{}, \textquotedbl{}coordinate\textquotedbl{}: [968, 81]\}\}
}
\\

\midrule

2
& \caseScreenshot{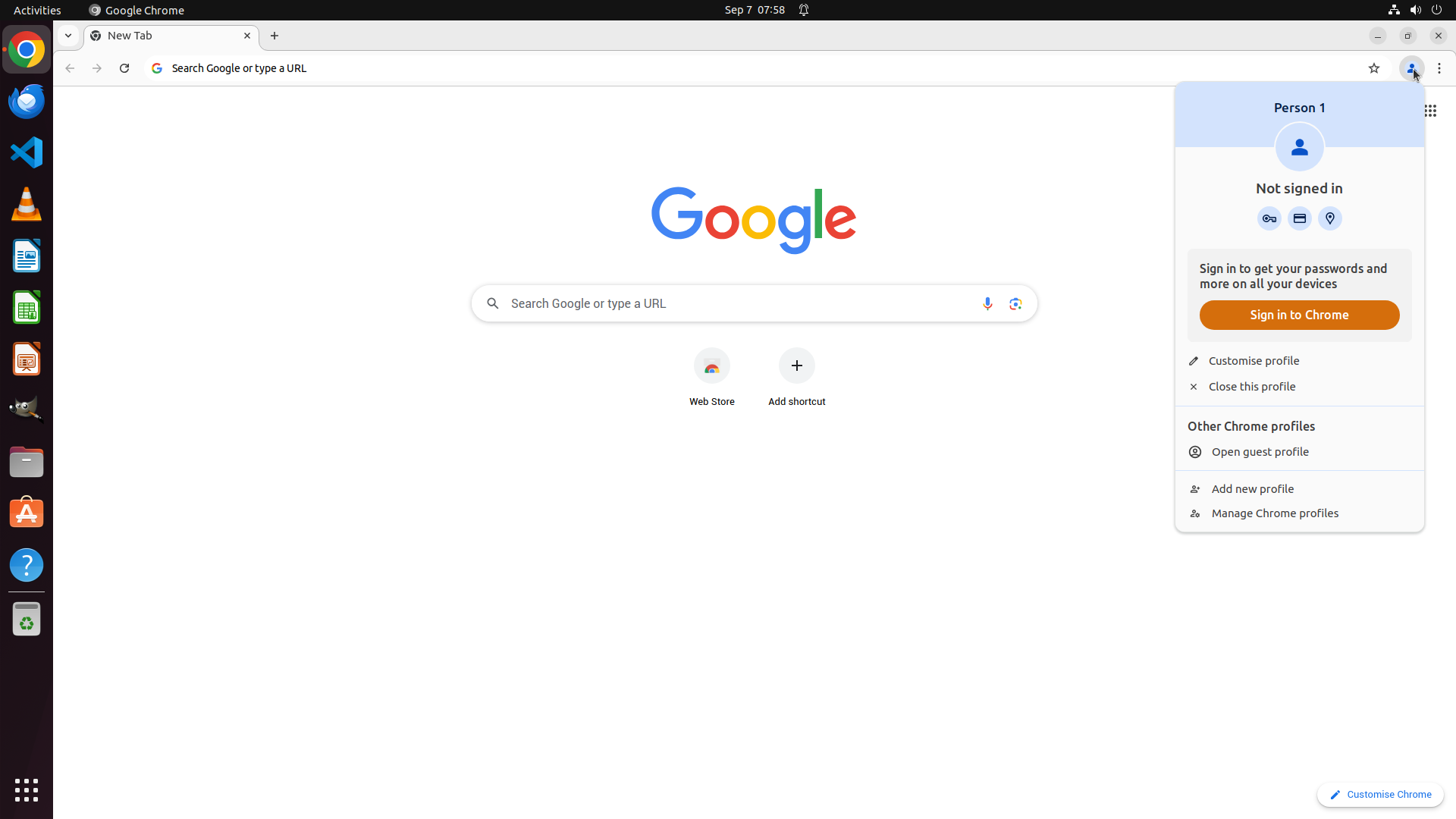}
&
Action: Click the ``Customise profile'' option in the open Chrome profile menu to open the profile customization settings.
\caseToolCall{
    \{\textquotedbl{}name\textquotedbl{}: \textquotedbl{}computer\_use\textquotedbl{}, \textquotedbl{}arguments\textquotedbl{}: \{\textquotedbl{}action\textquotedbl{}: \textquotedbl{}left\_click\textquotedbl{}, \textquotedbl{}coordinate\textquotedbl{}: [877, 438]\}\}
}
\\

\midrule

3
& \caseScreenshot{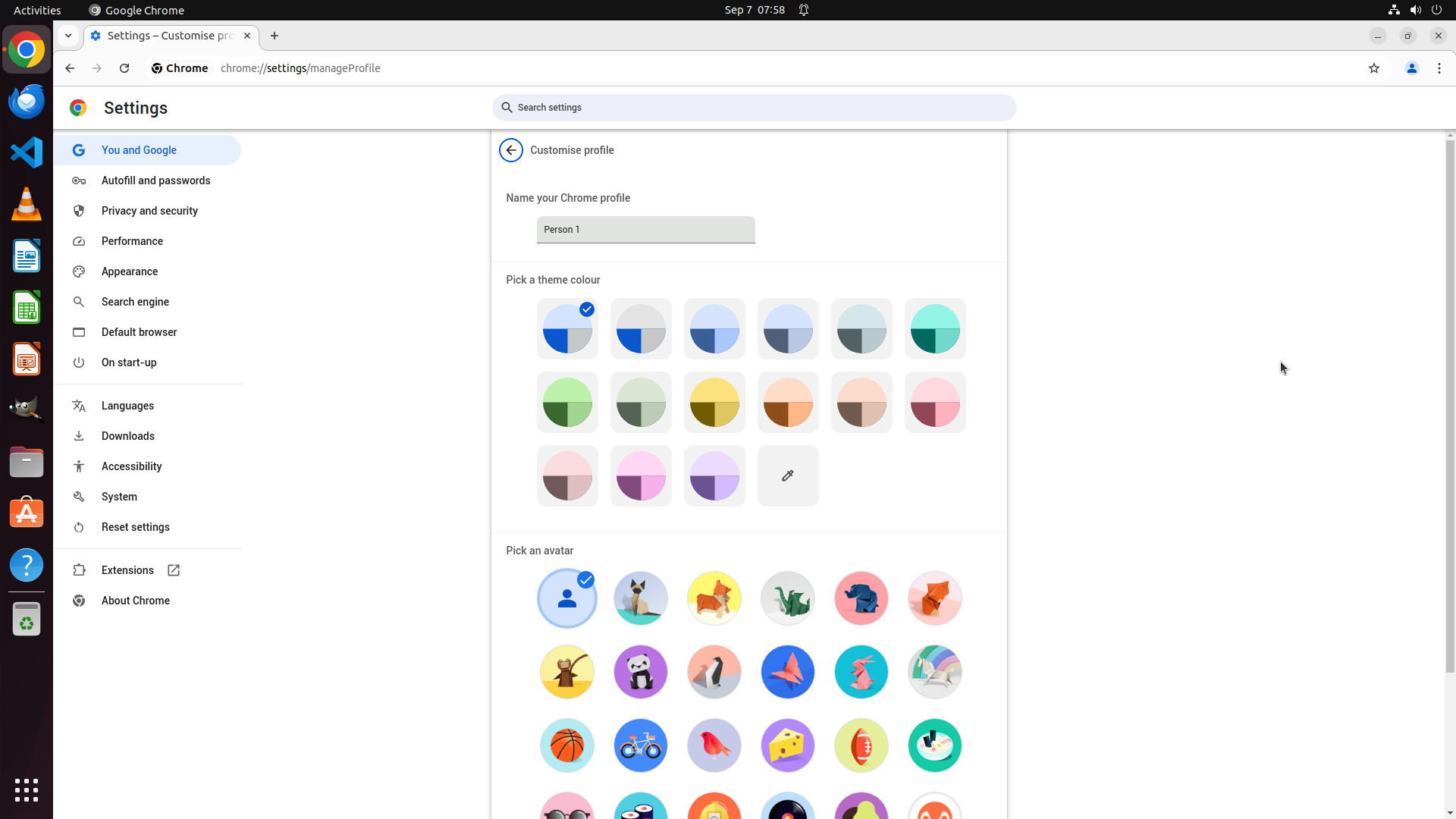}
&
Action: Click the ``Name your Chrome profile'' text input that currently displays ``Person 1'' to focus the field.
\caseToolCall{
    \{\textquotedbl{}name\textquotedbl{}: \textquotedbl{}computer\_use\textquotedbl{}, \textquotedbl{}arguments\textquotedbl{}: \{\textquotedbl{}action\textquotedbl{}: \textquotedbl{}left\_click\textquotedbl{}, \textquotedbl{}coordinate\textquotedbl{}: [404, 280]\}\}
}
\\

\midrule

4
& \caseScreenshot{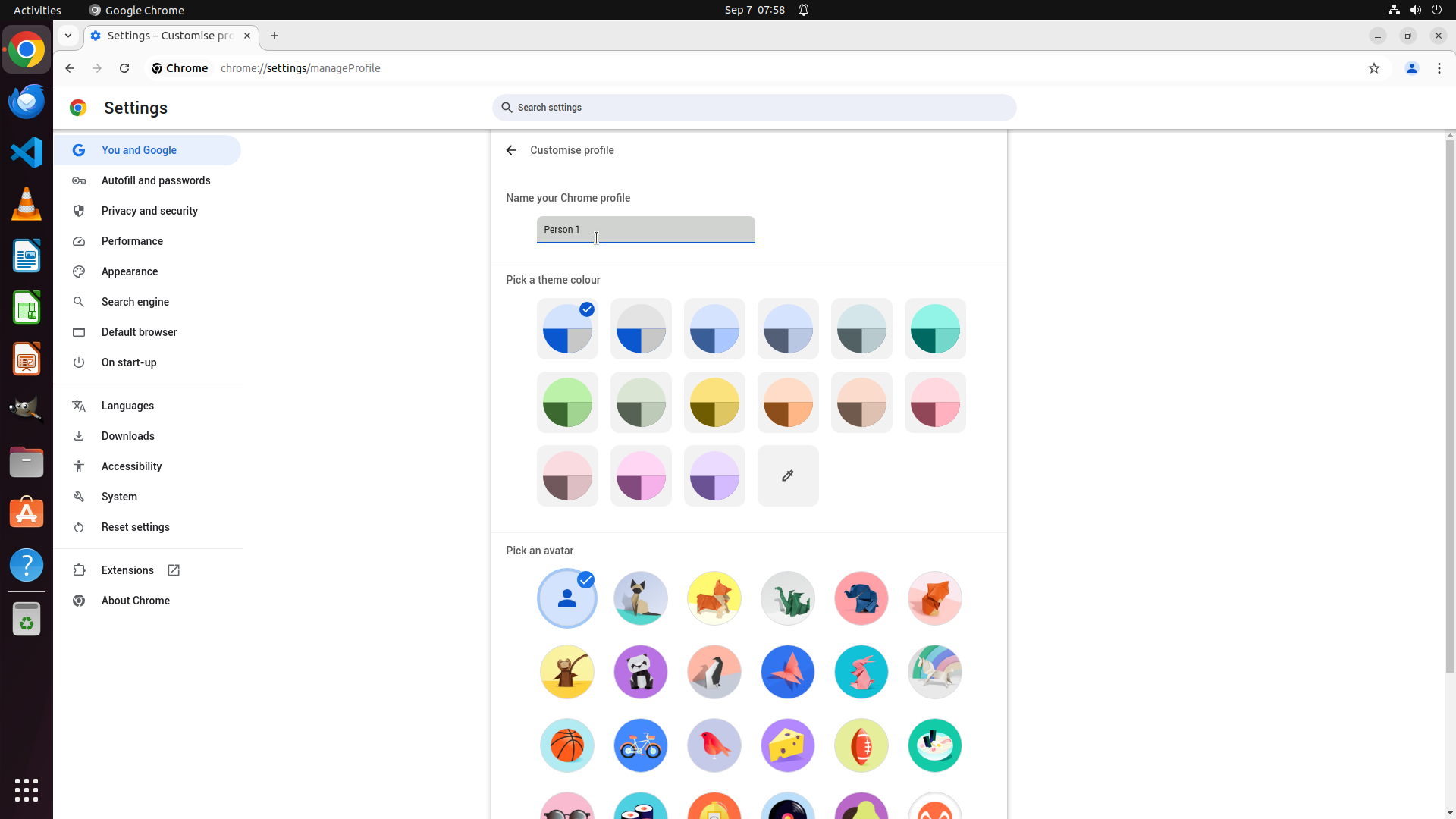}
&
Action: Press Ctrl+A to select all text in the focused ``Name your Chrome profile'' input field.
\caseToolCall{
    \{\textquotedbl{}name\textquotedbl{}: \textquotedbl{}computer\_use\textquotedbl{}, \textquotedbl{}arguments\textquotedbl{}: \{\textquotedbl{}action\textquotedbl{}: \textquotedbl{}key\textquotedbl{}, \textquotedbl{}keys\textquotedbl{}: [\textquotedbl{}ctrl\textquotedbl{}, \textquotedbl{}a\textquotedbl{}]\}\}
}
\\

\midrule

5
& \caseScreenshot{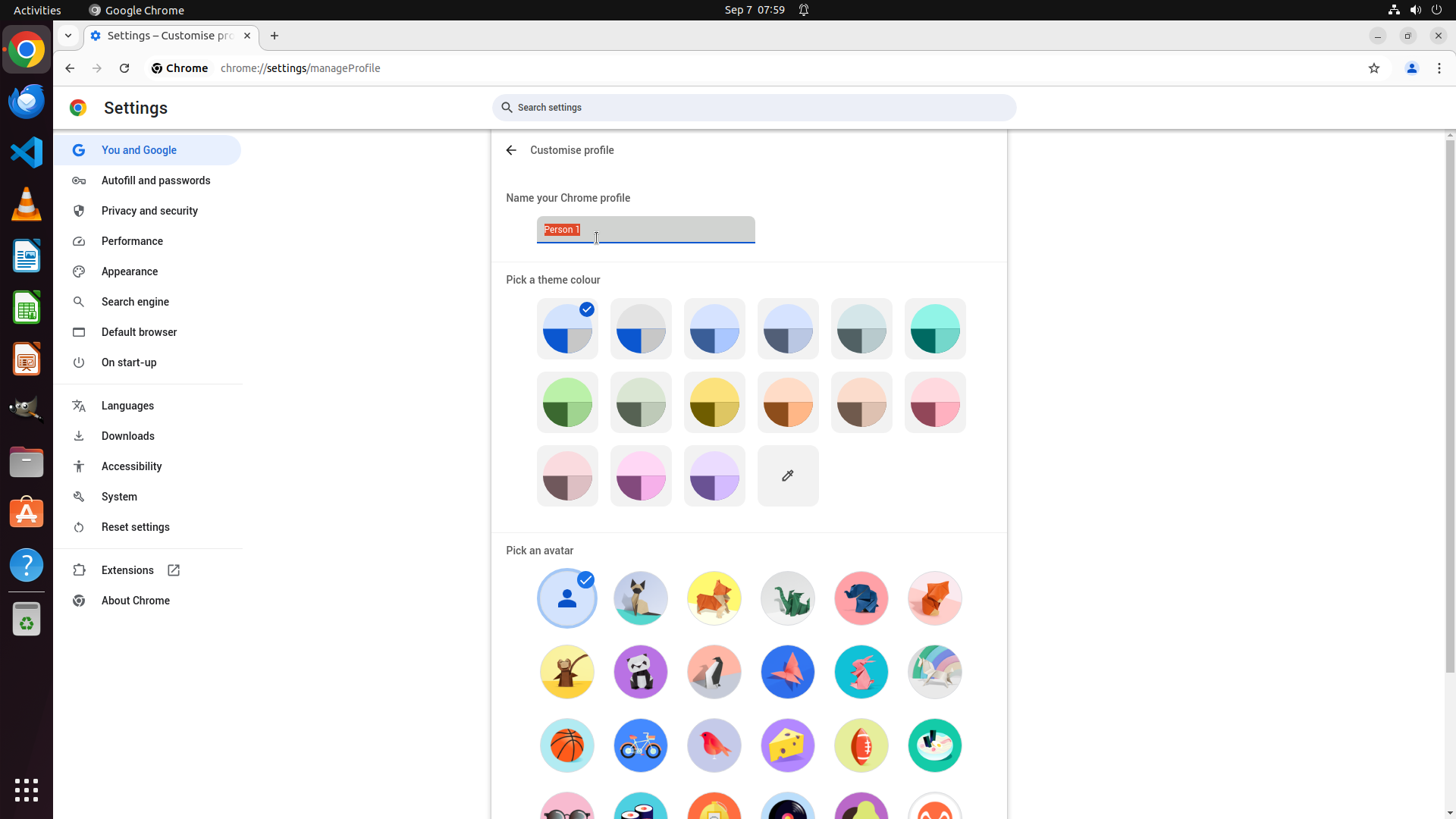}
&
Action: Type the text exactly: Thomas. This will replace the selected ``Person 1'' in the focused ``Name your Chrome profile'' field so it reads ``Thomas.''
\caseToolCall{
    \{\textquotedbl{}name\textquotedbl{}: \textquotedbl{}computer\_use\textquotedbl{}, \textquotedbl{}arguments\textquotedbl{}: \{\textquotedbl{}action\textquotedbl{}: \textquotedbl{}type\textquotedbl{}, \textquotedbl{}text\textquotedbl{}: \textquotedbl{}Thomas\textquotedbl{}\}\}
}
\\

\midrule

6
& \caseScreenshot{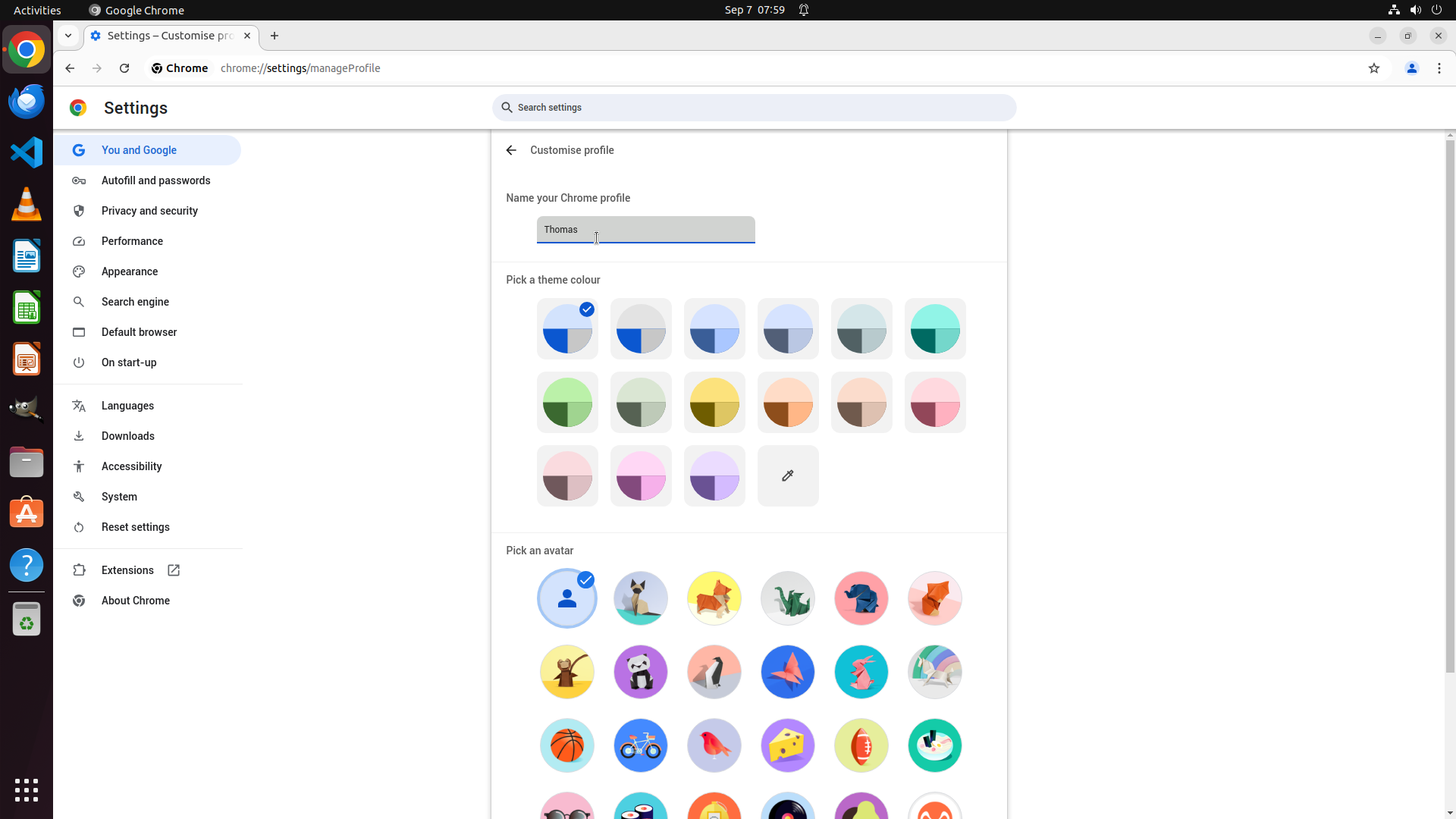}
&
Action: Click the left-pointing back arrow icon next to the ``Customise profile'' title at the top-left of the Chrome settings content area to return to the previous settings page.
\caseToolCall{
    \{\textquotedbl{}name\textquotedbl{}: \textquotedbl{}computer\_use\textquotedbl{}, \textquotedbl{}arguments\textquotedbl{}: \{\textquotedbl{}action\textquotedbl{}: \textquotedbl{}left\_click\textquotedbl{}, \textquotedbl{}coordinate\textquotedbl{}: [350, 181]\}\}
}
\\

\midrule

7
& \caseScreenshot{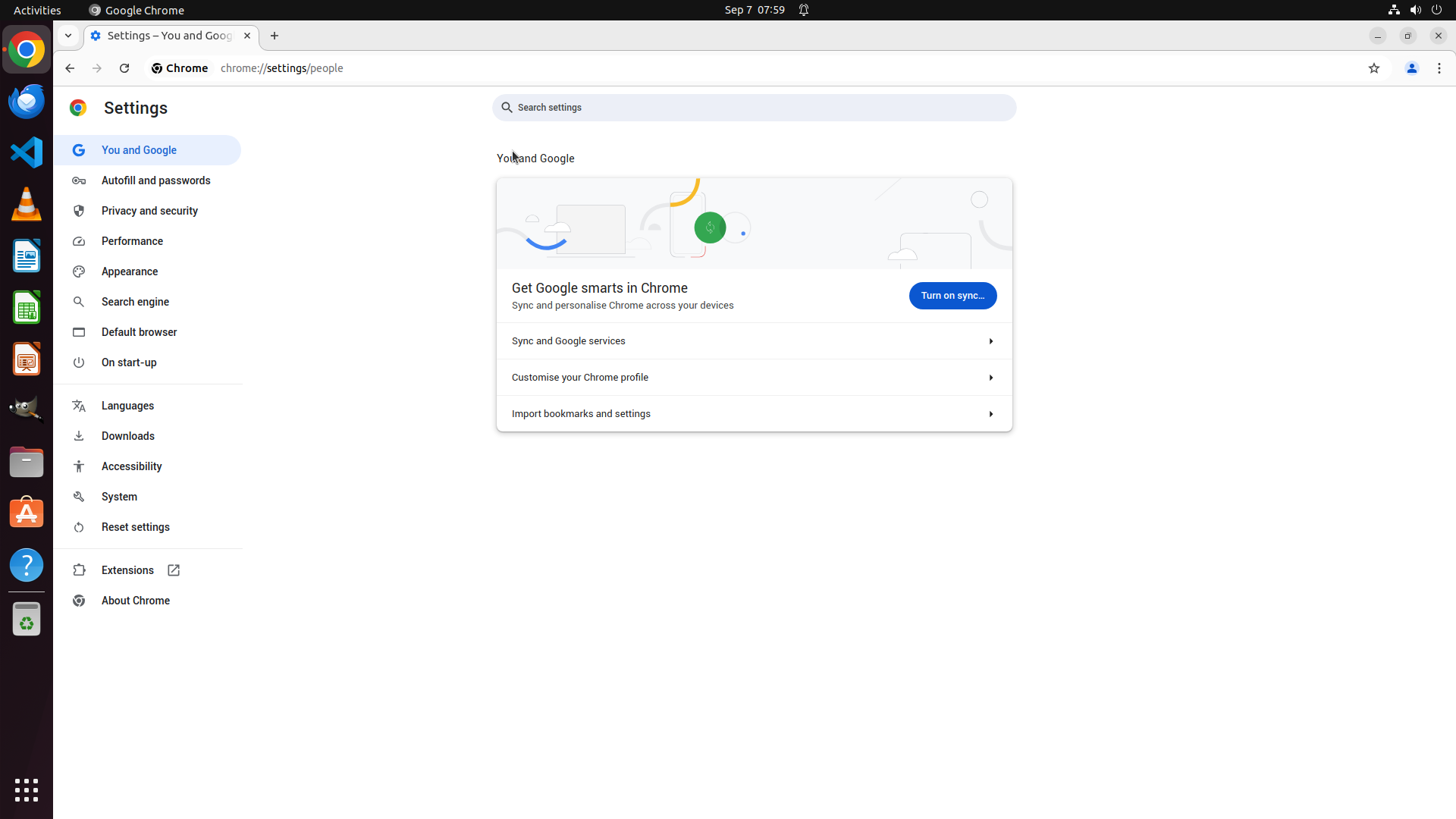}
&
Action: Click the circular user profile avatar button in the top-right of Chrome's toolbar to open the profile menu.
\caseToolCall{
    \{\textquotedbl{}name\textquotedbl{}: \textquotedbl{}computer\_use\textquotedbl{}, \textquotedbl{}arguments\textquotedbl{}: \{\textquotedbl{}action\textquotedbl{}: \textquotedbl{}left\_click\textquotedbl{}, \textquotedbl{}coordinate\textquotedbl{}: [968, 81]\}\}
}
\\

\midrule

8
& \caseScreenshot{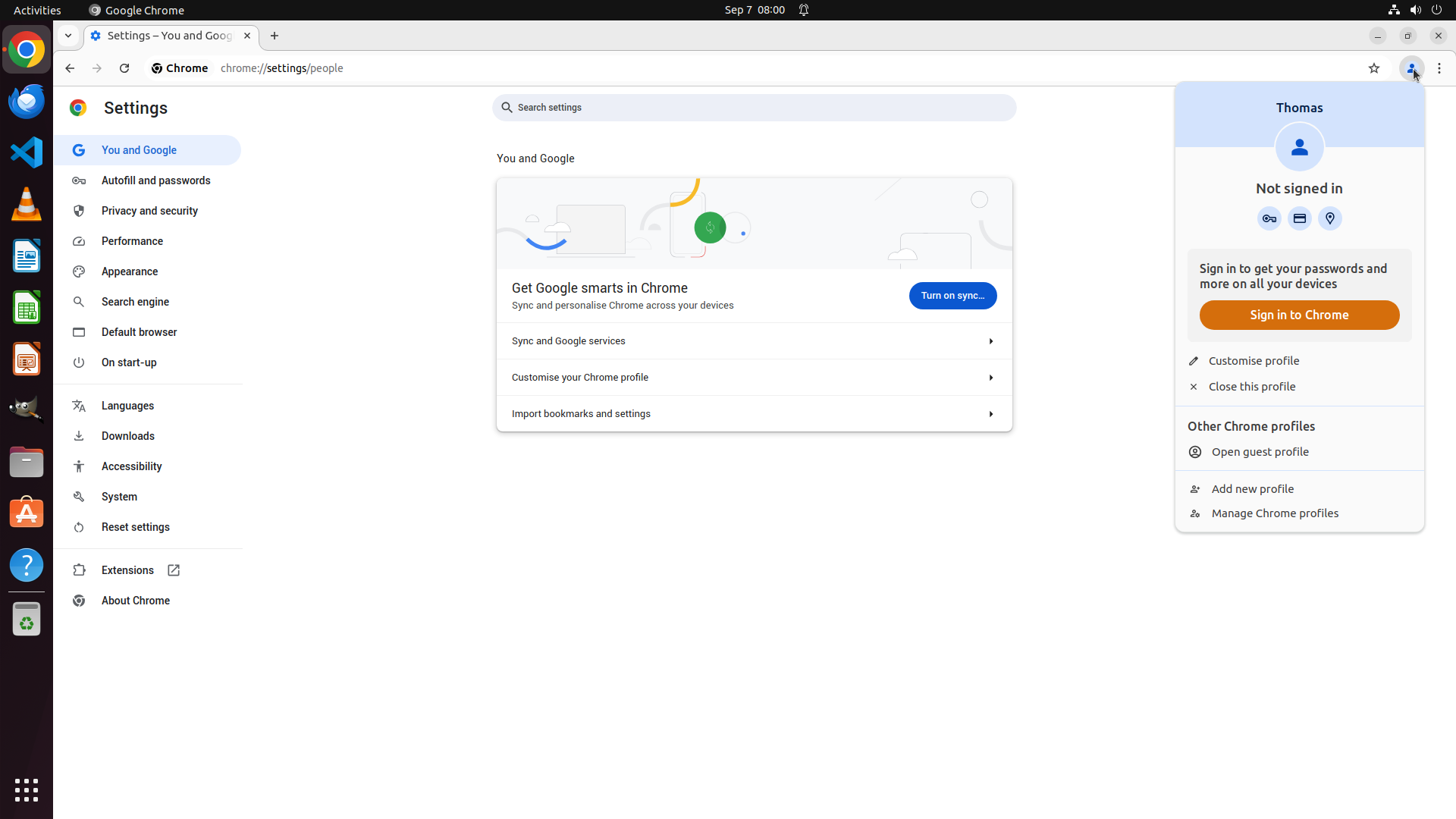}
&
Action: Click the ``Thomas'' profile name header at the top of the open Chrome profile menu in the top-right to open the Thomas profile home page.
\caseToolCall{
    \{\textquotedbl{}name\textquotedbl{}: \textquotedbl{}computer\_use\textquotedbl{}, \textquotedbl{}arguments\textquotedbl{}: \{\textquotedbl{}action\textquotedbl{}: \textquotedbl{}left\_click\textquotedbl{}, \textquotedbl{}coordinate\textquotedbl{}: [891, 137]\}\}
}
\\

\midrule

9
& \caseScreenshot{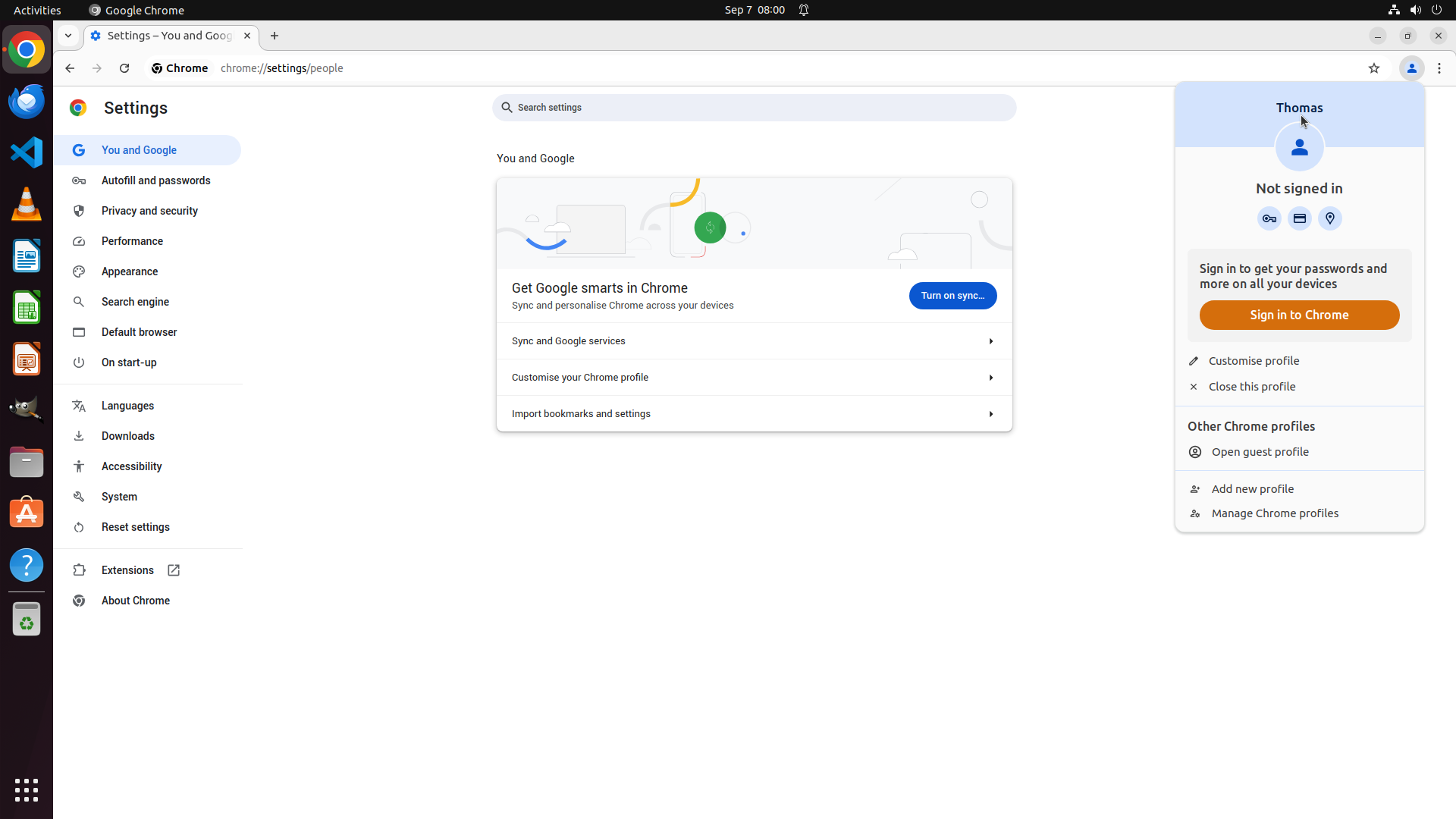}
&
Action: Terminate the session and mark the task as successfully completed.
\caseToolCall{
    \{\textquotedbl{}name\textquotedbl{}: \textquotedbl{}computer\_use\textquotedbl{}, \textquotedbl{}arguments\textquotedbl{}: \{\textquotedbl{}action\textquotedbl{}: \textquotedbl{}terminate\textquotedbl{}, \textquotedbl{}status\textquotedbl{}: \textquotedbl{}success\textquotedbl{}\}\}
}
\\

\bottomrule
\end{longtblr}

\endgroup

\end{document}